\documentclass[shortAfour,times,sageh]{sagej}

\usepackage{moreverb,url}
\newcommand\BibTeX{{\rmfamily B\kern-.05em \textsc{i\kern-.025em b}\kern-.08em
T\kern-.1667em\lower.7ex\hbox{E}\kern-.125emX}}

\usepackage{amsthm,amssymb,natbib,url,amsmath}
\usepackage{graphicx}
\usepackage{graphics}
\usepackage{clrscode3e}
\usepackage{caption}
\usepackage{subcaption}
\usepackage{bm}

\theoremstyle{plain}\newtheorem{thm}{Theorem}
\theoremstyle{definition}\newtheorem{defn}{Definition}
\theoremstyle{plain}
\theoremstyle{plain}

\usepackage{xcolor,colortbl}
\definecolor{Gray}{gray}{0.85}
\newcolumntype{g}{>{\columncolor{Gray}}c}
\newcolumntype{w}{>{\columncolor{white}}c}

\DeclareMathOperator{\proj}{proj}
\DeclareMathOperator{\codim}{codim} %A^\mathrm{o}

\newcommand{\smplr}{\psi}
\newcommand{\smplrSet}{\Psi}

\newcommand{\incremental}{incremental}
\newcommand{\focused}{focused}
\newcommand{\elements}{\id{elements}}
\newcommand{\mixed}{\id{mixed\_elements}}
\newcommand{\ext}[2]{#1{\uparrow^{#2}}}
\newcommand{\tgnt}[2]{T_{#1}(#2)}

\newcommand{\lzp}[1]{l_{#1}^P}
\newcommand{\lzg}[1]{l_{#1}^G}

\newcommand{\lzm}[1]{l_m^{#1}}
\def\D{d}%\def\D{\mathrm{d}}

\usepackage[disable]{todonotes}
\newcommand\note[1]{\todo[inline, color=blue!10, linecolor=blue!90,
  size=\footnotesize]{\linespread{0.9}\selectfont{{\bf TODO:} #1}\par}}

\def\volumeyear{2017}

\begin{document}

\runninghead{Garrett et al.}

\title{Sampling-Based Methods for Factored\\Task and Motion Planning}

\author{Caelan Reed Garrett\affilnum{1}, Tom\'as Lozano-P\'erez\affilnum{1}, and Leslie Pack Kaelbling\affilnum{1}}

\affiliation{\affilnum{1}MIT CSAIL, USA}

\corrauth{Caelan Reed Garrett,
Computer Science and Artificial Intelligence Laboratory,
32 Vassar Street,
Cambridge, MA 02139 USA}

\email{caelan@csail.mit.edu}

\begin{abstract}
%There has been a great deal of progress in developing probabilistically complete methods that move beyond motion planning to multi-modal problems including various forms of task planning.  
% In general, these new methods each require a new formulation, definition of robust feasibility, sampling methods, and search algorithm. 
This paper presents a general-purpose formulation of a large class of discrete-time planning problems, with hybrid state and control-spaces, as factored transition systems.
Factoring allows state transitions to be described as the intersection of several constraints each affecting a subset of the state and control variables. 
Robotic manipulation problems with many movable objects involve constraints that only affect several variables at a time and therefore exhibit large amounts of factoring.
We develop a theoretical framework for solving factored transition systems with sampling-based algorithms.
The framework characterizes conditions on the submanifold in which solutions lie, leading to a characterization of robust feasibility that incorporates dimensionality-reducing constraints.
It then connects those conditions to corresponding conditional samplers that can be composed to produce values on this submanifold.
We present two domain-independent, probabilistically complete planning algorithms that take, as input, a set of conditional samplers.
We demonstrate the empirical efficiency of these algorithms on a set of challenging task and motion planning problems involving picking, placing, and pushing. 
\end{abstract}

\note{It would be nice to highlight the ease of specification or something.}


\maketitle

%%%%%%%%%%%%%%%%%%%%%%%%%%%%%%%%%%%%%%%%%%%%%%%%%%%%%%%

\section{Introduction}

%We are interested in planning for autonomous systems operating in the physical world.
%These systems often involve both continuous and discrete state and control variables.
%Additionally, many systems are extremely high-dimensional. 
%The state of a robot operating in a human environment includes not only the configuration of the robot, but also the poses of each object in the world.
%Fortunately, the dynamics for these systems are often factorizable into pieces which only affect a small component of the state at once.
%% Motion planning example?

Many important robotic applications require planning in a high-dimensional space that includes not just the robot configuration, but also the ``configuration'' of the external world, including poses and attributes of objects. %, reaction states of chemical or biological processes, or intentions of other agents.
There has been a great deal of progress in developing probabilistically complete sampling-based methods that move beyond motion planning to these hybrid problems %multi-modal problems 
including various forms of task planning.  
These new methods each require a new formulation, theoretical framework, sampling method, and search algorithm.
We propose a general-purpose abstraction of sampling-based planning for a class of hybrid systems that implements each of these requirements.
As our motivating application, we apply this abstraction to robot task-and-motion planning.

%model robot task-and-motion planning problems
We model planning using {\em factored transition systems}, discrete-time planning problems involving mixed discrete-continuous state and control-spaces.
This formulation is able to highlight any {\em factoring} present within the problem resulting from constraints that only impact a few variables at a time.
Directly exposing factoring enables to us to design algorithms that are able to efficiently sample states and controls by sampling values for subsets of the variables at a time.
Additionally, factoring enables the use of efficient discrete search algorithms from artificial intelligence planning community.

The theoretical contribution of this paper is an analysis of the topology of a problem's solution space, particularly in the presence of dimensionality-reducing constraints. 
%The key insight is that the minimal-dimension manifold containing the set of solutions can often be constructed by composing projection preimages, intuitively submanifolds defined by fixing the values of some parameters.  
The key insight is that, in some cases, 
%the intersection of constraint manifolds can be constructed by composing the conditioned versions of the individual constraint manifolds in a sequence.
the intersection of several lower-dimensional constraints lies on a submanifold of the parameter-space that can be identified using only the individual constraints.
By understanding the topology of the solution space, we define a property that characterizes a large class of problems for which sampling-based planning methods can be successful. 

\begin{figure}[ht]
\centering
%\raisebox{.1\height}{\includegraphics[width=0.24\textwidth]{figures/several_small}}
%\raisebox{.3\height}{\includegraphics[width=0.11\textwidth]{figures/rearrangement_small}}
\includegraphics[width=0.245\textwidth]{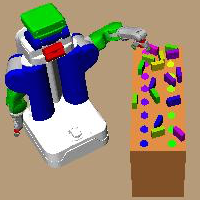}
\includegraphics[width=0.22\textwidth]{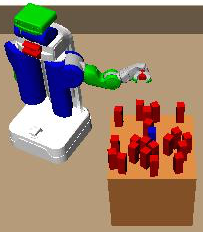}
\caption{{\it Experiment 1} (left): the robot must place each block at its corresponding goal pose. {\it Experiment 3} (right): the robot must move the blue block to another table.} \label{fig:exp3}
\end{figure}

The algorithmic contribution is the construction of two sampling-based planning algorithms that exploit the factored, compositional structure of the solution space to draw samples using {\em conditional samplers}.
%from a space in which solutions have positive measure.  
These algorithms search in a combined space that includes the discrete structure (which high-level operations, such as ``pick'' or ``place'' happen in which order) and parameters (particular continuous parameters of the actions) of a solution.  
Theoretically, these algorithms are probabilistically complete when given a set of sampling primitives that cover the appropriate spaces with probability one.
Practically, they can solve complex instances of task-and-motion planning problems as well as problems involving non-prehensile manipulation problems.

%\begin{itemize}
%\item Factored sampling - used in multiple states, reason about things that have not happened
%\item More efficient search using factoring
%\item lower-dimensionality actions
%\item Completeness with respect to generators
%\item Factoring is (relate to my FFRob Vega-Brown)
%\item Maybe show the values across time picture
%\item Low dimensional example (like the circle in the)
%\item Do I want to explain to decompose entirely and then combine?
%\item Why do we group variables the way we do? Several are mentioned in the same constraints together.
%\item Vega brown doesn't consider robot configurations or trajectories independently
%\item Starting with a factored description of the problem enables algorithms to take advantage of additional structure
%\item Increase sample efficiency by taking the cartesian product of values
%\item Can reason about controls in which most of the constraints are satisfied and plan to execute from a state that satisfies all the constraints
%\item The planning through an object phenomenon is because the same control works in multiple transitions. And we know of a pose that allows a collision free transition.
%\end{itemize}

%Simplicity of sampling
%Factored transition systems allow better sampling, grounding, and search
%Limited analytical descriptions of dynamics

\section{Related Work}

\note{Relation to mode sampling}

% Difference equation? Markov difference equation?
% https://en.wikipedia.org/wiki/Hybrid_system
% https://scholar.google.com/scholar?start=10&q=hybrid+systems&hl=en&as_sdt=0,22
%\cite{alur2000discrete} - Discrete Abstractions of Hybrid Systems -https://www.cis.upenn.edu/~alur/Ieee00.pdf
% \cite{alur1995algorithmic} - The algorithmic analysis of hybrid systems - https://www.sciencedirect.com/science/article/pii/030439759400202T
%\cite{henzinger2000theory} - The Theory of Hybrid Automata - http://www.informatik.uni-bremen.de/agbs/lehre/ss03/scs3/hintergrund-info/the_theory_of_hybrid_automata.pdf
% Discrete-Time Hybrid Modeling and Verification - https://pdfs.semanticscholar.org/c238/3d38daaae02a1c357a70594c6f44ab144504.pdf

% Describe other robustness conditions

Planning problems in which the goal is not just to move the robot without collision but also to operate on the objects in the world have been addressed from the earliest days of manipulation planning~\citep{LozanoPerez81,handeyICRA87,Wilfong89}.
\cite{Alami91,AlamiTwoProbs} decomposed manipulation problems using a {\em manipulation graph} that represents connected components of the configuration space given by a particular robot grasp. 
They observed that solutions are alternating sequences of {\em transit} and {\em transfer} paths corresponding to the robot moving while its hand is empty and the robot moving while holding an object. 
\cite{vendittelli2015decidability} proved a manipulation planning with a robot and two movable obstacles is decidable by providing a complete decomposition-based algorithm.
\cite{deshpande2016tamp} extended this result to general, prehensile task and motion planning.
Maintaining an explicit characterization of the free configuration space can be prohibitively expensive in high-dimensional planning problems.
% Instead than maintaining exact representations of the connected components of the configuration space, 

%In recent years, there have been a number of approaches to integrating discrete task planning and continuous motion planning~\citep{Cambon,Plaku} aimed at increasing the capabilities of autonomous robots.
\cite{simeon2004manipulation} extended the work of~\cite{Alami91,AlamiTwoProbs} by using probabilistic roadmaps (PRMs)~\citep{Kavraki96} to approximate each component of the configuration space.
%applied these ideas to manipulation planning using probabilistic roadmaps.
% No completeness analysis
The aSyMov system~\citep{Cambon} generalizes this strategy to task and motion planning with significant discrete structure.
It uses a heuristic computed by planning at just the symbolic level to guide the geometric exploration of its roadmaps.
\cite{Plaku} also use symbolic planning to influence geometric planning but for biasing sampling instead of guiding the search.

% Our work takes a view of planning as a problem of finding variable assignments subject to a set of constraints.  
% Some of the choices determine the discrete action sequence (the plan skeleton) and others the continuous geometric parameters.

\cite{StilmanWAFR06,StilmanICRA07} introduced the problem of robotic {\em navigation among movable obstacles} (NAMO), robotic motion planning in a reconfigurable environment.
They provide an algorithm for solving monotonic problem instances, problems that require moving each object at most one time.
\cite{van2009path} developed a probabilistically complete algorithm for robustly feasible NAMO problems.
% Conditions
\cite{krontirisRSS2015,krontiris2016icra} extended the work of \cite{StilmanWAFR06,StilmanICRA07} to nonmonotonic rearrangement problems, which require moving a set of objects from to specified goals poses, by using their algorithm as a primitive within a larger, complete search. 
% rearrangement
These algorithms are each specialized to a subclass of manipulation planning.

% Hauser uses sample-based instead of sampling-based
% HauserLatombe and Hauser
\cite{HauserIJRR11} introduced a framework and algorithm {\em multi-modal motion planning}, motion planning in overlapping spaces of non-uniform dimensionality.
% for probabilistically complete
\cite{barry2013hierarchical,barry2013manipulation} considered multi-modal motion planning using bidirectional rapidly-exploring random trees (RRT-Connect).
\cite{vega2016asymptotically} extended these ideas to optimal planning with differential constraints.
Because these approaches do not exploit any factoring present within a problem, they must sample entire states at once and are unable to take advantage of powerful heuristics to guide their search,
% Optimal, Sampling-Based Manipulation Planning

% Expansiveness - Multi-modal Motion Planning in Non-expansive Spaces
% http://journals.sagepub.com/doi/pdf/10.1177/0278364909352098

% Randomized multi-modal motion planning for a humanoid robot manipulation task
% Expansiveness - http://journals.sagepub.com/doi/pdf/10.1177/0278364910386985

% Robust feasibility
%  FFRob: Leveraging Symbolic Planning for Efficient Task and Motion Planning

% clearance - Path Planning among Movable Obstacles: A ProbabilisticallyCompleteApproach
% https://pdfs.semanticscholar.org/ff1f/4249d475c073ef2d69f671490dd60e941b04.pdf

% Asymptotically optimal planning under piecewise-analytic constraints
% http://wafr2016.berkeley.edu/papers/WAFR_2016_paper_11.pdf

\cite{dornhege09icaps,dornhege13irosws} introduced {\em semantic attachments}, external predicates evaluated on a geometric representation of the state, to integrate geometric reasoning into artificial intelligence planners.
Their algorithms assume a finite set of primitive actions which restricts them to discrete control-spaces.
They evaluate semantic attachments within their search which results in unnecessarily computing many expensive motion plans.
% No blackbox planner

\cite{HPN} introduced {\em generators} to select predecessor states in a goal regression search (HPN). % for task and motion planning.
\cite{GarrettIROS15} gave an algorithm (HBF) for planning in hybrid spaces by using approximations of the planning problem to guide the backward generation of successor actions to be considered in a forward search.
Both approaches requires that generators are specified according to an inverse model in order to be compatible with their backward searches.
Additionally, both approaches integrate search and sampling preventing them from leveraging discrete search algorithms as blackbox subroutines.

%\cite{Pandey12,deSilva} use {\em hierarchical task networks} (HTNs)~\citep{erol1994htn} to search over plan skeletons and backtrack upon failing to find satisfactory geometric parameters.  
\cite{Pandey12,deSilva} use {\em hierarchical task networks} (HTNs)~\citep{erol1994htn} to guide a search over {\em plan skeletons}, discrete sequences of actions with unbound continuous variables, using knowledge about the task decomposition.
The search over plan skeletons backtracks in the event that it is unable to bind the free continuous variables of a skeleton.
\cite{lozano2014constraint} take a similar approach but leverage constraint satisfaction problem (CSP) solvers operating on discretized variable domains to bind the free variables.
%to identify satisfactory geometric parameters from discretized domains.
%\cite{LagriffoulDSK12,lagriffoul2014efficiently} interleave the search for a discrete action sequence and the geometric parameters and focus on limiting the amount of geometric backtracking.
%perform a search over {\em plan skeletons}, discrete sequences of actions that does not yet assign values for the continuous variables.
\cite{LagriffoulDSK12,lagriffoul2014efficiently} also search over plan skeletons but are able to prune some plan skeletons from consideration using computed bounds on the constraints.
For each plan skeleton under consideration, they generate a set of approximate linear constraints, {\it e.g.} from grasp and placement choices, and use linear programming to compute a valid assignment of continuous values or determine that one does not exist. 
Similarly,~\cite{toussaint2015logic,toussaint2017multi} formulate the binding of geometric variables as a nonlinear constrained optimization problem and use a hierarchy of bounds on the nonlinear program to prune plan skeletons.
Because binding and pruning operate globally on entire plan skeletons, these approaches are unable to identify individual variables and constraints that primarily contributed to infeasibility.
Thus, the search over plan skeletons may evaluate many similar plan skeletons that exhibit the same behavior.
%all of these approaches may require evaluating many plan skeletons even when only a single constraint causes infeasibility. 
In contrast, by reasoning locally about individual conditional samplers, our \focused{} algorithm is able to retain samples that satisfy their associated constraints and focus the subsequent search and sampling on conditional samplers that failed to produce satisfactory samples.

%\cite{toussaint2015logic} formulates task and motion planning as a logic-geometric program, a non-linear constrained optimization problem augmented with a logic and knowledge base.  
%He hierarchically optimizes the final state, transfer configurations, and motion trajectories in a search over plan skeletons.
%In the event that an optimization is infeasible, the search may need to backtrack over many plan-skeletons.

% \cite{toussaint2015logic} expressed task and motion planning as a logic-geometric program, a constrained optimization problem endowed with a logic over a knowledge base.
% He provide an algorithm for solving problem instances by hierarchically decomposing problems into tree search and local optimization phases.

%Their approach performs a discrete search in the space of
%action sequences and uses a CSP solver (in discretized domains) to
%determine whether a valid set of action parameters completes the plan
%skeleton.

The FFRob algorithm of~\cite{GarrettWAFR14} samples a set of object poses and robot configurations and then plans with them using a search algorithm that incorporates geometric constraints in its heuristic.
An iterative version of FFRob that repeats this process until a solution is found is probabilistically complete and exponentially convergent~\citep{garrettIJRR2017}.
Our \incremental{} algorithm can be seen as generalizing this strategy of iteratively sampling then searching from pick-and-place domains to domains with arbitrary conditional samplers.
%The FFRob algorithm of~\cite{GarrettWAFR14,garrettIJRR2017} is related to the \incremental{} algorithm discussed in this paper.
%It also involves sampling a fixed set of object poses and robot configurations and then planning with them using artificial intelligence techniques. 
%An iterative version of FFRob is probabilistically complete and exponentially convergent~\citep{garrettIJRR2017}.
%However, the approach in FFRob is specialized to pick-and-place problems.  %and does not use planning to guide sampling. 
%Additionally, because it blindly samples poses, configurations, and motions, it may generate many unnecessary values resulting large planning overhead. 

~\cite{Erdem} plan at the task-level using a boolean satisfiability (SAT) solver, initially ignoring geometric constraints, and then attempt to produce motion plans satisfying the task-level actions. 
If an induced motion planning problem is infeasible, the task-level description is updated to indicate motion infeasibility using a domain-specific diagnostic interface.
%with additional action preconditions indicating the motion infeasibility. 
% Their algorithm requires a custom interface to diagnose 
%Their work requires a domain-specific interface for diagnosing motion infeasibility and updating the task-level description.
~\cite{dantam2016tmp} extend this work by using a satisfiability modulo theories (SMT) solver to incrementally add new constraints and restart the task-level search from its previous state.
Their approach also adjusts to motion planning failures automatically and in a way that allows previously failed motion planning queries to be reconsidered.
The algorithms of~\cite{Erdem} and~\cite{dantam2016tmp} both assume an {\it a priori} discretization of all continuous values apart from configurations, for example, objects placements.
~\cite{Srivastava14} remove this restriction by using symbolic references to continuous parameters.
Upon finding a task-level plan, they use a domain-specific interface, like~\cite{Erdem}, to bind values for symbolic references and update the task-level description when none are available.
Our \focused{} algorithm is related to these approaches in that it lazily evaluates constraints and plans with lazy samples before real values.
However, it is able to automatically manage its search and sampling in a probabilistically complete manner using each individual conditional sampler as a blackbox, without the need of a domain-specific interface.

Our work captures many of the insights in these previous approaches in a general-purpose framework.
It also highlights the role of factoring in developing efficient algorithms for sampling relevant values and searching discretized spaces.
%as well as more thoroughly exploiting the mutual constraints between plan skeletons and geometric parameters.
% Generalize

\section{Factored Transition System}

%  dynamic stability, reachability, and verification
We begin by defining a general class of models for deterministic, observable, discrete-time, hybrid systems. % dynamical, controllable, continuous
%We introduce a factored representation for modeling discrete-time, hybrid systems.
These systems are {\em hybrid} in that they have mixed discrete-continuous state and control-spaces.
However, they are also {\em discrete-time} meaning that transitions are discrete changes to the hybrid state~\citep{torrisi2001discrete}.
This is in contrast to a {\em continuous-time} hybrid systems~\citep{alur1995algorithmic,alur2000discrete,henzinger2000theory} which allow continuous transitions described as differential equations. 
%Intuitively, an agent controlling a system is restricted to executing a finite sequence of hybrid control inputs.
It is possible to address many continuous-time problems in this framework, as long as they can be solved with a finite sequence of continuous control inputs.
%For example, each control input could be a zero-order hold torque parameterized by the magnitude and duration of the applied force.

% Leslie - But the length doesn't need to be determined in advance.
%We will then explore conditions under which the problem of ``planning'', finding a finite sequence of controls that drives the system into a desired state, is feasible for these models using sampling-based planning approaches.  Finally, we will provide effective algorithms for finding solutions to problems in this class.
\begin{defn}
A discrete-time {\em transition system} ${\cal S} = \langle {\cal X}, {\cal U}, {\cal T} \rangle$ is defined by a set of states ({\em state-space}) ${\cal X}$, set of controls ({\em control-space}) ${\cal U}$, and a {\em transition relation} ${\cal T} \subseteq {\cal X} \times {\cal U} \times {\cal X}$.
\end{defn}
For many physical systems, ${\cal T}$ is a {\em transition function} from ${\cal X} \times {\cal U}$ to ${\cal X}$. 
We are interested in controlling a transition system to move from a given initial state to a state within a goal set.
%, but it will be useful to write it in this relation form.
% $f(x, u) \in {\cal X}$ which is defined for $x \in {\cal X}$ and $u \in U(x) = \{u \mid (x, u, x') \in {\cal T}\}$. %We often wish to find plans that control such systems to transition between sets of states. 
%For an initial state $x_0 \in {\cal X}$ and a goal set of states $X_* \subseteq {\cal X}$,
\begin{defn}
A {\em problem} ${\cal P} = \langle x^0, X^*, {\cal S} \rangle$ is an initial state $x^0 \subseteq {\cal X}$, a set of goal states $X^* \subseteq {\cal X}$, and a transition system ${\cal S}$.
\end{defn}
%A {\em domain} ${\cal D} = \{{\cal P},...\}$ is a set of problems.
\begin{defn}
A {\em plan} for a problem ${\cal P}$ is finite sequence of $k$ control inputs $(u^1, ..., u^k)$ and $k$ states $(x^1, ..., x^k)$ such that $(x^{i-1}, u^i, x^i) \in {\cal T}$ for $i \in \{1, ..., k\}$ and $x^k \in X^*$.
\end{defn}
When ${\cal T}$ is a transition function, a plan can be uniquely identified by its sequence of control inputs.

\subsection{Factoring}
% Introduce modes for the robot to move within?
We are particularly interested in transition systems that are factorable.
As we will show in sections~\ref{sec:samplers} and~\ref{sec:search}, factored structure is useful for developing efficient methods for sampling and searching transition systems.

\begin{defn}
A {\em factored transition system} is a transition system with state-space $\bar{\cal X} = {\cal X}_1 \times ... \times {\cal X}_m$ and control-space $\bar{\cal U} = {\cal U}_1 \times ... \times {\cal U}_n$ that is defined by $m$ {\em state variables} $\bar{x} = (x_1, ..., x_m)$ and $n$ {\em control variables} $\bar{u} = (u_1, ..., u_n)$. 
\end{defn}
%For a continuous system, each ${\cal X}_i \subseteq \mathbb{R}$ can be a degree-of-freedom of the system.
%A trivial decomposition always exists by letting $n=1$ and $m=1$. 
% $x = (x[1], ..., x[m])$
% $u = (u[1], ..., u[n])$
The transition relation is a subset of the {\em transition parameter-space} ${\cal T} \subseteq \bar{\cal X} \times \bar{\cal U} \times \bar{\cal X}$
%$${\cal T} \subseteq {\cal X}_1 \times ... \times {\cal X}_m \times {\cal U}_1 \times ... \times {\cal U}_n \times {\cal X}_1 \times ... \times {\cal X}_m$$
%$${\cal T} \subseteq \Big(\bigotimes_{i=1}^m {\cal X}_i\Big) \times \Big(\bigotimes_{i=1}^n {\cal U}_i\Big) \times \Big(\bigotimes_{i=1}^m {\cal X}_i\Big).$$
% But ${\cal T} = {\cal X}  \times {\cal U} \times {\cal X}$ is still true
Valid transitions are $(x_1, ..., x_m, u_1, ..., u_n, x'_1, ..., x'_m) \in {\cal T}$. % Indexing them together $2n + m$
%A parameter is distinct from a variable. $x_v$ defines a parameter
To simplify notation, we will generically refer to each $x_p$, $u_p$, or $x'_p$ in a transition as a {\em parameter} $z_p$ where $p \in \{1,...,2m+n\} = \Theta$ indexes the entire sequence of variables.
\note{Make $x_p$, $u_p$, or $x'_p$?}
%encodes $x$, $u$, or $x'$ as well as the variable $v$. Let ${\cal Z}_i$ be the corresponding domain of parameter $i$, either ${\cal X}_v$ or ${\cal U}_v$. 
For a subset of parameter indices $P = (p_1, ..., p_k) \subseteq \Theta$, let $\bar{z}_P = (z_{p_1}, ..., z_{p_k}) \in \bar{\cal Z}_P$ be the combined values and $\bar{\cal Z}_P = {\cal Z}_{p_1} \times ... \times {\cal Z}_{p_k}$ be the combined domain of the parameters.
% Should probably order this
% Could rename variables to parameters $v \in V \subseteq {\cal V}$

%Many transition relations are the combination of several types of transitions that each involve the same parameters.
Many transition relations are hybrid, in that there is a discrete choice between different types of operation, each of which has a different continuous constraint on the relevant parameters.
For example, a pick-and-place problem has transitions corresponding to a robot moving its base, picking each object, and placing each object.
% where each type of transition affects the state and control variables differently.
In order to expose the discrete structure, we decompose the transition relation ${\cal T} = \bigcup_{a=1}^\alpha T_a$ into the union of $\alpha$ smaller {\em transition components} $T_a$.
% Emphasize discrete structure more?
% Leslie - Should we say someplace that the different discrete action classes may have different parameterizations and that the U we speak of here is the union of all the parameter choices that might be needed by any of the action classes and that the constraints in the clause for each action class will just mention the relevant subset of those U variables (so in fact we don't have to come up with an assignment to all the U variables)?
A transition relation $T_a$ often is the intersection of several constraints on a subset of the transition parameters.
\begin{defn}
A {\em constraint} is a pair $C = \langle P, R \rangle$ where $P \subseteq \Theta$ is a subset of parameter indices and $R \subseteq \bar{\cal Z}_P$ is a relation on sets ${\cal Z}_{p_1}, ..., {\cal Z}_{p_k}$. 
% $\{i_1, ..., i_k\} = {P} \subseteq \{1...{2m+n}\}$.
\end{defn}
A tuple of values that satisfy a constraint is called a constraint element.
\begin{defn}
A constraint {\em element} $C(v_{p_1}, ..., v_{p_k})$ is composed of a constraint $C = \langle P, R \rangle$ and variable values $(v_{p_1}, ..., v_{p_k}) = \bar{v}_P \in R$ for parameter indices $P = (p_1, ..., p_k)$.
\end{defn}
%To simplify notation, we will write a constraint $C = \langle P, R \rangle$ in the form $R(p_1, .., p_k)$ for $P = (p_1, ..., p_k)$. 
For instance, {\it pick} transitions involve constraints that the end-effector initially is empty, the target object is placed stably, the robot's configuration forms a kinematic solution with the placement,
%the object is reachable, the trajectory taken is collision-free, 
and the end-effector ultimately is holding the object.
A constraint decomposition is particularly useful when $|{P}| << 2m+n$; 
{\it i.e.}, each individual constraint has low arity.
Let $\ext{C}{\Theta} = \{\bar{z} \in \bar{\cal Z}_\Theta \mid \bar{z}_P \in R\}$ be the {\em extended form} of the constraint over all parameter indices $\Theta$.
This alternatively can be seen as a Cartesian product of $R$ with $\bar{\cal Z}_{\Theta \setminus P}$ followed by permuting the parameter indices to be in a sorted order.
% to include parameters ${\Phi} \setminus P$.
%The extended form can be seen as a Cartesian product followed by a permutation.
%$\widehat{C} = \sigma \cdot (C \times \bar{\cal Z}_{{\Phi} - P})$ where $\sigma = {P || ({\cal P} - P) \choose {\cal P}}$ is a permutation.
\note{$\bar{P} = (P_1, ..., P_n)$ as parameter names. $\bar{p} = (p_1, ...p_n)$ as values?}
\note{Say parameter indices instead of parameters in may places?}

\begin{defn}
A transition component $T_a$ is specified as the intersection over a {\em clause} of $\beta$ constraints ${\cal C}_a = \{C_1, ..., C_\beta\}$ where $T_a = \bigcap_{b=1}^\beta \ext{C_b^a}{\Theta}$.
\end{defn}
%$T_a = \bigcap_{b = 1}^\beta \widehat{C}_b$.
%$${\cal T}^i_j = \{(z_1, ..., z_{2m+n}) \in {\cal X}  \times {\cal U} \times {\cal X} \mid (z_{s_1}, ..., z_{s_k}) \in {C}^i_j, \{(s_1, ..., s_k\} = S^i_j\}$$
%$$T = \{(z_1, ..., z_{2m+n}) \in \bigotimes_{i =1}^{2m+n} {\cal Z}_i \mid (z_{i_1}, ..., z_{i_k}) \in C\}$$
%$${\cal T}^a_b = \{(z_1, ..., z_{2m+n}) \in \bar{\cal Z}_{[1...{2m+n}]} \mid (z_{i_1}, ..., z_{i_k}) \in {C}^a_b\}$$
%$$T_a = \{\bar{z} \in \bar{\cal Z} \mid \forall \langle I, C\rangle \in {\cal C}_a, \bar{z}_I \in C\}.$$
%$$T_a = \{\bar{z} \in \bar{\cal Z} \mid \forall C \in {\cal C}_a, C = \langle P, R\rangle, \bar{z}_P \in R\}.$$
%$$T_a = \bar{\cal Z} \cap \bigcap_{C \in {\cal C}_a} C$$
%Define $C_1 \cap C_2 = \{\bar{z} \mid \bar{z} \in \bar{\cal Z}_{P_1 \cup P_2}, \bar{z}_{P_1} \in  R_1, \bar{z}_{P_2} \in R_2  \}$
%Define $\bar{Z} \cap C = \{\bar{z} \mid \bar{z} \in \bar{\cal Z}, \bar{z}_{P} \in R \}$
%Each clause represents a intersection of its constraints.
%A clause is analogous to a conjunctive clause from ???
Membership in $T_a$ is equivalent to the {\em conjunction} over membership for each ${\cal C}_a$: $[\bar{z} \in T_a] \iff \bigwedge_{b=1}^\beta [\bar{z} \in \ext{C_b^a}{\Theta}]$.
Within a clause, there are implicit variable domain constraints on each parameter $z_p$ of the form $\id{Var}_p = \langle (z_p), {\cal Z}_p \rangle$. 
% Tomas - I would have wanted to see some examples of clauses, maybe not even formally, words would be fine, e.g. a clause is a \lztype? of transition such as move or a grasp? is it one for each object?  Something to make it more concrete.
Finally, the transition relation ${\cal T}$ is the union of $\alpha$ clauses $\{{\cal C}_1, ..., {\cal C}_\alpha\}$.
Membership in ${\cal T}$ is equivalent to the {\em disjunction} over membership for each ${\cal T}$: $[\bar{z} \in {\cal T}] \iff \bigvee_{a=1}^\alpha [\bar{z} \in T_a]$.
Thus, membership in ${\cal T}$ is a logical expression in {\em disjunctive normal form} over literals $[\bar{z} \in \ext{C_b^a}{\Theta}]$.
%This is the set analog of disjunctive normal form in logic. 
%Again, a trivial decomposition always exists through using one conjunctive clause that involves one constraint.
%$\widehat{C}$

%\begin{defn}
%A {\em factored transition system} is a transition system given by  a state-space $\bar{\cal X} = {\cal X}_1 \times ... \times {\cal X}_m$, control-space $\bar{\cal U} = {\cal U}_1 \times ... \times {\cal U}_n$, and a transition relation composed of a set of $\alpha$ conjunctive constraint clauses %$\big\{\{\langle {I}^1_1, {C}^1_1 \rangle, ..., \langle {I}^1_{\beta_1}, {C}^1_{\beta_1} \rangle\}, ..., \{\langle {I}^\alpha_1, {C}^\alpha_1 \rangle, ..., \langle {I}^\alpha_{\beta_\alpha}, {C}^\alpha_{\beta_\alpha} \rangle\}\big\}$.
%$\{{\cal C}_1, ..., {\cal C}_\alpha\}$.
%%$\{\langle {I}^1_1, {C}^1_1 \rangle, ..., \langle {I}^1_{\beta_1}, {C}^1_{\beta_1} \rangle\}, ..., \{\langle {I}^\alpha_1, {C}^\alpha_1 \rangle, ..., \langle {I}^\alpha_{\beta_\alpha}, {C}^\alpha_{\beta_\alpha} \rangle\}$.
%\end{defn}

%While some constraints may still be fairly complex, such as avoiding collisions, 
Factoring the transition relation can expose constraints that have a simple equality form. 
Equality constraints are important because they transparently reduce the dimensionality of the transition parameter-space. 
%Equality constraints capture the discrete component of a transition clause that alone can inform which sequences of transition clauses are valid.  
%Talk about how the equality are usually for making the discrete search interesting? Although, I suppose it also depends on what planner you use 
%Consider the following two types of equality constraints. 
\begin{defn}
A {\em constant equality} constraint $\langle (z_p), \{\kappa\} \rangle$ (denoted $z_{p} = \kappa$) indicates that parameter $z_p$ has value $\kappa$. %${I} = \{i\}$, ${C} = \{\kappa\}$. 
\end{defn}
\begin{defn}
A {\em pairwise equality} constraint $\langle (z_p, z_{p'}), \{(v, v) \mid v \in {\cal Z}_{p} \cap {\cal Z}_{p'}\} \rangle$ (denoted $z_{p} = z_{p'}$) indicates that parameters $z_p, z_{p'}$ have the same value. %${I} = \{i, j\}$, ${C} = \{(z, z) \mid z \in {\cal Z}_i \cap {\cal Z}_j\}$. 
\end{defn}
%Let $z_{p} = \kappa$ and $z_{p} = z_{p'}$ be shorthand for constant and pairwise equality constraints respectively.
% Not true technically if a variable is discrete

%Many system in practice are factorable into a form that exposes constraints which have simple equality forms. Consider a constraint equality constraint $C(z_v) = [z_v = \kappa]$ which restricts the value of variable $z_v$ to a constant $\kappa$ for a transition. Another common constraint is a pairwise equality constraint $C(x_v, x'_v) = [x_v = x'_v]$ which indicates that state variable $v$ remains unchanged after a transition. Pairwise equality constraints are particular common in system in which transitions only affect a few state variables at a time.

For many high-dimensional systems, the transition relation is {\em sparse}, meaning its transitions only alter a small number of the state variables at a time. Sparse transition systems have transition relations where each clause contains pairwise equality constraints $x_p = x'_p$ for most state variables $p$. Intuitively, most transitions do not change most state variables.

The initial state $\bar{x}^0$ and set of goal states $\bar{X}^*$ can be specified using clauses ${\cal C}_0$ and ${\cal C}_*$ defined solely on state variables. Because $\bar{x}^0$ is a single state, its clause is composed of constant equality constraints.
%\subsection{Equality Constraints}
% The intersection of two constraints is a constraint

%These constraints may be a dominating component of a constraint clause. 
%$\{\langle {I}^a_1, {C}^a_1 \rangle, ..., \langle {I}^a_{\beta_a}, {C}^a_{\beta_a} \rangle\}$. 
%It often is more compact to implicitly assume pairwise equality constraints between $x_v$ and $x'_v$ unless $x'_v$ is mentioned in another constraint within the clause. Under this perspective, a clause can be thought of as constraining the difference between states $x$ and $x'$. 

%Furthermore, some variables may be constrained by a pairwise equality constraint in each clause of the transition relation. This indicates that the variables are {\em static}, or have fixed values throughout a plan. An example static variable is the pose of a object that a robot cannot manipulate. The pose may vary based on the problem instance but is fixed throughout any plan.

%It can be more compact to write ${C}_i(x, u, x')$ as ${C}_i(x, u, \Delta x)$ where $\Delta x$ only includes the variables that change. This can be done by removing $[x_v = x'_v]$ from ${C}_i(x, u, x')$ .
% Discuss parameter lifting here? Probably too early

% Leslie - This is always a sticking point that we have to be very careful about  terminologically:  there are sets of constraints and sets of points in the domain.  Adding to the set of constraints generally decreases the size of the set of points it represents.  A clause is a kind of syntactic thing (a set of (I_j, C_j) constraints);  it represents a set of (x,u,x') tuples.

\subsection{Constraint Satisfaction}

\note{Choose $c$ for clause?}
Planning can be thought of as a combined search over discrete clauses and hybrid parameter values. 
%For each action there is a choice of a discrete transition type %(such as {\it pick} or {\it move}) 
%and of continuous parameters. 
To select a type is to select a clause from the transition relation;  to select its parameters is to select the $\bar{x}, \bar{u}, \bar{x}'$ values.
\begin{defn}
%A finite sequence of clauses $\vec{a} = (a_1, ..., a_k)$ is a {\em plan skeleton}~\citep{lozano2014constraint}. 
A finite sequence of transition components $\vec{a} = (a_1, ..., a_k)$ is a {\em plan skeleton}~\citep{lozano2014constraint}. 
\end{defn}
For example, solutions to pick-and-place problems are sequences of {\it move}, {\it pick}, {\it move-while-holding}, and {\it place} clauses involving the same object.
%Because the transition relation is the disjunction of conjunctive constraint clauses, a transition $(x, u, x')$ is valid if it is contained within the relation specified by at least one clause ${\cal C}_a$. 
%Each conjunctive clause in ${C}(x, u, x')$ can be thought of as an {\em action skeleton}. Because of the disjunction over these action skeletons, a planner can choose which action skeleton to satisfy when performing a transition. Extending this reasoning, solution to a problem can be analyzed to looking at valid values of a {\em plan skeleton}, a finite sequence of action skeletons. 
% Define a conjunctive clause?

%Plan skeletons are useful tools for analyzing a set of solutions to a problem that have the same form. 
\begin{defn}
The {\em plan parameter-space} for a plan skeleton $\vec{a} = (a_1, ..., a_k)$ is an alternating sequence of states and controls $(\bar{x}^0, \bar{u}^1, \bar{x}^1, ..., \bar{u}^k, \bar{x}^k) = \vec{z} \in \bar{\cal X} \times (\bar{\cal U} \times \bar{\cal X})^k = \vec{\cal Z}$. 
\end{defn}
Here, we generically refer to each variable in the plan parameter-space as $z_p$ where now $p \in \{1,..., m + k(m+n)\} = \Theta$.
When applying the constraints for clause $a_t$, plan state $\bar{x}^{t-1}$ is the transition state $\bar{x}$ and likewise $\bar{x}^t$ is $\bar{x}'$. 
%The constraint interpretation of this is the conjunction each of the conjunctive transition clauses. 
%\begin{figure}[ht]
%\centering
%\includegraphics[width=0.25\textwidth]{figures/overlap.pdf}
%\end{figure}
Solutions using this skeleton must satisfy a single clause of all plan-wide constraints
% Should to adapt the indices to refer to the correct variables over the full plan
${\cal C}_{\vec{a}} = {\cal C}_0 \cap {\cal C}_{a_1} \cap ... \cap {\cal C}_{a_k} \cap {\cal C}_*.$
%$${\cal C}_\psi = {\cal C}_0 \cap {\cal C}_* \cap \bigcap_{l = 1}^k {\cal C}_{a_l}.$$
%A state sequence and action sequence for a plan can be merged to form an 
%The set of solutions using this skeleton are then plans in the parameter-space that satisfy the plan-wide clause
%is then simply the intersection of each transition clause 

%$$\bigcap_{i=1}^k (\bar{\cal X} \times \bar{\cal U})^{i-1} \times {\cal T}^{a_i} \times (\bar{\cal U} \times \bar{\cal X})^{k-i}$$.
%$$\{\bar{z} \in \bar{\cal Z} \mid \bar{z}_{I_0} \in {\cal C}_0, \bar{z}_{I_*} \in {\cal C}_*, \forall l \in \{1...k\}\; \bar{z}_{I_{a_l}} \in {\cal C}_{a_l}\}.$$
%$$\{\bar{z} \in \bar{\cal Z} \mid  \forall \langle I, C \rangle \in {\cal C}_\psi\; \bar{z}_{I} \in C\}.$$

%Given a plan skeleton, solutions using that plan skeleton are assignments of the values of  $x_0, ..., x_n$, $u_1, ..., u_n$ which conjunctive satisfy the initial constraints, goal constraints, and action skeleton constraints. 

%$$\bigcup_{i=1}^k\bigcup_{b=1}^{\beta_{a_i}} \langle \{{I}^{a_i}_b, {C}^{a_i}_b \rangle\}.$$
%$$\Xi_{a_1...a_k} = \{\langle {I}^{a_i}_b, {C}^{a_i}_b \rangle \mid i \in \{1...k\}, b \in \{1...\beta_{a_i}\}\}.$$
% Could also do (x, u, x', x u, x', ...) and add additional equality constraints

\begin{defn} \label{defn:satisfiable}
A set of constraints ${\cal C}$ is {\em satisfiable} if there exists parameter values $\vec{z} \in \vec{Z}$ such that $\bar{z} \in \bigcap_{C \in {\cal C}} \ext{C}{\Theta}$.
\end{defn}
\begin{defn} \label{defn:feasible}
A problem ${\cal P}$ is {\em feasible} if there exists a plan skeleton $\vec{a}$ that ${\cal C}_{\vec{a}}$ is satisfiable.
\end{defn}

\begin{figure}[ht]
\centering
%\includegraphics[width=0.48\textwidth]{figures/simple_factor_graph.pdf}
%\caption{Factor graph of where $m=1$, $n=1$, and $\beta_a = 1$.} \label{fig:simple_factor_graph}
%\includegraphics[width=0.48\textwidth]{figures/factor_graph.pdf}
\includegraphics[width=0.48\textwidth]{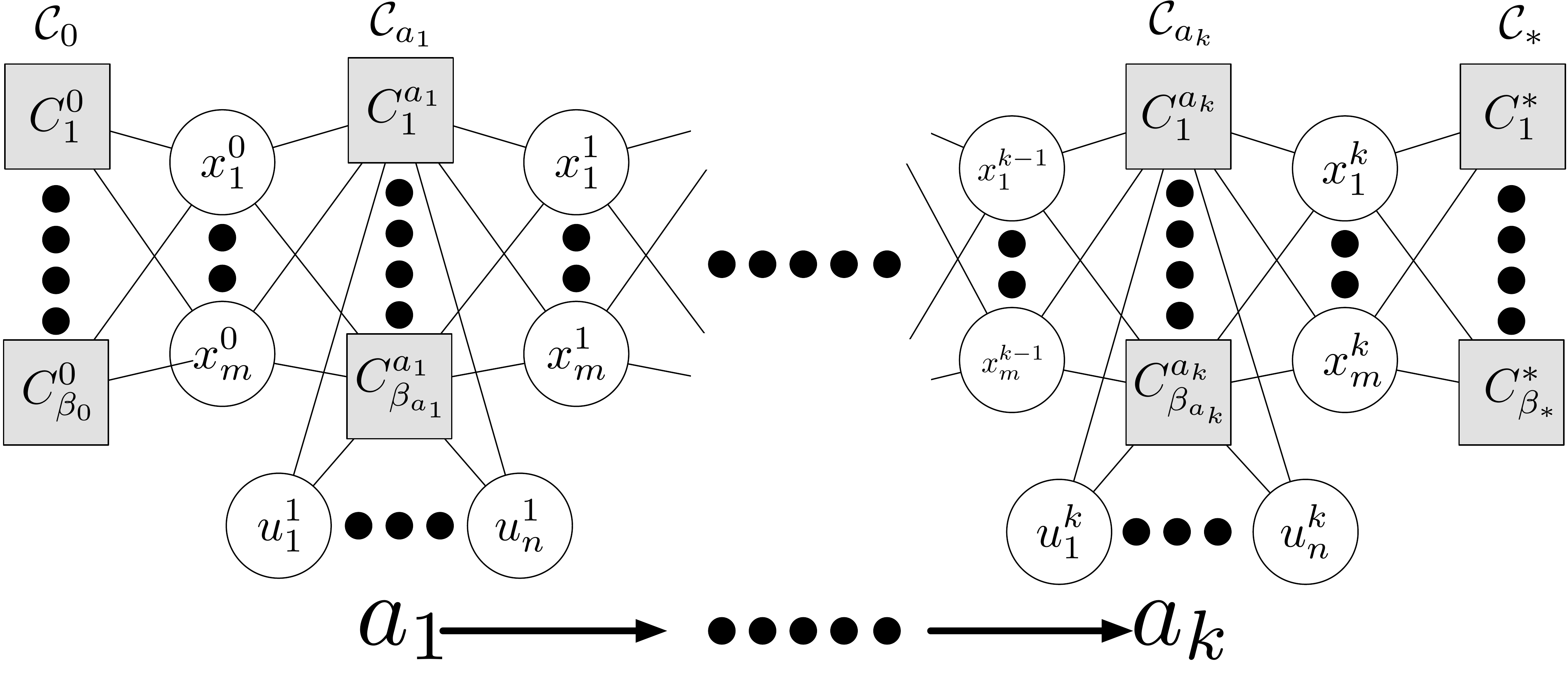}
\caption{A constraint network for a generic plan skeleton $\vec{a} = (a_1, ..., a_k)$ and parameters $\vec{z} = (\bar{x}^0, \bar{u}^1, \bar{x}^1, ..., \bar{u}^k, \bar{x}^k)$.} \label{fig:full_factor_graph}
\end{figure}

%If the transition system is sparse, many state variables will have the same values across time steps, so there will be many constant and pairwise equality constraints.  
%Pre-processing of the clause can be used to eliminate variables that are constrained via equality.
For a given plan skeleton, %the number of plan parameters is fixed, so 
finding satisfying parameter values $\vec{z}$ is a hybrid {\em constraint satisfaction problem}. 
%Of course, identifying a correct plan skeleton itself is a nontrivial problem. 
The joint set of constraints forms a {\em constraint network}, a bipartite graph between constraints and parameters~\citep{dechter1992constraint,lagriffoul2014efficiently}. 
An edge between a constraint node $C = \langle P, R \rangle$ and state or control node $x^{t-1}_p$, $u^t_p$, or $x^t_p$ is defined if and only if $p \in P$. 
%In the constraint satisfaction community, these graphs are a type of {\em constraint network}. 
%Figure~\ref{fig:simple_factor_graph} displays a constraint network for a transition system that only has a trivial factorization, {\it i.e.} $m=1$, $n=1$, and $\beta_a = 1$.  
%A constraint network is specific to a single problem and plan skeleton.
%Leslie - Even for plans of the same length, the choices of action classes in the skeleton determine which variables and constraints are actually involved.
Figure~\ref{fig:full_factor_graph} displays a general constraint network. 
Many transition systems in practice will have constraint networks with many fewer edges because each $P$ contains only a small number of parameters. 
%Thus, the constraint network can expose conditional independence relationships among the plan parameters. 
% Given choices of some parameters, others are independent
% We take advantage of the small number of constraints which involve a variable rather than any conditional independence

% Primal constraint graph - vertices are nodes, constraints edges, undirected graph (markov random field)
% Dual constraint graph - vertices are constraints, hyper-edge shared edge
% Show and example solution and an example sampling network

\note{Modeling mode systems}
\note{$a$ is used for both the clause name and the pose of object $a$}
\note{$p, q$ could be used to emphasize the data type}

%\section{Example Domains}
\section{Example Transition Systems}

%I can always do the Hauser strategy were I mention several domains, but only implement one
%Should I automatically apply equality constraints in these?

%We are interested in developing algorithms to address particular {\em domains}. 
%We are interested in a general algorithmic framework that can be applied in many {\em domains}.
We are interested in a general algorithmic framework that can be applied in many factored transition systems.
% Leslie - Can we say that our algorithms apply to all factored transition systems
%A domain ${\cal D} = \{{\cal P}, {\cal P}',...\}$ is loosely defined as a set of problems that share similar variable, constraint, and transition forms.
%A {\em domain} ${\cal D} = \{{\cal P},...\}$ is a set of problems.
% Domains usually have similar transition functions. In pick and place, want to count problems with a varying number of objects as part of the domain. Can't just change initial and goal
Consider the following two applications and their representation as factored transition systems.
We begin with a motion planning application to illustrate the approach, and then describe a pick-and-place application.
%We give Python code for these examples in the extended version of this paper~\citep{TODO}.

\subsection{Motion Planning}

\note{Constraints must be contained within clauses, so at most can use two of one state variable}
\note{Why even bother putting the variables in them then? Could instead just make two for every action}
\note{Maybe a constraint is just a set that we give meaning when we pass in parameters}
\note{Distinction between parameter values and names}

Many motion planning problems may be defined by a bounded configuration space ${\cal Q} \subset \mathbb{R}^d$ and collision-free configuration space $Q_{\it free} \subseteq {\cal Q}$. 
We will consider planning motions composed of a finite sequence of straight-line trajectories $t$ between waypoints $q, q'$.
Problems are given by an initial configuration $q_0 \in {\cal Q}$ and a goal configuration $q_* \in {\cal Q}$.
Motion planning can be modeled as a transition system with state-space $\bar{\cal X} = {\cal Q}$ and control-space $\bar{\cal U} = {\cal Q}^2$. 
The transition relation ${\cal T} = \{{\cal C}_{Move}\}$ has a single clause
\begin{equation*}
{\cal C}_{Move} = \{\id{Motion}, \id{CFree}\}.
%{\cal C}_{Move} = \{\id{Motion}(x_q, u_t, x_q'), \id{CFree}(u_t)\}.
%{\cal C}_{Move} = \{\id{Motion}(x_R, u_T, x_R'), \id{CFree}(u_T)\}.
\end{equation*}
% $t$ could instead be control effort
% Explain how this will work for general motions under robustness constraints
The transition relation does not exhibit any useful factoring.
A motion constraint $\id{Motion}$ enforces $u_t$ is a straight-line trajectory between $x_q$ and $x_q'$.
\begin{align*}
\id{Motion} = \langle (x_q, u_t, x_q'),  \{&(q, t, q') \mid q, q' \in {\cal Q}^2, \\
&t(\lambda) = \lambda q + (1- \lambda) q'\} \rangle
\end{align*}
A collision-free constraint \id{CFree} ensures all configurations on the trajectory are not in collision.
\begin{equation*}
%\id{CFree} = \{t = (q, q') \mid \forall \lambda \in [0, 1].\; \lambda q + (1- \lambda) q' \in Q_{\it free}\}
\id{CFree} = \langle (u_t), \{t \mid \forall \lambda \in [0, 1].\; t(\lambda) \in Q_{\it free}\} \rangle.
\end{equation*}
The initial clause is ${\cal C}_{0} = \bar{x}^0 = \{x_q = q_0\}$ and the goal clause is ${\cal C}_{*} = \{x_q = q_*\}$.
\note{Zi: is q index or state or control? q \in Q? Does t denote discretized time steps or state? }

\note{Make the robot's variable be named $R$?}

%However, because each degree-of-freedom $j$ is mentioned in the same set of constraints, the factoring does not expose an interesting decomposition of the transition relation.
% More generally say that we can can combine cliques of variables mentioned within the same constraints
The system state could alternatively be described as $\bar{x} =  (x_{j_1}, ..., x_{j_d})$ where $j$ is a single robot degree-of-freedom. 
For simplicity, we combine individual degrees of freedom included within the same constraints into a single variable.
%For example, rather than describe a state variable for each individual robot joint, we consider a single variable for the combined robot configuration.
This is possible because the set of robot joints always occurs together when mentioned as parameters within motion and kinematic constraints.
% This doesn't affect the constraint network but affects completeness/sampling
% We could instead do the cartesian product of joint values, particularly when sampling configurations for motion planning

\begin{figure}[h]
\centering
\includegraphics[width=0.48\textwidth]{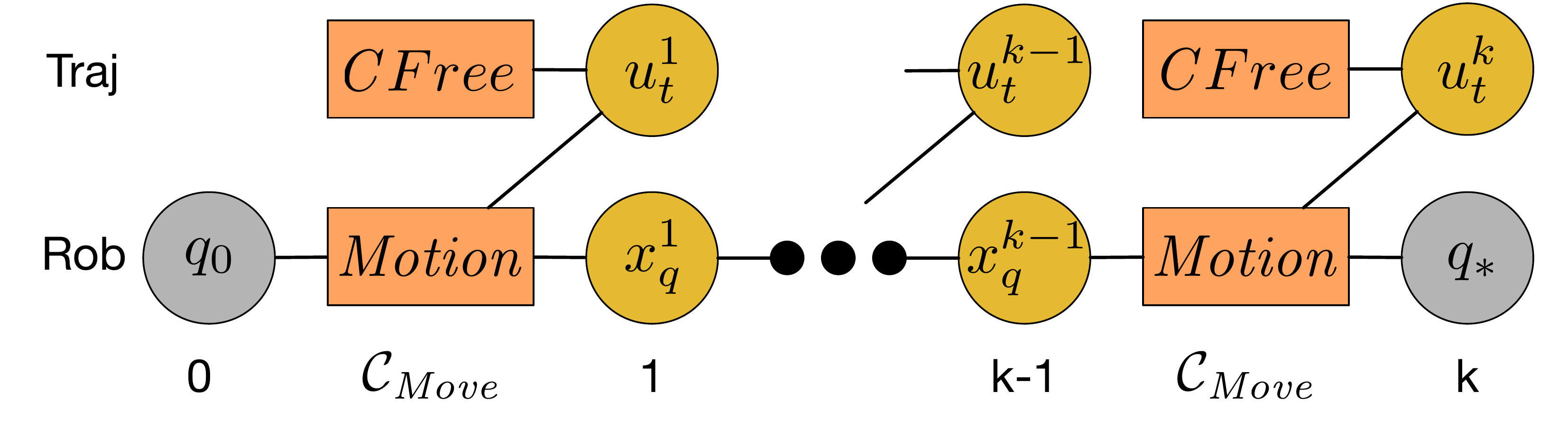}
\caption{Motion planning plan skeleton of length $k$.} \label{fig:motion_solution}
\end{figure}

Figure~\ref{fig:motion_solution} displays a constraint network for a plan skeleton of length $k$. 
Because the transition relation has a single clause, all solutions have this form.
Dark gray circles are parameters, such as the initial and final configurations, that are constrained by constant equality. 
Free parameters are yellow circles.
Constraints are orange rectangles.
% I could split this into the trajectory form

\subsection{Pick-and-Place Planning}
%Maybe consider a problem in which $B$ has to be moved out of the way

%Pick-and-place planning~\citep{garrettIJRR2017} is a subset of task and motion planning.  
% I could either directly use the pick-and-place formation I previously described in FFRob, or I could do something else
% Show that we can reduce sampling a trajectory to sampling several parameter values.
A pick-and-place problem is defined by a single robot with configuration space ${\cal Q} \subset \mathbb{R}^d$, 
a finite set of moveable objects ${\cal O}$, 
%a set of stable placement poses ${\cal S}_{o} \subset \text{SE}(3)$ for each object $o$, 
%and a set of ${\cal G}_{o} \subset \text{SE}(3)$ relative grasp poses for each object $o$.
a set of stable placement poses ${\cal S}_{o} \subset \text{SE}(3)$ for each object $o \in {\cal O}$, 
and a set of grasp poses relative to the end-effector ${\cal G}_o \subset \text{SE}(3)$ for each object $o \in {\cal O}$. 
%Let the combined set of poses for an object $o_i$ be ${\cal P}_i = {\cal S}_{i} \cup {\cal G}_{i}$.
% \subseteq {\cal Q}^l$ composed of most $l-1$ linear movements. %(as opposed to a single linear step)
The robot has a single manipulator that is able to rigidly attach itself to a single object at a time when the end-effector $g$
performs a grasping operation.
%The robot can execute trajectories $t \in T$ that respect joint limits and avoid fixed obstacles. 
As before, the robot can execute straight-line trajectories $t$ between waypoints $q, q'$.
%specified by a sequence of configurations  that respect joint limits and avoid fixed obstacles. 
%We will assume that each trajectory also encodes %the start configuration, end configuration, as well as 
%the grasp of the object that the robot may be holding.

\note{SE(3) but I consider 2d examples?}
% We will assume that $p$ becomes a grasp (relative pose from the end-effector) when it is grasped
Pick-and-place problems can be modeled as a transition system with state-space 
%$\bar{\cal X} = {\cal Q} \times {\cal P}_{1}  \times ... \times {\cal P}_{n}  \times \{\kw{None}, 1, ..., n\}$. 
$\bar{\cal X} = {\cal Q} \times \text{SE}(3)^{|{\cal O}|} \times (\{\kw{None}\} \cup {\cal O})$. 
States are $\bar{x} = (x_q, x_{o_1}, ..., x_{o_{|\cal O|}}, x_h)$.
Let $h \in {\cal O}$ indicate that the robot is holding object $h$ and $h = \kw{None}$ indicate that the robot's gripper is empty. %Each $p_o$ is the pose of object $o$. 
When $h = o$, the pose $x_o$ of object $o$ is given in the end-effector frame. 
Otherwise, $x_o$ is relative to the world frame.
By representing attachment as a change in frame, the pose of an object remains fixed, relative to the gripper, as the robot moves.
% Hauser has explicit mode transformation variables as well which represent grasps. He also updates the real pose though
Controls are pairs $\bar{u} = (u_t, u_g)$ composed of trajectories $u_t$ and boolean gripper force commands $u_g$.
Let $u_g = \kw{True}$ correspond to sustaining a grasp force and $u_g = \kw{False}$ indicate applying no force.
%The control-space is $\bar{\cal U} = \bigcup_{i=1}  T$. 

%because we will only consider intermediate waypoints as control inputs
The transition relation ${\cal T}$ has $1 + 3|{\cal O}|$ clauses because {\it pick}, {\it move-while-holding}, and {\it place} depend on $o$:
\begin{equation*}
{\cal T} = \{{\cal C}_{Move}\} \cup \{{\cal C}_{MoveH}^o, {\cal C}_{Pick}^o, {\cal C}_{\id Place}^o \mid o \in {\cal O}\}.
\end{equation*}
%Let $a \in A$ be shorthand for a constraint $\langle \{a\}, A \rangle$.
%\begin{align*}
%{\cal C}_\id{Move} &= \{{Motion}(x_q, u_t, x_q'), \id{CFree}(u_t), x_h = \kw{None}, \\
%&x_h = x_h'\} \cup \{x_{o'} = x'_{o'}, \id{CFree}_{o'}(t, x_{o'}) \mid o' \in {\cal O}\}
%\end{align*}
%\begin{align*}
%{\cal C}_\id{MoveH}^o &= \{\id{Motion}(x_q, u_t, x_q'), \id{CFreeH}_o(u_t, x_o), x_h = o, \\
%&x_h = x_h'\} \cup \{x_{o'} = x'_{o'}, \id{CFree}_{o'}(t, x_{o'}) \mid o' \in {\cal O}\} \\
%&\cup \{\id{CFreeH}_{o, o'} (x_t, x_o, x_{o'}) \mid o' \in {\cal O}, o \neq o'\} 
%\end{align*}
%\begin{align*}
%{\cal C}_\id{Pick}^o &= \{ \id{Stable}_o (x_o), \id{Grasp}_o (x'_o), x_q = x_q', x_h = \kw{None}, \\
%&x_h' = o, \id{Kin}_o (x_o', x_o, x_q) \} \cup \{x_{o'} = x'_{o'} \mid o' \in {\cal O}, o \neq o'\}
%\end{align*}
%\begin{align*}
%{\cal C}_\id{Place}^o &= \{ \id{Grasp}_o (x_o), \id{Stable}_o (x'_o), x_q = x_q', x_h = o, \\
%&x_h' = \kw{None}, \id{Kin}_o (x_o, x_o', q) \} \cup \{x_{o'} = x'_{o'} \mid o' \in {\cal O}, o \neq o'\}
%\end{align*}
${\cal C}_\id{Move}$ and ${\cal C}_\id{MoveH}^o$ clauses correspond to the robot executing a trajectory $u_t$ while its gripper is empty or holding object $o$:
\begin{align*}
%{\cal C}_\id{Move} =& \{{Motion}(x_q, u_t, x_q'), \id{CFree}(u_t), \\
%&x_h = \kw{None}, x_h = x_h', u_g = \kw{False} \} \;\cup \\
%&\{x_{o'} = x'_{o'}, \id{CFree}(u_t, x_{o'}) \mid o' \in {\cal O}\}
{\cal C}_\id{Move} =& \{\id{Motion}, \id{CFree}, x_h = \kw{None}, x_h = x_h', \\
&u_g = \kw{False} \} \;\cup \{x_{o'} = x'_{o'}, \id{CFree}_{o'} \mid o' \in {\cal O}\}
\end{align*}
\begin{align*}
%{\cal C}_\id{MoveH}^o =& \{\id{Motion}(x_q, u_t, x_q'), \id{CFreeH}(u_t, x_o),  \\
%&x_h = o, x_h = x_h', u_g = \kw{True}\} \;\cup \\
%%\cup \{x_{o'} = x'_{o'}, \id{CFree}(u_t, x_{o'}) \mid o' \in {\cal O}\} \\
%& \{\id{CFreeH}(u_t, x_o, x_{o'}) \mid o' \in {\cal O}, o \neq o'\} 
{\cal C}_\id{MoveH}^o =& \{\id{Motion}, \id{CFreeH}_o, x_h = o, x_h = x_h', \\
& u_g = \kw{True}\} \;\cup \{\id{CFreeH}_{o, o'} \mid o' \in {\cal O}, o \neq o'\}.
\end{align*}
%$\id{Motion}$ is the set of legal start configurations, trajectories, and end configurations.
%$\id{CFree}$ is the set of collision-free trajectories with respect to the environment.
$\id{CFree}_o$ is a constraint containing robot trajectories $u_t$ and object $o$ poses $x_o$ that are not in collision with each other.
$\id{CFreeH}_{o}$ is a constraint composed of robot trajectories $u_t$ and object $o$ grasps $x_o$ relative to the end-effector that are not in collision with the environment.
$\id{CFreeH}_{o, o'}$ is a constraint containing robot trajectories $u_t$, object $o$ grasps $x_o$ relative to the end-effector, and object $o'$ poses $x_o'$ that are not in collision with each other.

${\cal C}_\id{Pick}^o$ and ${\cal C}_\id{Place}^o$ clauses correspond to instantaneous changes in what the gripper is holding:
\begin{align*}
%{\cal C}_\id{Pick}^o =& \{ \id{Stable}(x_o), \id{Grasp}(x'_o), \id{Kin}(x_o', x_o, x_q), \\
%&x_q = x_q', x_h = \kw{None}, x_h' = o \} \;\cup \\
%&\{x_{o'} = x'_{o'} \mid o' \in {\cal O}, o \neq o'\}
{\cal C}_\id{Pick}^o =& \{ \id{Stable}_o, \id{Grasp}_o', \id{Kin}_o, x_q = x_q', x_h = \kw{None}, \\
&x_h' = o \} \;\cup \{x_{o'} = x'_{o'} \mid o' \in {\cal O}, o \neq o'\}
\end{align*}
\begin{align*}
{\cal C}_\id{Place}^o =& \{ \id{Grasp}_o, \id{Stable}_o', \id{Kin}'_o, x_q = x_q', x_h = o, \\
&x_h' = \kw{None}\}  \;\cup \{x_{o'} = x'_{o'} \mid o' \in {\cal O}, o \neq o'\}.
\end{align*}
As a result, they do not involve any control variables.
%Let ${Grasp}_o = {\cal G}_o$ and ${Stable}_o = {\cal S}_o$.
$\id{Grasp}_o = \langle (x_o), {\cal G}_o \rangle$ is a constraint that $x_o$ is a grasp transform from the object frame to the end-effector frame.
Let $\id{Grasp}'_o = \langle (x_o'), {\cal G}_o \rangle$ be the same constraint but on $x_o'$.
Similarly, $\id{Stable}_o = \langle (x_o), {\cal S}_o \rangle$ is a constraint that $x_o$ is a stable placement, and $\id{Stable}'_o = \langle (x_o'), {\cal S}_o \rangle$.
Finally, $\id{Kin}_o$ is a constraint composed of kinematic solutions involving object $o$ for a grasp $g$, pose $p$, and robot configuration $q$:
\begin{equation*}
\id{Kin}_o = \langle (x_o', x_o, x_q),  \{(g, p, q) \mid \proc{kin}(q) = pg^{-1}\} \rangle.
\end{equation*}
$\id{Kin}_o'$ is the equivalent constraint but on state variables $(x_o, x_o', x_q)$, where $x_o$ and $x_o'$ are swapped.
Because $\id{Kin}_o$ and $\id{Kin}_o'$ refer to the same relation and only involve different parameters, we will just refer to $\id{Kin}_o$.
% world_from_manip * manip_from_stable =  world_from_stable
% world_from_stable * manip_from_stable ^ {-1} =  
%${\cal U} \subset {\cal Q}^{l-2}$ because we will only consider intermediate waypoints as control inputs
% Could also think about trajectories that separate base and arm movement
\note{Ensure I keep the correct grasp, pose \id{Kin} order}

%Variables not completely constrained by equality.
Pick-and-place transition systems are substantially factorable.
Each constraint involves at most 3 variables.
Additionally, in each clause, many variables are entirely constrained by equality.
For ${\cal C}_\id{Move}, {\cal C}_\id{MoveH}^o$ clauses, only half of the variables are not constrained by equality:
% $3 + |O|$
% $2*(2+|O|) + 2$ = $6+2*|O|)$
%\begin{equation*}
$\{x_q, u_t, x_q'\} \cup \{x_{o'} \mid o' \in {\cal O}\}.$
%\end{equation*}
For ${\cal C}_\id{Pick}^o, {\cal C}_\id{Place}^o$ clauses, only $\{x_o, x_o', x_q\}$ variables are not constrained by equality.
\note{Should I use upper case letters for objects to indicate that they are names?}

\begin{figure}[h]
\centering
\includegraphics[width=0.49\textwidth]{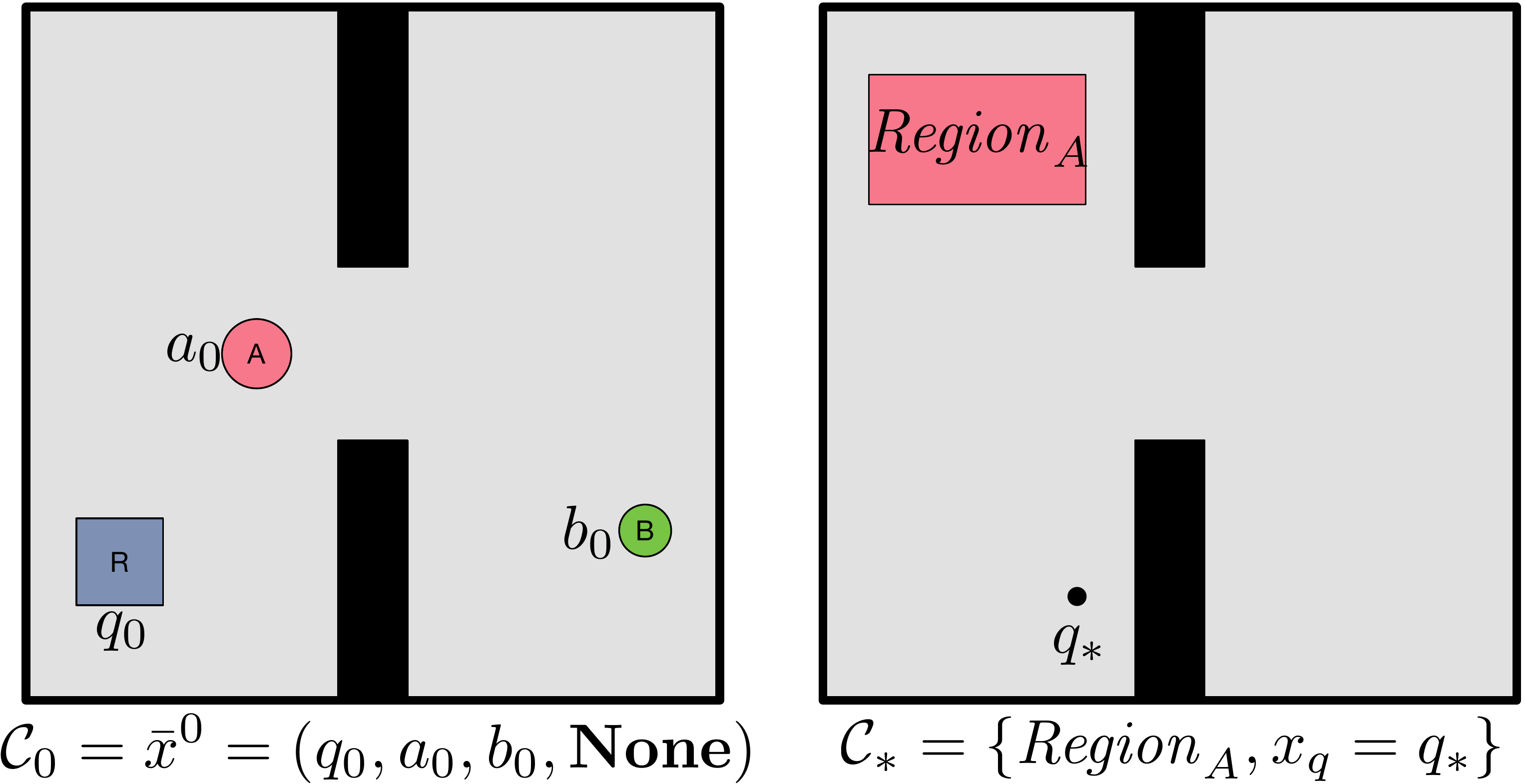}
\caption{The initial state (left) and goal constraints (right) for a pick-and-place problem involving a square robot $R$ and movable circle objects $A$ and $B$.} \label{fig:pp_initial_goal}
\end{figure}
\note{Change $C_0$ in this figure}

We will use the pick-and-place problem shown in figure~\ref{fig:pp_initial_goal} with two movable objects $A, B$ as a running example.
The initial state $\bar{x}^0 = (q_0, a_0, b_0, \kw{None})$ is fully specified using equality constraints ${\cal C}_0 =  \{\bar{x}_q = q_0, \bar{x}_A = a_0, \bar{x}_B = b_0, \bar{x}_h = \kw{None}\}$. 
We assume that $a_0$ and $b_0$ are stable poses: $a_0 \in {\cal S}_A$ and $b_0 \in {\cal S}_B$.
The goal states $\bar{X}^*$ are given as constraints ${\cal C}_* = \{\id{Region}, x_q = q_*\}$ 
where the region constraint $\id{Region}_A = \langle (x_A), {\cal R}_A \rangle$, for ${\cal R}_A \subseteq {\cal S}_A$  is a subset of the stable placements for object $A$.

\begin{figure}[h]
\centering
\includegraphics[width=0.49\textwidth]{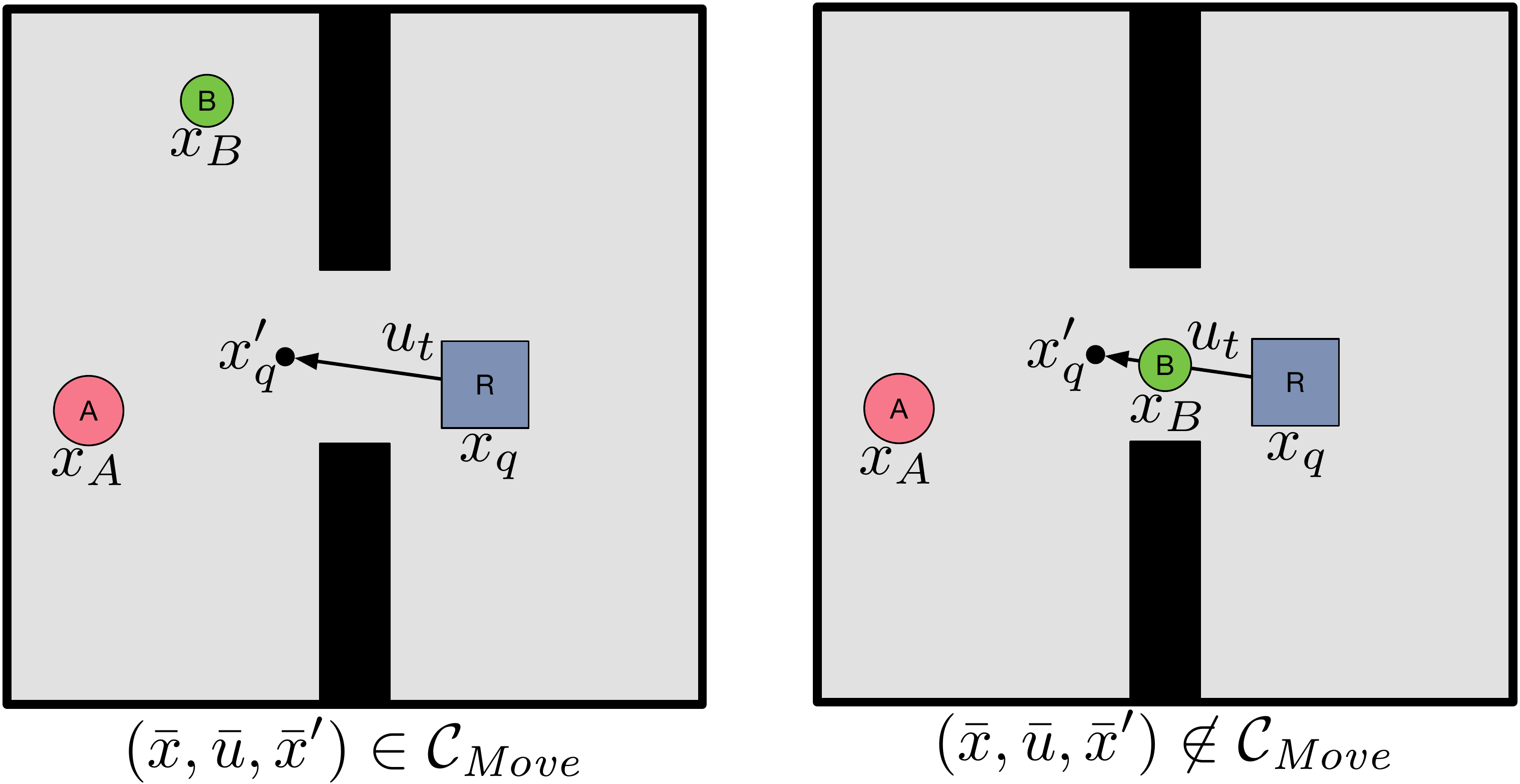}
\caption{A valid transition (left) and an invalid transition (right) for the same control trajectory $u_t$. The right transition is invalid because it violates the $\id{CFree}_B$ collision-free constraint.} \label{fig:pp_transition}
\end{figure}

A useful consequence of factoring is that the same control values $\bar{u}$ can be considered in many transitions.
Consider the two candidate transitions in figure~\ref{fig:pp_transition}, both using the same control trajectory $u_t$.
The application of $u_t$ in the left figure results in a valid transition for clause $\id{Move}$.
In fact, for a majority of the combinations of placements of $x_A$, $x_B$, $u_t$ is a valid transition. 
Thus, for a single value of $u_t$, we are implicitly representing many possible transitions.
The right figure, however, shows an instance in which this $u_t$ does not correspond to a legal transition as it would result in a collision with $B$, thus violating $\id{CFree}_B$.
%Factoring exposes the fact a single control trajectory $u_t$ may be legal in many transitions depending on the poses of $A$ and $B$.
% Could also make a note about states, but we really don't talk about state legality, just transition legality

\begin{figure*}[ht]
\centering
\includegraphics[width=0.99\textwidth]{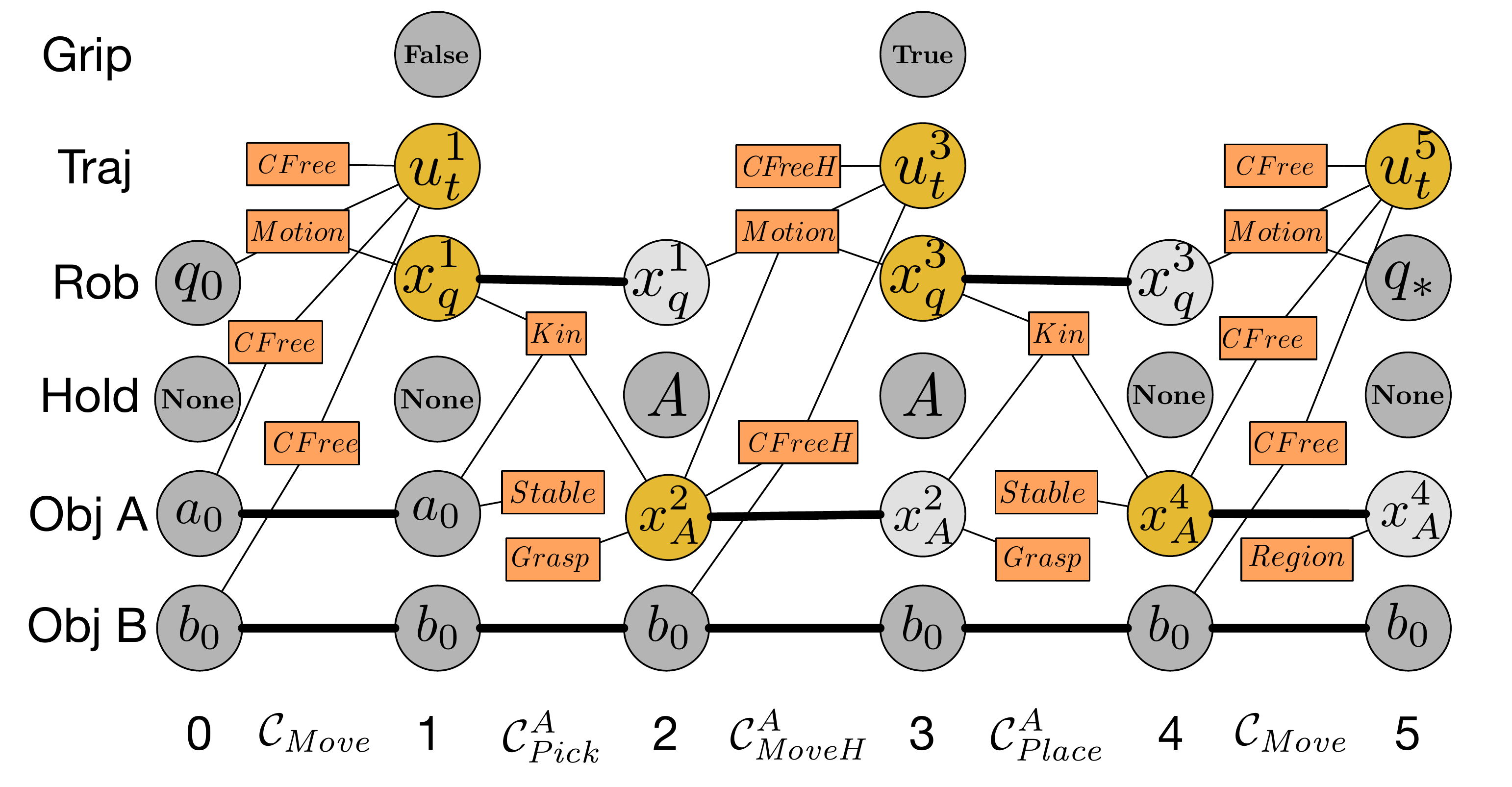}
\caption{Pick-and-place constraint network for a plan skeleton $\vec{a} = (\id{Move}, \id{Pick}^A, \id{MoveH}^A, \id{Place}^A, \id{Move})$.} \label{fig:pp_network}
\end{figure*}

Figure~\ref{fig:pp_network} displays the constraint network for a plan skeleton 
%\begin{equation*}
$\vec{a} = (\id{Move}, \id{Pick}^A, \id{MoveH}^A, \id{Place}^A, \id{Move})$
%\end{equation*}
that grabs $A$, places $A$ in $\id{Region}_A$, and moves the robot to $q_*$.
%and then $B$.
%In general, minimal solutions to pick-and-place problems are alternating sequences of {\it move}, {\it pick}, {\it move-holding}, and {\it place} transitions. 
Thick edges indicate pairwise equality constraints. 
Light gray parameters are transitively fixed by pairwise equality.
We will omit the constraint subscripts for simplicity.
% to either a constant or free parameter. 
%Figure~\ref{fig:free_parameters} shows the same constraint network, identifying just the free parameters. 
Despite having $29$ total parameters, only $7$ are free parameters. 
This highlights the strong impact of equality constraints on the dimensionality of the plan parameter-space.

%\begin{figure*}[ht]
%\centering
%\begin{tabular}{c}
%%\includegraphics[width=0.9\textwidth]{figures/tmp_factor.pdf} \\
%\includegraphics[width=0.9\textwidth]{figures/small_pp_network.pdf} \\
%Complete constraint network\\
%\hline
%\rule{0pt}{1ex}    \\
%%\includegraphics[width=0.9\textwidth]{figures/tmp_factor_3.pdf} \\
%\includegraphics[width=0.9\textwidth]{figures/small_pp_free.pdf} \\
%Network with free parameters only\\
%\hline
%\rule{0pt}{1ex}    \\
%%\includegraphics[width=0.9\textwidth]{figures/bayesian_3.pdf}\\
%\includegraphics[width=0.9\textwidth]{figures/small_pp_dag.pdf}\\
%Sampling graph
%\end{tabular}
%\caption{A constraint networks for a pick-and-place problem with
%  8 transitions. } \label{fig:simple_factor_graph} 
%%\includegraphics[width=0.98\textwidth]{figures/tmp_factor_2.pdf}
%%\caption{TMP factor graph.} \label{fig:simple_factor_graph}
%\end{figure*}

% Can move the old figures to the appendix
%\begin{figure}[ht]
%\centering
%\includegraphics[width=0.49\textwidth]{figures/small_pp_free.pdf}
%\caption{Network with free parameters only.} \label{fig:free_parameters}
%\end{figure}

% Figure~\ref{fig:simple_factor_graph_2} displays the same factor graph with only free parameters. 
% \begin{figure*}[ht]
% \centering
% \caption{The constraint network in figure~\ref{fig:simple_factor_graph} that only includes free parameters.} \label{fig:simple_factor_graph_2}
% \end{figure*}

\section{Sampling-Based Planning} \label{sec:analysis}

%So far, we have allowed each constraint ${C}$ to be an arbitrary set of values. However, to plan with these constraints, we need a finite characterization of their values. This is easy when ${C}$ is finite, as commonplace in the constraint satisfaction community. However, we are interested in problems where ${C}$ may instead be uncountably infinite. Such sets are often difficult to both describe and reason with.
%Such sets require knowledge of a special purpose representation, such as a semi-algebraic decomposition, in order to compactly describe.

Constraints involving continuous variables are generally characterize uncountably infinite sets, which are often difficult to reason with explicitly.
Instead, each constraint can be described using a blackbox, implicit {\em test}. % rather than an an explicitly characterization. 
%Let $z_{I} = (z_{i_1}, ..., z_{i_k})$ where $\{i_1, ..., i_k\} = {I}$.
A test for constraint $C = \langle {P}, {R} \rangle$ is a boolean-valued function $t_C: \bar{\cal Z}_{P} \to \{0, 1\}$ where $t_C(\bar{z}_{P}) = [\bar{z}_{P} \in {R}]$. Implicit representations are used in sampling-based motion planning, where they replace explicit representations of complicated robot and environment geometries with collision-checking procedures.  

In order to use tests, we need to produce potentially satisfying values for $\bar{z}_P = (z_{p_1}, ..., z_{p_k})$ by sampling ${\cal Z}_{p_1}, ..., {\cal Z}_{p_k}$. 
Thus, we still require an explicit representation for ${\cal X}_1, ..., {\cal X}_m$ and ${\cal U}_1, ..., {\cal U}_n$; however, these are typically less difficult to characterize. 
We will assume ${\cal X}_1, ..., {\cal X}_m$, ${\cal U}_1, ..., {\cal U}_n$ are each bounded manifolds.
%of a topological space. % with boundary. 
This strategy of sampling variable domains and testing constraints is the basis of {\em sampling-based planning}~\citep{Kavraki96}. These methods draw values from ${\cal X}_1, ..., {\cal X}_m$ and ${\cal U}_1, ..., {\cal U}_n$ using deterministic or random {\em samplers} for each space and test which combinations of sampled values satisfy required constraints. 
% A sampler is an object $\psi$ which has the following function \proc{sample}($\phi$)
%\note{Formally define sampler. Is a sampler a set? Is it a procedure with state?}
%A sampler produces a deterministic or random sequence of values from its sample space.
%A sampling-based method terminates when has produced a sequence of states and controls that satisfy the initial, transition, and goal constraints. 
%This general approach can also be seen as simultaneously sampling parameter values for each possible plan skeleton at once. 

Sampling-based techniques %avoid directly reasoning about the intersection of constraints, they 
are usually not complete over all problem instances. 
First, they cannot generally identify and terminate on infeasible instances.  
Second, they are often unable to find solutions to instances that require identifying values from a set that has zero measure in the space from which samples are being drawn.
% has very small or even 
In motion planning, these are problems in which all paths have zero {\em clearance} (sometimes called {\em path-goodness}), the infimum over distances from the path to obstacles~\citep{kavraki1998analysis,Kavraki98probabilisticroadmaps}.
% Clearance is weaker than epsilon goodness
Thus, sampling-based motion planning algorithms are only theoretically analyzed over the set of problems admitting a path with positive clearance. 
Conditions similar to positive clearance include an open configuration space~\citep{Laumond:1998:RMP:521883}, positive $\epsilon$-{\em goodness}
% Epsilon goodness relates to the portion of the configuration space that is visible
~\citep{kavraki1995randomized,Barraquand97,Kavraki98probabilisticroadmaps} and {\em expansiveness}~\citep{hsu1997path,Kavraki98probabilisticroadmaps}.
% Expansiveness is stronger than $\epsilon$-goodness
% A random sampling scheme for path planning \cite{Barraquand97}
% Expansiveness 
% Configuration space is open. Manifold story is like this
%However, under typical nonzero clearance conditions~\citep{kavraki1998analysis} this transition system can be reduced to robustly feasible transition system restricted to straight-line trajectories.
% (this is referred to as the ``narrow passage'' problem in motion planning). 
% Finding Narrow Passages with Probabilistic Roadmaps:  The Small Step Retraction Method - http://ai.stanford.edu/~latombe/papers/iros05/paper.pdf
% Narrow Passage Sampling for Probabilistic Roadmap Planning - https://users.cs.duke.edu/~reif/paper/sunz/bridge/bridge.pub.pdf
%and therefore are concerned only with {\em semi-completeness}. 
% Analysis of probabilistic roadmaps for path planning
% https://personalrobotics.ri.cmu.edu/files/courses/papers/Kavraki98-prm.pdf

The ideas of positive clearance~\citep{van2009path,garrettIJRR2017}, positive $\epsilon$-goodness~\citep{berenson2010probabilistically,barry2013manipulation}, and expansiveness~\citep{HauserLatombe, HauserIJRR11} have been extended to several manipulation planning contexts.
In manipulation planning, zero-clearance related properties can take on additional forms.
For instance, a goal constraint that two movable objects are placed within a tight region may only admit pairs of goal poses lying on a submanifold of the full set of stable placements~\citep{van2009path,garrettIJRR2017}. 
At the same time, kinematic constraints resulting from a pick operation define a set lying on a submanifold of the robot's configuration space~\citep{berenson2010probabilistically,garrettIJRR2017}.
Thus, these clearance-related properties typically identify a set of anticipated submanifolds, such as a set of inverse kinematic solutions, that can be directly sampled.

%produce submanifolds of the robot's configuration space.
% where zero-clearance related phenomenon arise from  in other forms. 

% Kinematics, placement
% Intersection of manifolds lower dimensional
% Kinematic solution that induces a zero clearance path planning problem

% Random geometric graphs for Vega-Brown?
% Jenny uses open in 6.1 in her paper
% 6.2 deals with mode transitions

More generally, sampling-based algorithms are typically only complete over {\em robustly feasible} problems~\citep{karaman2011sampling}.
%problems for which there exists a plan-skeleton with a set of solutions which has nonzero measure in its plan parameter-space. 
When directly applied to factored transitions, a problem is robustly feasible if there exists a plan skeleton $\vec{a}$ such that 
%$\mu( \bar{\cal Z}_{\vec{a}} \cap \bigcap_{C \in {\cal C}_{\vec{a}}} \widehat{C}) > 0$
%$\mu(\bigcap_{C \in {\cal C}_{\vec{a}}} \ext{C}{\Theta} ) > 0$ where $\mu$ is a product measure on the 
$\bigcap_{C \in {\cal C}_{\vec{a}}} \ext{C}{\Theta}$ contains an open set
%sample space $\bar{\cal Z}$. 
in plan-parameter-space $\bar{\cal Z}$. 
% formed from measures for the codomain of each ${\cal Z}_p$.

\note{I could use ${\cal P}_{\vec{a}}$ and $\bar{\cal Z}_{\vec{a}}$}

\subsection{Dimensionality-reducing constraints}

%{\bf Dimensionality-reducing constraints}
% The dimensionality of the solution cannot drop more than the dropped dimensionality of each constraint
% We seem to decompose constraints into the smallest possible such that all the dimensionality reduction is contained within them
%A constraint is embedded in ${C} \subseteq {\cal Z}_{i_1} \times ... \times {\cal Z}_{i_k}$. 
%The constraint space is manifold because the Cartesian product of two \textwidtho manifolds is itself a manifold.
% However, the constraint may be a submanifold of the constraint space because the dimensionally is reduced
% Hauser uses coparameter as the fixed values within the manifold 
%We know a manifold that contains the set of solutions
Some problems of interest involve individual constraints that only admit a set of values on a lower-dimensional subset of their parameter-spaces. 
%A {\em dimensionality-reducing constraint} $C$ is one in which $\mu(C) = 0$ for all problems in the domain. 
A {\em dimensionality-reducing constraint} $C = \langle P, R \rangle$ is one in which $R \subseteq \bar{\cal Z}_P$ does not contain an open set. %. for all problems in the domain. 
Consider the ${\id Stable}_o$ constraint. %  subset of poses that are stable placements ${\cal S}$. 
The set of satisfying values lies on a 3-dimensional manifold.
% Is it okay to call this a single manifold even if it's made up of complete disconnected pieces?
% because its free parameters are a discrete index for the surface involved, an $(x, y)$ position, and a $\theta$ rotation about the $z$ axis.
By our current definition, all plans involving this constraint are not robustly feasible. 
%We could instead attempt to directly sample pose parameters from ${\id Stable}$, 
%% Sample and define robust feasibility with respect to
%however they are often in ${\id Kin}$, another dimensionality-reducing constraint.
%The set of end-effector transforms and manipulator
%configurations that admit a kinematic solution inherently has low-dimensionality. 
%%More formally, a {\em dimensionality-reducing constraint} ${C} \subseteq {M}$ is a subset of a submanifold ${M}$ embedded within its product manifold ${\cal Z}_{I}$ where $\dim{M} < \dim{\cal Z}_{I}$. Typically ${M}$ is the minimal dimension submanifold for which ${C}$ has nonzero measure.
When a problem involves dimensionality-reducing constraints, we have no choice but to sample at their intersection. 
This, in general, requires an explicit characterization of their intersection, which we may not have. %However, the number
%It may be difficult to explicit produce this representation. Moreover, 
Moreover, the number of dimensionality-reducing constraint combinations can be unbounded as plan skeletons may be arbitrarily long.
% which can be intractable to describe
However, in some cases, we can produce this intersection automatically using explicit characterizations for only a few spaces.
%individual constraints. % or small subsets
%In order to effectively solve problems with dimensionality-reducing constraints using sampling-based techniques, we need to somehow incorporate all the dimensionality-reducing constraints into our construction of the sample space.
%We will do this by composing spaces that are conditioned on values of previously chosen parameter values.
%defined relative to values of other parameters. 
%When this process is possible, it will allow us to identify a space that is the intersection of these constraints without ever requiring an explicit form of their intersection. 

\note{Do I want to say that we only consider constraints that are low-dimensional for the full domain?}
% I guess I kind of assume the constraint dimensionality is fixed across the domain
% One could have meta-parameters that adjust it
\note{I currently talk about TAMP by only considering poses}

\begin{figure}[h]
\centering
\includegraphics[width=0.49\textwidth]{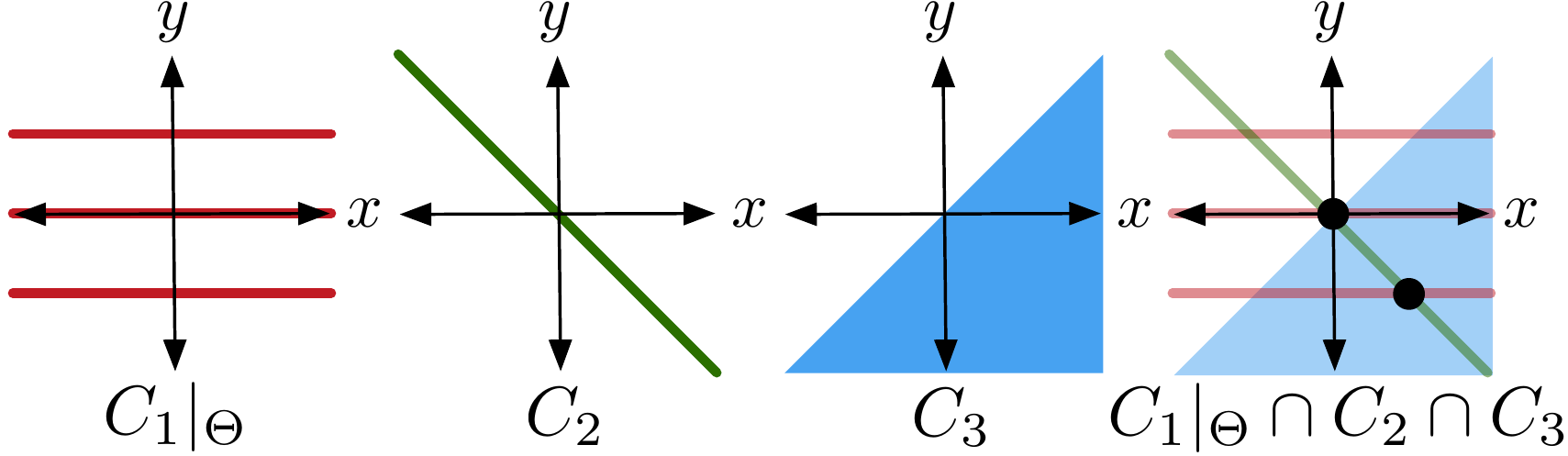}
\caption{Intersection of three constraints, two of them dimensionality-reducing.} \label{fig:2d_example}
\end{figure}
%Maybe make this example something real. Suppose we have an action to aim at a target or something where $a$ is the target and $b$ is the amount of thrust?

We motivate these ideas with an example visualized in figure~\ref{fig:2d_example}.
Consider a plan skeleton $\vec{a}$ with parameters $\Theta = (z_x, z_y)$ where ${\cal Z}_x = {\cal Z}_y = (-2, +2)$ and constraints ${\cal C}_{\vec{a}} = \{C_1, C_2, C_3\}$ where
\begin{align*}
C_1 &= \langle (y), \{-1, 0, 1\}\rangle \\
C_2 &= \langle (x, y), \{(x, y) \mid x + y = 0\}\rangle \\
C_3 &= \langle (x, y), \{(x, y) \mid x - y \geq 0\}\rangle.
\end{align*}
%\begin{itemize}
%\item $C_1 = \langle (y), \{-1, 0, 1\}\rangle$ %, $\widehat{C}_1 = \{(x, -1), (x, 0), (x, +1) \mid x \in [-2, +2]\}$
%\item $C_2 = \langle (x, y), \{(x, y) \mid x + y = 0\}\rangle$ %, $\widehat{C}_2 = \{(x, y) \mid x + y = 0\}$
%%\item $C_2 = \langle (x, y), \{(x, y) \mid |x - y| = 1\}\rangle$
%\item $C_3 = \langle (x), \{x \mid x \geq 0\}\rangle$ %, $\widehat{C}_3 = \{(x, y) \mid y \geq 0\}$
%\end{itemize}
%(a line on a plane), 
%(discrete set on a line),
The set of solutions $\ext{C_1}{\Theta} \cap C_2 \cap C_3  = \{(1, -1), (0, 0)\}$ is 0-dimensional while the parameter-space is 2-dimensional. This is because $C_1$ and $C_2$ are both dimensionality-reducing constraints.
 %that lose one degree-of-freedom. 
 A uniform sampling strategy where $X, Y \sim \text{Uniform}(-2, +2)$ has zero probability of producing a solution.
%The following random sampling approaches have zero probability of producing a solution:
%\begin{itemize}
%\item $(X, Y)$ where $X, Y \sim \text{Uniform}(-2, +2)$
%\item $(X, Y)$  where $X \sim \text{Uniform}(-2, +2)$, \\$Y \sim \text{Uniform}(\{-1, 0, +1\})$
%\item $(T, -T)$ where $T \sim \text{Uniform}(-2\sqrt{2}, +2\sqrt{2})$
%\end{itemize}
%
%The other 3 combinations of ways to sample $x$ from ${\cal Z}_x$ or $C_2$ and $y$ from ${\cal Z}_y$, $C_1$, or $C_2$ all also have zero probability of producing a solution. 

To solve this problem using a sampling-based approach, we must sample from $\ext{C_1}{\Theta} \cap C_2$. 
%$\widehat{C}_1 \cap \widehat{C}_2 = \{(1, -1), (0, 0), (-1, 1)\}$
Suppose we are unable to analytically compute $\ext{C_1}{\Theta} \cap C_2$, but we do have explicit representations of $C_1$ and $C_2$ independently. In particular, suppose we know $C_2$ conditioned on values of $y$, 
%$R_2(y) = \{(-y, y)\}$.
$C_2(y) = \langle (x), \{-y\}\rangle$. 
Now, we can characterize $\ext{C_1}{\Theta} \cap C_2 = \{(x, y) \mid y \in R_1, x \in R_2(y)\} = \{(1, -1), (0, 0), (-1, 1)\}$. With respect to a counting measure on this discrete space, $\ext{C_1}{\Theta} \cap C_2 \cap C_3$ has positive measure. This not only gives a representation for the intersection but also suggests the following way to sample the intersection: $Y \sim \text{Uniform}(\{-1, 0, +1\})$,  $X = -Y$, and reject $(X, Y)$ that does not satisfy $C_3$.
%Only sampled pairs that additionally satisfy $C_3$ are solutions though.
% Conditional sampler?
This strategy is not effective for all combinations of dimensionality-reducing constraints. Suppose that instead $C_1 = \langle (x, y), \{(x, y) \mid x - y = 1\}\rangle$. Because both constraints involve $x$ and $y$, we are unable to condition on the value of one parameter to sample the other. % NOTE - remove sample?
%Intuitively, this strategy is effective when there is an order in which the parameters can be chosen conditional on previous parameters.
%conditionally sampled from the low-dimensional constraints.
\note{It would be nice if we sampled $X$ as well...}

%We can generalize to to sample for pose, then inverse kinematic config, then motion plan. 
%Intuitively, this works when there is a sequence in which the variables can be sampled. The sampling process is able to take into account the loss of dimensionality.

%%%%%%%%%%%%%%%%%%%%%%%%%%%%%%

\subsection{Intersection of Manifolds}

\note{The denseness argument doesn't depend on the measure stuff. I don't have to worry about saying a densely sample the coordinate space. Instead, I can just say that I hit the subset of configurations that satisfy this property}

In this section, we develop the topological tools to generalize the previous example.
Our objective is to show that by making assumptions on each dimensionality reducing constraint individually, we can understand the space formed by the intersection of many dimensionality constraints. 
First, we overview several topological ideas that we will use~\citep{tu2010manifolds}.

\subsubsection{Topological Tools:}

A $d$-dimensional {\em manifold} $M$ is a topological space that is locally homeomorphic to $d$-dimensional Euclidean coordinate space.  
Let $d = \dim{M}$ be the dimension of the coordinate space of $M$.
An {\em atlas} for an $d$-dimensional manifold $M \subseteq \mathbb{R}^m$ is a set of charts $\{(U_\alpha, \varphi_\alpha), ...\}$ such that $\bigcup_\alpha U_\alpha = M$. Each {\em chart} $(U_\alpha, \varphi_\alpha)$ is given by an open set  $U_\alpha \subseteq M$ and a {\em homeomorphism}, a continuous bijection with a continuous inverse, $\varphi: U_\alpha \to \mathbb{R}^d$. 
Let $N$ be a regular submanifold of ambient manifold $M$.
Then, $\codim{N} = \dim{M} - \dim{N}$ is the {\em codimension} of $N$.
% bounded
%worrying about boundaries or crossing points 

Define $\tgnt{x}{M}$ to be the {\em tangent space} of manifold $M$ at $x \in M$, which intuitively contains the set of directions in which one can pass through $M$ at $x$.
A smooth map of manifolds $f: M \to N$ is {\em submersion} at $x \in M$ if its {\em differential}, a linear map between tangent spaces, $\D f_x: \tgnt{x}{M} \to \tgnt{f(x)}{N}$ is surjective. 
% Pushforward (differential)
When $\dim{M} \geq \dim{N}$, this is equivalent to the Jacobian matrix of $f$ at $x$ having maximal rank equal to $\dim{N}$.
When $f$ is a submersion, by the {\em preimage theorem} (also called the implicit function theorem and the regular level set theorem), the preimage $f^{-1}(y)$ for any $y \subseteq N$ is a submanifold of $M$ of dimension ${\dim{M} - \dim{N}}$.
% Implicit function theorem, submersion theorem, preimage theorem, regular level set
% Regular value, smooth map, tangent map is surjective - map is trivially smooth because it is linear 
% Critical point, singular point
Similarly, by the {\em local normal submersion theorem}, there exists coordinate charts $\varphi$ and $\varphi'$ local to $x \in M$ and $f(x) \in N$ such that $f$ is a projection in coordinate space.
% This restriction prunes degenerate {\em critical points} from the conditional constraint manifold. 
Thus, the first $\dim{N}$ coordinates of $\varphi(x)$ and $\varphi(f(x))$ are the same and $\varphi(x)$ contains an extra $\codim{N}$ coordinates.
%the canonical submersion. 
%and the other $\dim{M} - \dim\proj_{I}(\interior(M))$ depend potentially on 
% https://math.stackexchange.com/questions/2340180/global-analog-of-local-submersion-theorem
% Should I put this before the preimage theorem?

For manifolds $N_1$ and $N_2$, consider projections $\proj_1: (N_1 \times N_2) \to N_1$ and $\proj_2: (N_1 \times N_2) \to N_2$.
For all $\bar{y} = (y_1, y_2) \in N_1 \times N_2$, the map
$$(\D\proj_1, \D\proj_2): \tgnt{\bar{y}}{N_1 \times N_2} \to \tgnt{y_1}{N_1} \times \tgnt{y_2}{N_2}$$
is an isomorphism.
Thus, the tangent space of a product manifold is isomorphic to the Cartesian product of the tangent spaces of its components.

Let $N_1, N_2$ both be submanifolds of the same ambient manifold $M$.
An intersection is {\em transverse} if $\forall x \in (N_1 \cap N_2).\; \tgnt{x}{N_1} + \tgnt{x}{N_2} = \tgnt{x}{M}$ where
$$\tgnt{x}{N_1} + \tgnt{x}{N_2} = \{v + w \mid v \in \tgnt{x}{N_1}, w \in \tgnt{x}{N_2}\}.$$
%$$\codim{N_1} = \dim{M} - \dim{N_1}$$
% Sum of all elements within both
If the intersection $N_1 \cap N_2$ is transverse, then $N_1 \cap N_2$ is a submanifold of codimension 
$$\codim(N_1 \cap N_2) = \codim{N_1} + \codim{N_2}.$$
%The corresponding dimension is 
%$$\dim{M} - (\dim{M} - \dim{N_1}) - (\dim{M} - \dim{N_2}).$$
Intuitively, the intersection is transverse if their combined tangent spaces produce the tangent space of the ambient manifold.

%\begin{lem} 
%$f: M \to N$
%By the {\em submersion theorem}, there exists charts 
%Charts $(U, \phi)$ for $M$ centered at $x$ 
%Charts $(V, \psi)$ for $N$ centered at $f(x)$ such that
%$(\psi \circ f \circ \phi^{-1})(r^1, ..., r^m, r^{m+1}, ..., r^n) = (r^1, ..., r^m)$
%In a neighborhood of both
%$(\psi \circ f \circ \phi^{-1})\phi(x) = \psi(y)$
%\end{lem}

\subsubsection{Conditional Constraints:}

We start by defining conditional constraints, a binary partition of $P$ for a constraint $\langle P, R \rangle$.

\note{Finite chart business only affects algorithms}
\note{Projection is open set within}

\note{In future work, I can incorporate domain constraints in the projection to allow it to be a manifold. That way, the preimage can guarantee the intersection works}
\begin{defn}
A {\em conditional constraint} $\langle I, O, R \rangle$ for a constraint $C = \langle P, R \rangle$ is a 
partition of $P$ into a set of {\em input parameters} $I$ and a set of {\em output parameters} $O$. %, and a relation $R$ defined on $I \cup O$ where $I \cap O = \emptyset$. 
%$P = I \cup O$ and $I \cap O = \emptyset$.
\end{defn}
% Independent and dependent parameters (IP) and (DP)
% Oriented manifold
% Projections would count as a new manifold
% Could also call this the inverse projection, inverse image, projection level-set, fiber, ...

Define $\proj_{I}(\bar{z}) = \bar{z}_P$ to be the set-theoretic projection of $\bar{z}$ onto parameters $P$.
Let its inverse, the projection preimage $\proj^{-1}_I(\bar{z}_I)$, be the following:
%$$\proj^{-1}_I(x) = \{\bar{z}_P \in M \mid \bar{z}_I =  x\}.$$
$$\proj^{-1}_I(\bar{z}_I) = \{\bar{z}_P \in R \mid \bar{z}_P = (\bar{z}_I, \bar{z}_O)\}.$$
%Let $\proj_{I}(\bar{z}_P) = \bar{z}_I$ be the set-theoretic projection of $\bar{z}$ onto parameters $I$.
%Let $\proj_P(\bar{Z}) = \{\bar{Z}_P \mid \bar{z} \in \bar{Z}\}$ be the set-theoretic projection of $\bar{Z}$ onto parameters $P$.
%$$\proj_{I}(z_{I_1}, ..., z_{I_{|I|}}, z_{O_1}, ..., z_{O_{|O|}}) = (z_{I_1}, ..., z_{I_{|I|}})$$
%Under certain conditions, ...
Conditional constraints will allow us to implicitly reason about intersections of $\proj_I(R)$ rather than $R$ directly.
%This is advantageous when $\proj_I(R)$ contains an open set.
In order to cleanly describe a lower-dimensional space produced by the intersection of several constraints, we will consider a simplified manifold $M$ that is a subset of $R$.
Additionally, we will make an assumption regarding the relationship between $\proj_I(M)$ and $M$ to use the preimage theorem.
%We identify a technical condition to prune points from $R$ such that, when combined with other constraints, might produce a set of varying dimensionality.
\begin{defn}
A {\em conditional constraint manifold} $\langle I, O, M \rangle$ for a conditional constraint $\langle I, O, R \rangle$ is a nonempty smooth manifold $M \subseteq R$ such that $\proj_I: M \to {\cal Z}_I$ is a submersion $\forall \bar{z}_P \in M$.
% $M$ defined by a finite set of charts. 
\end{defn}

\note{Can just use a version of the original charts for $M$ (i.e. without projection) that is organized to satisfy the submersion theorem. In this case, only need to assume there are a finite set of these charts. This avoids having to worry about parameterized charts. Although, really this kind is a parameterized chart.}
%\begin{equation}
%%\Phi_{\bar{\alpha}} (\bar{z}) = (\varphi_{\alpha_1}^1(\bar{z}_{P_1}), ..., \varphi_{\alpha_n}^n(\bar{z}_{P_n}))
%\Phi_{\bar{\alpha}} (\bar{z}) = (\proj_{d_1} \varphi_{\alpha_1}^1(\bar{z}_{P_1}), ..., \varphi_{\alpha_n}^n(\bar{z}_{P_n}))
%\end{equation}
% Could require infinite charts for a submanifold for some sort of infinite sequence of planes

In our context, $M$ is a submanifold of $\bar{\cal Z}_P = {\cal Z}_{P_1} \times ... \times {\cal Z}_{P_{|P|}}$ because $R \subset \bar{\cal Z}_P$.
The projection $\proj_I = f$ is trivially smooth map between manifold $M$ and product manifold $\bar{\cal Z}_I$. % = {\cal Z}_{I_1} \times ... \times {\cal Z}_{I_{|I|}}$.
Thus, $\proj_I$ is a submersion at $\bar{z}_P \in M$ when $\D{\proj_I}: \tgnt{\bar{z}_P}{M} \to \tgnt{\bar{z}_I}{{\cal Z}_I}$ is surjective.
%The image of $\D{\proj_I}$ is equal to its codomain $T_{\bar{z}_I} {\cal Z}_I$.
%The map is $\D\proj_I$ the same at each point (also part of an identify matrix)
When ${\cal Z}_I = \mathbb{R}^d$, this is equivalent to the subspace of tangent space $\tgnt{\bar{z}_P}{M}$ on parameters $I$ having rank $d$.
% subspace of the tangent space at $\bar{z}_P$ on on variables $I$ has rank $d$. Alternatively, the projection on $I$ of the tangent space has rank $d$.
A more intuitive interpretation is locally at $\bar{z}_P$, there exists coordinates to control $M$ in any direction of $I$.
This restriction is used to ensure that the intersection of $M$ and other conditional constraint manifolds will not be transdimensional.
%contain components of greater dimensionality than expected.
%As a result, $M$ is only composed of {\em regular points} (no critical points) of the projection $\proj_{I}$ and thus $\proj_I$ is a submersion.
This implies the preimage $\proj_{I}^{-1}(\bar{z}_{I})$ for particular values of input parameters $\bar{z}_{I} \in \proj_{I}(M)$ is a submanifold of $M$ of codimension $\dim\bar{\cal Z}_I$.
%is a $(\dim M -\dim{\cal Z}_I)$-dimensional manifold.
Importantly, $\proj_{I}(M)$ is an open set within $\bar{\cal Z}_I$.
This is the key consequence of the submersion assumption which is useful when intersecting arbitrary conditional constraint manifolds.
\note{This is similar to only considering an open subset of the configuration space when doing motion planning} 

%Let $\widehat{M} = \{\bar{z} \in \Theta \mid \bar{z}_P \in M\}$ be its extended form over parameters $\Theta$ where $P \subseteq \Theta$.
%$\widehat{M}$ is a submanifold in $\bar{\cal Z}_{\Theta}$ of dimension $\dim\bar{\cal Z}_{\Theta \setminus P} + \dim{M}$. 
%$$\dim\bar{\cal Z}_{\Theta \setminus P} + \dim{M} = \dim\bar{\cal Z}_{\Theta} + \dim{M} - \dim\bar{\cal Z}_{P}$$

%%%%%%%%%%%

\note{The incremental construction perspective is nice because it is more clear where the new degrees of freedom come from}
\note{Each $\bar{z}_{O_i}$ is effectively one variable (vector)}
%For each $v \in T_x M$, $v = \proj_{I}v + \proj_{O}v$
\note{Vector projection}
\note{Just globally reorder the indices}

%We will perform our analysis on {\em constraint manifolds} rather than on the original constraints themselves.
%We will primarily consider the case when $\proj_I(M)$ an open set within $\bar{\cal Z}_I$, indicating that $\dim \proj_{I}(M) =  \dim \bar{\cal Z}_{I}$.
%We will only consider conditional constraint manifolds where $\proj_I(M)$ an open set within $\bar{\cal Z}_I$. %, indicating that $\dim \proj_{I}(M) =  \dim \bar{\cal Z}_{I}$.
%Suppose, we wish to understand the intersection of manifolds $M$ each defined on a set of parameters $P$. 
%We will relate constraint manifolds back to constraints involving arbitrary relations in the subsequent section.

%\begin{figure}[ht]
%\centering
%\includegraphics[width=0.49\textwidth]{figures/oriented.pdf}
%\caption{Conditional constraint.} \label{fig:simple_factor_graph}
%\end{figure}

%When ${P} = {I}$, $\invprj_{I}(\bar{z}_{I})$ is a $0$-dimensional submanifold which is defined for $\bar{z}_{I} \in \proj_{P}({M}) = {M}$. 
%Thus, the domain of $\invprj_{I}$ can be interpreted as an inclusion map, {\it i.e.} a test for ${M}$. 
%When ${P} = \emptyset$, $\invprj_\emptyset() = {M}$ is simply the manifold itself.
%A constraint manifold $\langle P, M \rangle$ admits $2^{|P|}$ possible conditional constraint manifolds. 

Suppose we are given a set of conditional constraint manifolds $\{\langle I_1, O_1, M_1 \rangle, ..., \langle I_n, O_n, M_n \rangle\}$.
Let $\Theta = \bigcup_{j = 1}^{n} P_j$ be the set of parameters they collectively mention.
Let $S = \bigcap_{j=1}^n \ext{M_j}{\Theta}$ be their intersection when each constraint is extended on $\Theta$. 
%constraint manifolds ${\cal M} = \{\langle P_1, M_1,  \rangle, ..., \langle P_n, M_n \rangle\}$. 
We now present the main theorem which gives a sufficient condition for when $S$ is a submanifold of $\bar{\cal Z}_\Theta$.
This theorem is useful because it identifies when the intersection of several possibly dimensionality-reducing constraints is a space that we can easily characterize.

\note{No need to provide charts. Can also change charts inductively}
\note{Maybe I should do the subtractive form of this that asserts all manifolds are extended?}
\note{I could intersect with its level sets or something if I preferred}
\note{The level sets are disjoint}
\note{I can just say that locally homeomorphic}
\note{Can I do this directly by examining each element?}

\begin{thm} \label{thm:intersection}
%$S$ is a manifold of dimension $\sum_{i=1}^n\big(\dim{M_i} - \dim{{\cal Z}_{I_i}}\big)$ 
If $\{O_1, ..., O_n\}$ is a partition of $\Theta$ and there exists an ordering of $(\langle I_1, O_1, M_1 \rangle, ..., \langle I_n, O_n, M_n \rangle)$ 
such that $\forall i = \{1,...,n\}\; I_i \subseteq \bigcup_{j = 1}^{i-1} O_j$, then
$S = \bigcap_{j=1}^n \ext{M_j}{\Theta}$ is a submanifold of $\bar{\cal Z}_\Theta$ of codimension $\sum_{j=1}^n\codim{M_j}$.
\begin{proof}

%Let $S_i = \bigcap_{j=1}^i \widehat{M}_j$ be the intersection of the first $i$ constraint manifolds.
%Finally, define $N_{i} = \proj_{I_{i}}(M_{i})$ as a result of the input parameter projection for the $i$th constraint manifold.
%
%We proceed by induction on $i$. 
%As a base case, $I_1 = \emptyset$ and $S_1 = M_1$.
%For the inductive step, we assume that after the $i$th intersection, $S_i$ is a manifold of dimensionality $\dim{S_i}$ on parameters $\Theta_i$.
%We will show that $S_{i+1}$ is a manifold of dimensionality $\dim{S_i} + \big(\dim{M_{i+1}} - \dim{{\cal Z}_{I_{i+1}}}\big) $ on parameters $\Theta_{i+1}$.
%
%By assumption $N_{i+1}$ is an open set within ${\cal Z}_{I_{i+1}}$.
%Consider $\widehat{N}_{i+1}$, its extended form over $\Theta_i$ formed by the Cartesian product of $N_{i+1}$ and $\bar{Z}_{\Theta_i \setminus I_{i+1}}$ followed by a rearrangement of indices.
%$\widehat{N}_{i+1}$ is open in $\bar{Z}_{\Theta_i}$ because it is the Cartesian product of open sets.
%Thus, $S_i \cap \widehat{N}_{i+1}$ is a manifold of dimensionality $\dim{S_i}$ as the intersection of an open set and a manifold within a topological space is manifold of the dimensionality, if the intersection exists.

%%%%%%%%%%%%%%%%%%%%

%$$S_{i+1} = S_i \cap M_i$$
Define $\Theta_i = \bigcup_{j = 1}^{i} P_j$ to be union of the first $i$ sets of parameters. 
Notice that $\Theta_i = \bigcup_{j = 1}^{i} O_j$ follows from the second assumption.
Let $S_i = \bigcap_{j=1}^i \ext{M_j}{\Theta_i}$ be the intersection of the first $i$ constraint manifolds over parameters $\Theta_i$.
On the final intersection, $\Theta_n = \Theta$ and $S_n = S$.
%Define $\widehat{S}_i = S_i \times \bar{\cal Z}_{O_{i+1}}$ to be $S_i$ extended with the domain for parameters $O_{i+1}$ by a Cartesian product.
%Let $\widehat{M}_i$ be the extended form of $M_i$ over parameters $\Theta_i$ via the Cartesian product of $M$ and $\bar{\cal Z}_{\Theta_i \setminus P_i}$ followed by a permutation of variable indices.
We can also write $S_i$ recursively as ${S_i = \ext{S_{i-1}}{\Theta_i} \cap \ext{M_i}{\Theta_i}}$ where $S_0 = \emptyset$.
%We can write $S_i$ recursively as $S_i = \widehat{S}_{i-1} \cap \widehat{M}_i$ where $S_0 = \emptyset$.
%Let $S_i = \bigcap_{j=1}^i \widehat{M}_j$ be the intersection of the first $i$ constraint manifolds.
%Finally, define $N_{i} = \proj_{I_{i}}(M_{i})$ as a result of the input parameter projection for the $i$th constraint manifold.
Here, $S_{i-1}$ and $M_i$ are extended by Cartesian products with $\bar{\cal Z}_{O_i}$ and $\bar{\cal Z}_{\Theta_i \setminus P_i}$ respectively. 

We proceed by induction on $i$. 
For the base case, $I_1 = \bigcup_{j = 1}^{0} O_j = \emptyset$.
Thus, $\ext{S_0}{\Theta_1} = \bar{\cal Z}_{O_1}$ and $\ext{M_1}{\Theta_1} = M_1$. 
Since, $M_1 \subseteq \bar{\cal Z}_{O_1}$, 
$$S_1 = (\ext{S_0}{\Theta_1} \cap \ext{M_1}{\Theta_1}) = (\bar{\cal Z}_{O_1} \cap M_1) = M_1$$ 
is a submanifold of codimension $\codim{M_1}$ within $\bar{\cal Z}_{O_1}$.

For the inductive step, we assume that after the ${(i-1)}$th intersection, $S_{i-1}$ is a submanifold of codimensionality $\codim{S_{i-1}}$ on parameters $\Theta_{i-1}$.
%We will show that $S_i$ is a manifold of dimensionality $\dim{S_i} + \big(\dim{M_{i+1}} - \dim{{\cal Z}_{I_{i+1}}}\big) $ on parameters $\Theta_{i+1}$.
We will show that $S_i$ is a manifold of codimension $\codim{S_{i-1}} + \dim{M_i}$ on parameters $\Theta_{i}$.
By isomorphism of product manifold tangent spaces, $\forall \bar{z} \in \ext{S_{i-1}}{\Theta_i}$, $\tgnt{\bar{z}}{\ext{S_{i-1}}{\Theta_i}}$ is isomorphic to 
$\tgnt{\bar{z}_{\Theta_{i-1}}}{S_{i - 1}} \times \tgnt{\bar{z}_{O_i}}{\bar{\cal Z}_{O_i}}$. 
Similarly, $\forall \bar{z} \in \ext{M_i}{\Theta_i}$, $\tgnt{\bar{z}}{\ext{M_i}{\Theta_i}}$ is isomorphic to 
$\tgnt{\bar{z}_{P_i}}{M_i} \times \tgnt{\bar{z}_{\Theta_i \setminus P_i}}{\bar{\cal Z}_{\Theta_i \setminus P_i}}$. % Could make an indicies comment
As a result, the projection for the extended constraint 
$\D{\proj_{\Theta_{i-1}}}: \tgnt{\bar{z}}{\ext{M_i}{\Theta_i}} \to \tgnt{\bar{z}_{\Theta_{i-1}}}{\bar{\cal Z}_{\Theta_{i-1}}}$ is itself surjective for all $\bar{z}$ and therefore is a submersion.
%$$T_{\bar{z}} \widehat{S}_{i-1} = T_{\bar{z}_{\Theta_{i-1}}} S_{i-1} \times T_{\bar{z}_{O_i}} \bar{\cal Z}_i$$

%Recall our assumption that $\proj_{I_i}: M_i \to \bar{\cal Z}_{I_i}$ is a submersion.
%By virtue of the Cartesian product, the projection for the extended constraint manifold $\proj_{\Theta_{i-1}}: \widehat{M}_i \to \bar{\cal Z}_{\Theta_{i-1}}$ is also a submersion.
%Thus, for all $a \in T_{\bar{z}} \bar{\cal Z}_{\Theta_{i-1}},$ there exists $b \in T_{\bar{z}} \widehat{M}_i$ such that $\proj_{\Theta_{i-1}}(b) = a$.
% $$\forall a \in T_x \bar{\cal Z}_{\Theta_{i-1}}, \exists b \in T_x \widehat{M}_i.\; \proj_{\Theta_{i-1}}(b) = a$$
%Consider a tangent $b$ for $a = \proj_{\Theta_{i-1}}(v)$. 
%Their difference $v - b$ is only nonzero on $O_i$. 
%Notice that $\proj_{O_i}(v - b) \in T_{\bar{z}_{O_i}} \bar{\cal Z}_{O_i}$.
%Thus, we have proved $v \in T_{\bar{z}} S_i$ as $b \in T_{\bar{z}} \widehat{M}_i$, $(v - b) \in T_{\bar{z}} \widehat{S}_{i-1}$,
%and $T_{\bar{z}} S_i = T_{\bar{z}} \widehat{M}_i + T_{\bar{z}} \widehat{S}_{i-1}$
%\note{Could decompose $v$ into all its tangent pieces to show that each are in $O_i$.}

%We will prove $S_i$ is a submanifold of codimension $\codim{\widehat{M}_i} + \codim{\widehat{S}_{i-1}}$ in the ambient manifold is $\bar{\cal Z}_{\Theta_i}$.
%Consider each $\bar{z}_{\Theta_i} \in S_i$, we will show that for all $u \in T_{\bar{z}_{\Theta_i}} \bar{\cal Z}_{\Theta_i}$

Because $S_i \subseteq \bar{\cal Z}_{\Theta_i}$, for any $\bar{z} \in S_i$, $\tgnt{\bar{z}}{\ext{M_i}{\Theta_i}} + \tgnt{\bar{z}}{\ext{S_{i-1}}{\Theta_i}} \subseteq \tgnt{\bar{z}}{\bar{\cal Z}_{\Theta_i}}$
For each $\bar{z} \in S_i$, consider any tangent $v \in \tgnt{\bar{z}}{\bar{\cal Z}_{\Theta_i}}$.
% http://mathworld.wolfram.com/OrthogonalDecomposition.html
%Decompose the tangent space into two planes $O_i$ and $\Theta_{i-1}$
Let $v = x + y$ be an orthogonal decomposition of $v$ into a component $x$ defined on parameters $\Theta_{i-1}$ and component $y$ defined on $O_i$.
The first component $x$ is isomorphic to an element of $\tgnt{\bar{z}}{\ext{M_i}{\Theta_i}}$ because its tangent space is surjective to $\bar{\cal Z}_{\Theta_{i-1}}$.
The second component $y$ is isomorphic to an element of $\tgnt{\bar{z}}{\ext{S_{i-1}}{\Theta_i}}$ by the isomorphism to the product space involving $\bar{\cal Z}_{O_i}$.
Thus, $\tgnt{\bar{z}}{\bar{\cal Z}_{\Theta_i}} \subseteq \tgnt{\bar{z}}{\ext{M_i}{\Theta_i}} + \tgnt{\bar{z}}{\ext{S_{i-1}}{\Theta_i}}$.
As a result, $\ext{M_i}{\Theta_i}$ and $\ext{S_{i-1}}{\Theta_i}$ intersect transversally implying that $S_i$ is a smooth submanifold of $\bar{\cal Z}_{\Theta_i}$ with codimension 
$$\codim{\ext{M_i}{\Theta_i}} + \codim{\ext{S_{i-1}}{\Theta_i}} = \codim{M_i} + \codim{S_{i-1}}.$$
%Recall that $\codim{\widehat{M}_i} = \codim{M_i}$.
% By assumption, $\proj_I T_{\bar{z}_P} M = T_{\bar{z}_I} \bar{\cal Z}_I$
% \note{Could state something about the preimage being an open set as well}
\note{Generalize this to when the projection is not the full space but a previous submanifold. This allows us to deal with domain constraints.}

%\note{We are giving a constructive proof of this}
%We construct a coordinate space for $S$ using the new coordinates introduced on each iteration.
%We can construct charts 
%%$$\Phi(\bar{z}_{O_1}, \bar{z}_{O_2} ,..., z_{O_n}) = (\phi_1(\bar{z}_{P_1}), \proj \phi_2(\bar{z}_{P_1}), ...$$
%$$\Phi(\bar{z}_{O_1}, \bar{z}_{O_2} ,..., \bar{z}_{O_n}) = (\phi_1(\bar{z}_{P_1}), \phi_2(\bar{z}_{P_2}), ..., \phi_n(\bar{z}_{P_n}))$$ where $\phi_i$ is a chart at $\bar{z}_{I_i}$.   
%$\phi_1$ is a bijection (diffeomorphism) by default

After $n$ iterations, the entire intersection is a submanifold of codimension $\codim{S} = \codim{M_1} + ... + \codim{M_n}$ in $\Theta$.
Let $d_i = \dim{M_i} - \dim{\bar{\cal Z}_{I_i}}$ be the dimension of the submanifold defined by the projection preimage $\proj_{I_i}^{-1}(\bar{z}_I)$.
Finally, $\dim S$ can be seen alternatively as the sum of $\sum_{i=1}^n d_i$, the number of coordinates introduced on each iteration.
\begin{align*}
\dim{S} &= \dim\bar{\cal Z}_\Theta  - \sum_{i=1}^n \codim{M_i} \\
&= \dim\bar{\cal Z}_\Theta  - \sum_{i=1}^n\big(\dim{\bar{\cal Z}_{P_i}} - \dim{M_i}\big) \\
&= \dim\bar{\cal Z}_\Theta  - \sum_{i=1}^n\big(\dim{\bar{\cal Z}_{O_i}} + \dim{\bar{\cal Z}_{I_i}} - \dim{M_i}\big) \\
&= \sum_{i=1}^n\big(\dim{M_i} - \dim{\bar{\cal Z}_{I_i}}\big) = \sum_{i=1}^n d_i \\
\end{align*}
%$$\widehat{S}_i = S_i \times {\cal Z}_{O_i} \times  {\cal Z}_{\Theta \setminus \Theta_i}$$  
%$$\widehat{M}_i = M_i \times {\cal Z}_{\Theta \setminus \Theta_i}$$ % This technically has stuff in it
%$$\forall x \in S_i, \forall v \in T_x \bar{\cal Z}_{\Theta_i}.\; v \in T_x S_i$$
%$$v = \proj_{e_{O_i}}(v) + \proj_{e_{\Theta_i \setminus O_i}}(v)$$ % Project on standard basis
%$$\proj_{e_{\Theta_i \setminus O_i}}(v) \in T_x {\cal Z}_{O_i}$$
%$$\proj_{e_{\Theta_i \setminus O_i}}(v) \in \widehat{S}_i$$
%$$\forall a \in T_x \bar{\cal Z}_{\Theta_{i-1}}, \exists b \in T_x \widehat{M}_i.\; \proj_{\Theta_{i-1}}(b) = a$$
%By the submersion assertion \\
%$v - b$ is only defined on $O_i$. $v - b \in T_x(\bar{Z}_{O_i})$.
%$S_i$ is undefined on parameters $i+1, ..., |\Theta|$. 
%Each $M_i$ is undefined on parameters $i+1, ..., |\Theta|$. 
%Thus, $\widehat{M}_i$ has tangent space $T_x \times T_x$
%Meanwhile, $M_{i+1}$     $(T_|\Theta_i| + T_x) \times ...$
%The relevant tangents are $O_{i+1}$ (given by $S_i$) and $P_{i+1}$ given by $M_{i+1}$.
% Can just use the vectors that are zero for other coordinates
% Should this be a subset or a superset?
%https://math.stackexchange.com/questions/413766/tangent-space-of-product-manifold
% https://tex.stackexchange.com/questions/204621/matrix-in-latex
% https://tex.stackexchange.com/questions/59517/label-rows-of-a-matrix-by-characters
% https://tex.stackexchange.com/questions/337696/latex-command-for-diagonal-matrix-of-this-kind
\qed
\end{proof}
\end{thm}

%A coordinate space for $S_i$ can be given by the coordinate space for $S_{i-1}$ concatenated with $d_i = \dim{M_i} - \dim{\bar{\cal Z}_{I_i}}$ new coordinates from the submanifold defined by the projection preimage $\proj_{I_i}^{-1}(\bar{z}_I)$.
\note{Show that the manifold story admits samplers}

We will call $S$ a {\em sample space} when theorem~\ref{thm:intersection} holds.
%We will call the resulting manifold the {\em composed constraint manifold} $W$. 
From the partition condition, each parameter must be the output of exactly one conditional constraint manifold. 
Intuitively, a parameter can only be ``chosen'' once.
From subset condition, each input parameter must be an output parameter for some conditional constraint manifold earlier in the sequence. 
Intuitively, a parameter must be ``chosen'' before it can be used to produce values for other parameters.
Theorem~\ref{thm:intersection} can be understood graphically using {\em sampling networks}. 
A sampling network is a subgraph of a constraint network using constraints corresponding to conditional constraint manifolds.
The graphical relationship between a constraint network and a sampling network is analogous to the relationship between a factor graph and a Bayesian network from probabilistic inference~\citep{jensen1996introduction}.
However, this resemblance is purely structural because constraint networks and sampling networks represent sets given as the intersection of several constraints while graphical models represent joint distributions over sets.
%defined on constraint manifolds ${\cal M}$
Each parameter node in a sampling network has exactly one incoming edge. 
Directed edges go from input parameters to constraints or constraints to output parameters.
%A directed edge from a constraint to a parameter indicates that the parameter is sampled from the constraint. 
% A directed edge from a parameter to a constraint indicates that the parameter is an input to the constraint's test.
%A directed edge from a parameter to a constraint indicates that the parameter is a projection parameter for the constraint. 
%Because a parameter must be sampled by exactly one conditional sampler, 
Each parameter is the output of exactly one constraint.
% (${\cal X}$ and ${\cal U}$ count as implicit constraints)
% The graph is a DAG with repect to constraints and a tree with respect to parameter 
% Every constraint network can be transformed into a tree/DAG, but not necessarily one which satisfies the preimage constraint
Finally, the graph is acyclic. 
%As a result, its topological sorts are sample spaces.
%sample the parameters and evaluate the constraints.
% Picture? Do i even have space for this. Could also just refer to the examples

% Still need to decide if I am considering completeness with respect to conditional manifolds. I could just provide all orientations of a constraints. I can also just exhaustively check that my samplers satisfy all possible plans. I feel like the manifolds are something you have the freedom to choose. The idea is more for math land then anything. Also, the conditional versions always exist 
% Do include conditional constraints in the manifolds? The theory nicely extends to them, but it might seem strange to talk about measure in this space if you have all the constraints anyways

%For a given domain, we will assume we have a set of conditional constraint manifolds ${\cal M}$ defined on $(\bar{x}, \bar{u}, \bar{x}')$.
%In order to
When analyzing robustness properties, % for a problem ${\cal P}$, 
we will assume that we are given a set of conditional constraint manifolds ${\cal M}$. % defined on $(\bar{x}, \bar{u}, \bar{x}')$.
This set ${\cal M}$ is typically composed of dimensionality-reducing constraints that have a known analytic form. % across the domain.
We may have multiple conditional constraint manifolds per constraint $C$, resulting from the different ways of conditionalizing $C$. 
Implicit variable domain constraints $\langle (), (x_i), {\cal X}_i \rangle, \langle (), (u_j), {\cal U}_i \rangle \in {\cal M}$ for each variable are always present within ${\cal M}$.
Given a set of constraints ${\cal C}$ defined on parameters $\Theta$, we can produce the corresponding set of conditional constraint manifolds on $\Theta$ by substituting constraints for the conditional constraint manifolds.
For a constraint $C = \langle P, R \rangle$, let ${\cal M}_C$ be the set of conditional constraint manifolds $\{\langle I, O, M \rangle,...\} \subseteq {\cal M}$ associated with $C$ by substituting the input and output parameters $I, O$ for each conditional constraint manifold for with the parameters for $P$. 
For a set of constraints, let ${\cal M}_{\cal C} = \cup_{C \in {\cal C}} {\cal M}_C$ be the union for each constraint $C \in {\cal C}$.
%$$\Pi_{\cal M}({\cal C})$$

% Generate all possible $\Pi$

%When $S$ is a sample space, we can define a measure $\mu_S$ on it. % of appropriate dimensionality. 
%Let $\mu_S$  be the uniform measure on the Euclidean codomain of $S$. 
%Now we can provide a more general definition of robust feasibility. 
%We will define robustness properties with respect to a set of conditional constraint manifolds ${\cal M}$. 
Now we generalize our definition of robust feasibility to be with respect to a set of conditional constraint manifolds ${\cal M}$. 
Note that when ${\cal M}$ is solely composed of variable domain constraints, the new definition is equivalent to the previous definition.
%Robust feasibility is dependent on the constraint manifolds ${\cal M}$ specified.
Specifying ${\cal M}$ allows us to analyze the set of solutions in a lower-dimensional space $S$ given by the constraint manifolds. % where it may have nonzero measure.
Intuitively, a set of constraints is robustly satisfiable ${\cal C}$ if for some parameter values $\bar{z} \in S$, all parameter values $\bar{z}'$ in a neighborhood of $\bar{z}$ satisfy ${\cal C}$.
% Hausdorff dimension
%\note{Don't define a measure. Just state that it contains a plan with a neighborhood that are also solutions. What I have is fine under invariance of zero measure, but unnecessary}

\note{I could also just do just feasible}

\begin{defn} \label{defn:robustly-satisfiable}
A set of constraints ${\cal C}$ is {\em robustly satisfiable} with respect to conditional constraint manifolds ${\cal M}$ 
%if there exists parameter values $\vec{z} \in \bigcap_{C \in {\cal C}} \widehat{C}$ such 
if there exists a sample space $S = \bigcap_{i=1}^n \widehat{M}_i$ formed from conditional constraint manifolds $\{\langle I_1, O_1, M_1 \rangle, ..., \langle I_n, O_n, M_n \rangle\} \subseteq {\cal C}_{\cal M}$ where there exists a satisfying $\vec{z}$ with a neighborhood of parameter values ${\cal N}(\vec{z})$ in $S$ such that ${\cal N}(\bar{z}) \subseteq \bigcap_{C \in {\cal C}} \ext{C}{\Theta}$.
%$\vec{z}'$ in some neighborhood of $\vec{z}$ satisfy ${\cal C}$.
%$\mu_S(S \cap \bigcap_{C \in {\cal C}} \widehat{C}) > 0.$
\end{defn}

%We typically are interested in analyzing algorithms not just for one problem but for multiple 
% Alternatively, I could say that each transition system is the same.
% Conditional samplers and manifolds are a function of the transition system used: fixed object geometry, surfaces, robot geometry, etc
% Can always make auxillary state variables that define things that don't change but define geometries and static info
% The number of objects and robot conf is fixed though
% Constraints not mentioned within ${\cal M}$ can be anything though.
% The goals are given by constraints though. These can be parameterized by auxila

We are typically interested in evaluating robustness properties with respect to ${\cal M}$ for a set of problems defined with the same transition system ${\cal S}$ instead of for just a single problem ${\cal P}$. 
Thus, we a define a domain to be a set of problems ${\cal D}$ to be analyzed with respect to a common set of conditional constraint manifolds ${\cal M}$.
A domain here is similar to the notion of a domain in automated planning~\citep{mcdermott1998pddl} because both fix the dynamics of the system and the objective criteria across a set of problems.

\begin{defn}
A {\em domain} $\langle {\cal D}, {\cal M} \rangle$ is given by a set of problems ${\cal D}$, where each ${\cal P} \in {\cal D}$ has an identical transition system ${\cal S}$, and a set of conditional constraint manifolds ${\cal M}$.
% All problems?
% \forall \bar{x} \in \bar{X}, 
% Can't do all goals because need to describe them nicely using constraints
% Can parameterize the constraints though
\end{defn}

% Parameterized by {\cal S}
This allows us to describe the robustly feasible subset of problems within a domain.

\begin{defn} \label{defn:robustly-feasible}
%A factored transition problem ${\cal P}$ is {\em robustly feasible} with respect to conditional constraint manifolds ${\cal M}$ if there exists a plan skeleton $\vec{a}$ such that ${\cal C}_{\vec{a}}$ is robustly satisfiable with respect to ${\cal M}$.
A factored transition problem ${\cal P} \in {\cal D}$ within a domain $\langle {\cal D}, {\cal M} \rangle$  is {\em robustly feasible} if there exists a plan skeleton $\vec{a}$ such that ${\cal C}_{\vec{a}}$ is robustly satisfiable with respect to ${\cal M}$.
\end{defn}

\note{I could spell out ${\cal M}$}
\note{I think I should include the robustness sampling networks here}

\subsection{Robust Motion Planning}

%Motion planning does not involve any dimensionality-reducing constraints. 
%Thus, the configuration space itself is the only appropriate constraint manifold ${\cal M} = \{\langle (q), {\cal Q} \rangle\}$. 

Our motion planning system involves a single dimensionality reducing constraint \id{Motion}. 
The implicit variable domain constraint $\id{Var}_q = \langle (x_q), {\cal Q} \rangle$ has full dimensionality by default.
Each straight-line trajectory $t$ is uniquely described by its start configuration $q = t(0)$ and end configuration $q' = t(1)$.
Thus, we can notate a straight-line trajectory as $t = (q, q') \in {\cal Q}^{2} \subseteq \mathbb{R}^{2d}$.
As a result, \id{Motion} is a $2d$-dimensional submanifold of ${\cal Q}^{4} \subseteq \mathbb{R}^{4d}$.
We will consider sample spaces resulting from constraint manifolds ${\cal M} = \{\id{Var}_q, \id{Motion}\}$. 
%The collision-free \id{CFree} constraint is not included within ${\cal M}$ because its composition varies from problem to problem within a domain.

\begin{figure}[h]
\centering
\includegraphics[width=0.4\textwidth]{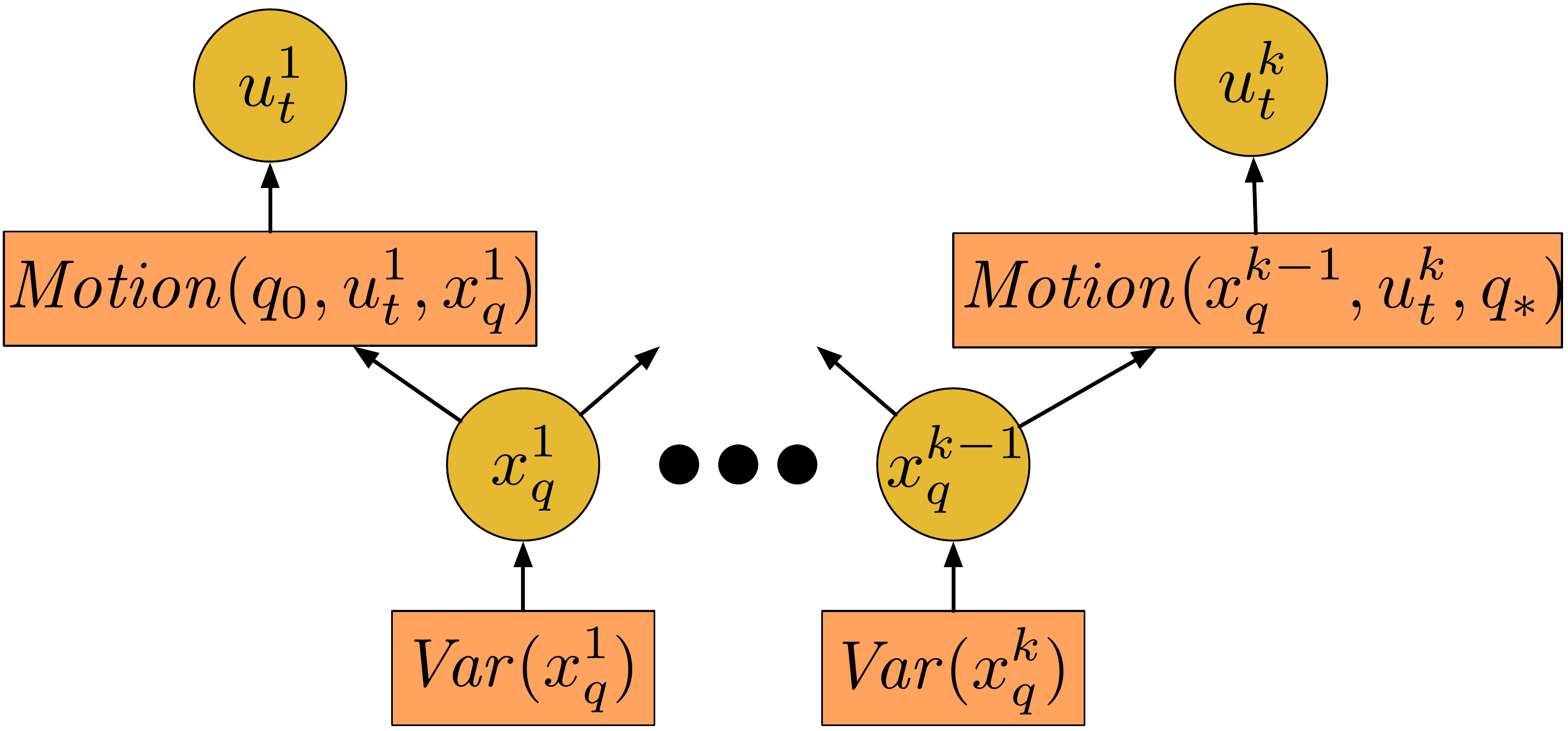}
\caption{Motion planning sampling network for constraint network in figure~\ref{fig:motion_solution}.} \label{fig:motion_dag}
\end{figure}

Figure~\ref{fig:motion_dag} shows a sampling network for the generic motion planning constraint network in figure~\ref{fig:motion_solution}.
This sampling network uses conditional constraint manifolds $\{\langle (), (x_q), \id{Var}_q \rangle, \langle (x_q,x_q'), (u_t), \id{Motion} \rangle\}$.
The projection $\proj_{x_q,x_q'}(\id{Motion}) = {\cal Q}^{2}$ and is thus trivially a submersion.
The projection preimage $\proj_{x_q,x_q'}^{-1}(q, q') = \{t\}$ is a single point corresponding to the straight-line trajectory.
%Thus, the sample space $S \subseteq {\cal Q}^{d(k-1) + 2dk}$ is a submanifold of dimensionality $d(k-1)$.
Thus, the sample space $S \subseteq {\cal Q}^{3k -1} \subseteq \mathbb{R}^{d(3k -1)}$ is a submanifold of dimensionality $d(k-1)$.
Intuitively, the space of satisfying values is parametrizable by each configuration $x_q^i$ for $i = 1, ..., (k-1)$.
As a result, a possible coordinate space of $S$ is ${\cal Q}^{k-1}$ corresponding to Cartesian product of ${\cal Q}$, the variable domain for each $x_q^i$.

% Use graphviz to better plot these?

\begin{figure*}[h]
\centering
\includegraphics[width=0.99\textwidth]{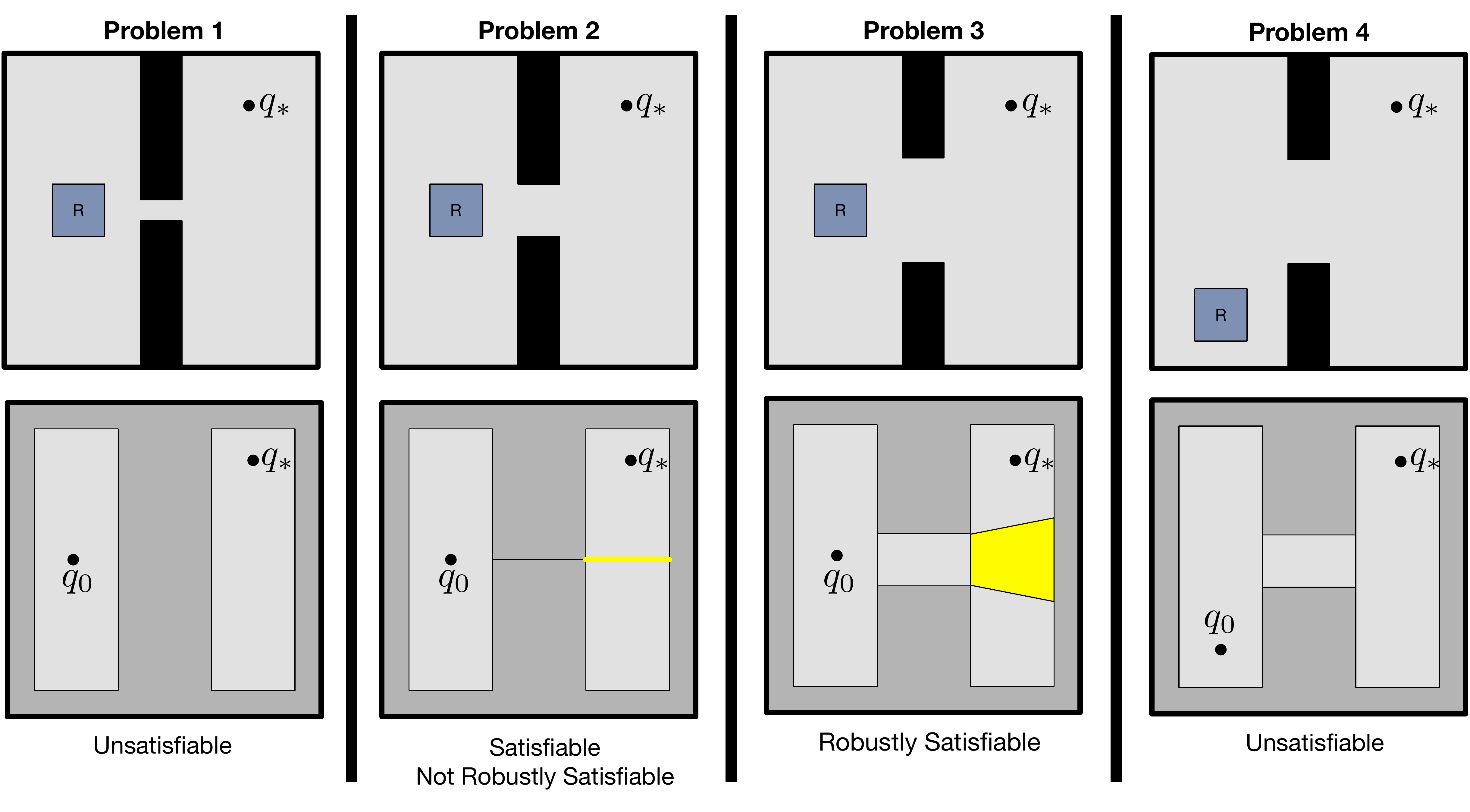}
\caption{From left to right, an unsatisfiable, satisfiable (but not robustly satisfiable), robustly satisfiable, and unsatisfiable motion planning problem for a plan skeleton involving two transitions. The top row displays each problem, varying the environment. The bottom row displays the collision-free configuration space $Q_\id{free}$. Yellow regions of configuration space correspond to values of $x_q^1$ satisfying the plan skeleton.} \label{fig:motion_satisfiable}
\end{figure*}

Subject to this sample space, we can analyze robustly feasible motion planning problems.
Figure~\ref{fig:motion_satisfiable} fixes the plan skeleton $\vec{a} = (\id{Move}, \id{Move})$ and investigates robustness properties of four problems varying the environment geometry. 
The plan skeleton has the following free parameters: $\{u_t^1, x_q^1, u_t^2\}$. However, $u_t^1, u_t^2$ can be uniquely determined given $x_q^1$.
The choices of these parameters must satisfy the following constraints:
\begin{align*}
\{&\id{Motion}(q_0, u_t^1, x_q^1), \id{CFree}(u_t^1), \\
&\id{Motion}(x_q^1, u_t^2, q_*), \id{CFree}(u_t^2)\}.
\end{align*}
Varying the environment only affects the \id{CFree} constraint.
The top row displays a top-down image of the scene and the bottom image shows the robot's collision-free configuration space $Q_\id{free}$ in light grey.
Linear trajectories contained within the light grey regions satisfy their \id{CFree} constraint.
The yellow region indicates values of $x_q^1$ that will result in a plan.
Problem 1 is unsatisfiable because no values of $x_q^1$ result in a plan.
Problem 2 has only a 1 dimensional interval of plan, thus it is satisfiable but not robustly satisfiable.
Problem 3 has a 2 dimensional region of solutions, so it is robustly satisfiable.
Problem 4 is unsatisfiable for the current plan skeleton. 
However, it is robustly satisfiable for a plan skeleton $\vec{a} = (\id{Move}, \id{Move}, \id{Move})$.

\note{Assume objects are open sets?}

\subsection{Robust Pick-and-Place}

\note{I already define dimensionality reducing constraint}

In pick-and-place problems, \id{Stable}, \id{Region}, \id{Grasp}, \id{Kin}, and \id{Motion} are all individually dimensionality-reducing constraints. 
%\id{Stable}, \id{Region} lie on a 3-dimensional manifold corresponding to an object's position resting on a 2-dimensional surface and rotation about the surface normal.
%\id{Grasp} 
Fortunately, we generally understand explicit representations of these sets. % barring collisions with fixed obstacles.
We will consider the following constraint manifolds:
\begin{equation*}
{\cal M} = \{\id{Var}_q, \id{Motion}\} \cup \{\id{Stable}_o, \id{Grasp}_o, \id{Kin}_o \mid o \in {\cal O}\}.
\end{equation*}
We will only consider problems in which $\dim \id{Stable}_o = \dim \id{Region}_o$ in which case $\id{Stable}_o$ captures the reduction of dimensionality from $\id{Region}_o$.
Again, $\id{CFree}$, $\id{CFree}_o$, $\id{CFreeH}_o$, and $\id{CFreeH}_{o,o'}$ are not assumed to be dimensionality-reducing constraints.

\begin{figure}[ht]
\centering
\includegraphics[width=0.49\textwidth]{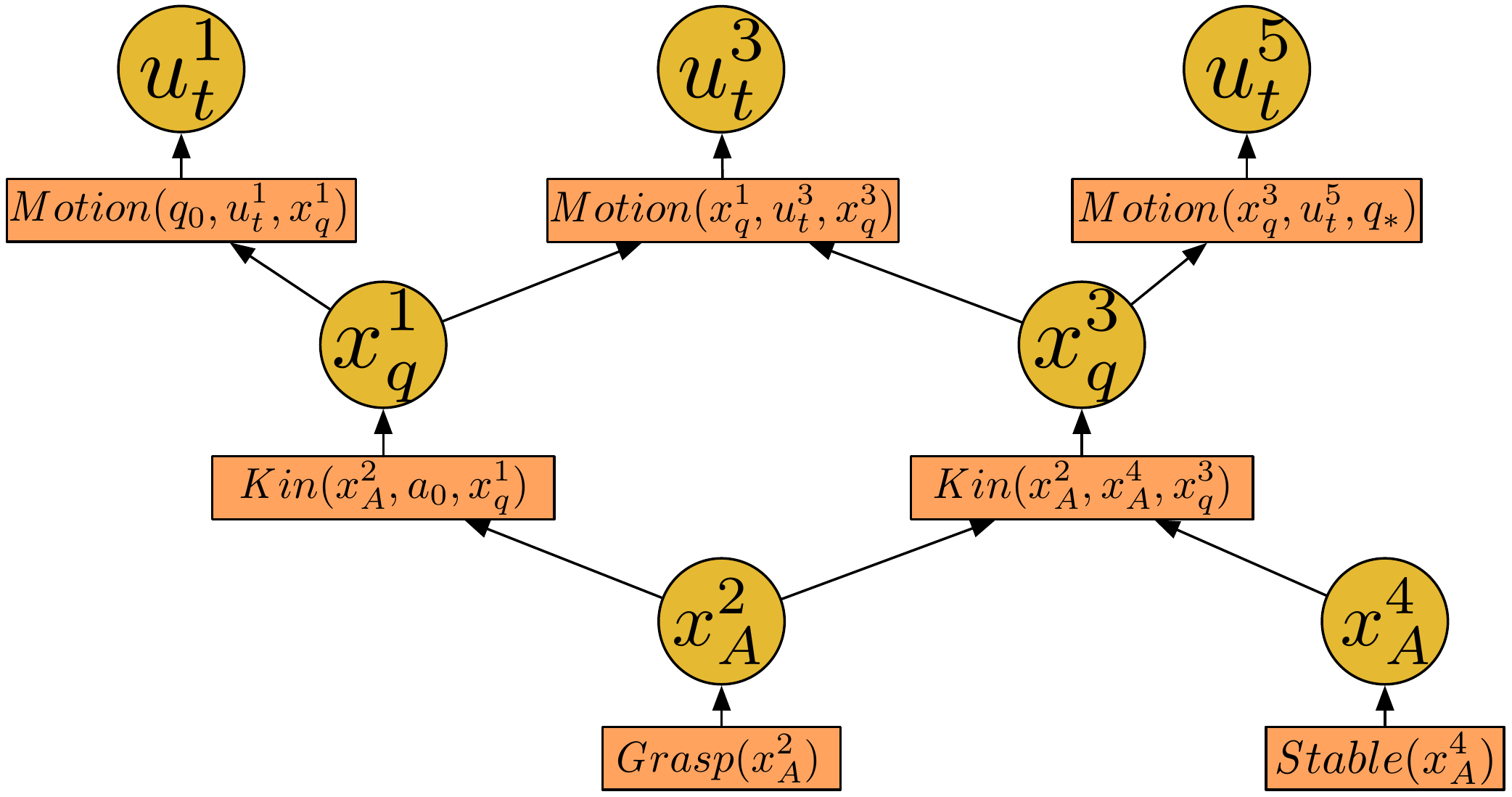}
\caption{Pick-and-place sampling network for the constraint network in figure~\ref{fig:pp_network}.} \label{fig:pp_dag}
\end{figure}

%Pick-and-place problems admit constraint networks that only require a small number of samplers.
% with fully specified initial states (not necessary) and independent conditions on the pose or grasp of a variable and robot
%Let us reconsider the pick-and-place problem and constraint network in figure~\ref{fig:simple_factor_graph_3} as an initial example.
Figure~\ref{fig:pp_dag} shows a sampling network for the pick-and-place constraint network in figure~\ref{fig:pp_network}.
It uses the following conditional constraint manifolds:
\begin{align*}
\{&\langle (x_q,x_q'), (u_t), \id{Motion} \rangle\} \cup \{\langle (), (x_o), \id{Stable}_o \rangle, \\
&\langle (), (x_o), \id{Grasp}_o \rangle, \langle (x_o, x_o'), (x_q), \id{Kin}_o \rangle \mid o \in {\cal O}\}.
\end{align*}
The sampling network satisfies the graph theoretic conditions in theorem~\ref{thm:intersection}. 
% illustrating  the transformation of a constraint network to a sampling network.
%Thus, this graph uses the following conditional samplers: $\proc{Region}(p) \mid \{\}$, $\proc{Grasp}(p) \mid \{\}$, $\proc{Kin}(g, p, q) \mid \{p, g\}$, $\proc{Motion}(q, \tau, q') \mid \{q, q'\}$, $\proc{MotionH}(q, \tau, q', g) \mid \{q, q', g\}$. 
% $\proc{Region}_{p}$, $\proc{Grasp}_{p}$
%$\proc{Kin}({g=?}, {p=?}, q)$
%${Kin}_{g, p, q}(g, p)$
%${Kin}_{q}(g, p)$
%This network uses the following conditional samplers: ${Region}_p$, ${Grasp}_p$, ${Kin}_{q}(g, p)$, ${Motion}_\tau(q, q')$, ${MotionH}_\tau(q, q', g)$. 
Additionally, each conditional constraint manifold has full dimensionality in its input parameter-space. 
$\id{Var}_q$, $\id{Stable}_o$ and $\id{Grasp}_o$ have no input parameters and therefore trivially have full input dimensionality.
The projection $\proj_{x_o,x_o'}(\id{Kin}_o)$ has full dimensionality under the assumption that the robot gripper's workspace has positive measure in SE(3).
We will consider a manifold subset of $\id{Kin}_o$ that satisfies the submersion conditions by omitting kinematic singularities. 
% http://www.cs.columbia.edu/~allen/F15/NOTES/jacobians.pdf
As before, $\proj_{x_q,x_q'}(\id{Motion})$ has full input dimensionality.

\note{I really do need a move here...}
This sampling network structure generalizes to all pick-and-place problems with goal constraints on object poses and robot configurations. % (which use these manifolds).
% Essentially eliminating redundant pick/place and move
Solutions are alternating sequences of $\id{Move}$, \id{Pick}, $\id{MoveH}$, and \id{Place} transitions where \id{Move} and \id{MoveH} may be repeated zero to arbitrarily many times.
% $\id{Move}^*$, \id{Pick}, $\id{MoveH}^*$
Each new cycle introduces a new grasp parameter, pose parameter, two configuration parameters, and two trajectory parameters. 
However, the only interaction with the next cycle is through the beginning and ending configurations which serve as the input parameters for the next move transition. 
Thus, this small set of conditional constraint manifolds is enough to define sample spaces for a large set of pick-and-place problems involving many objects.

\begin{figure*}[ht]
\centering
\includegraphics[width=0.99\textwidth]{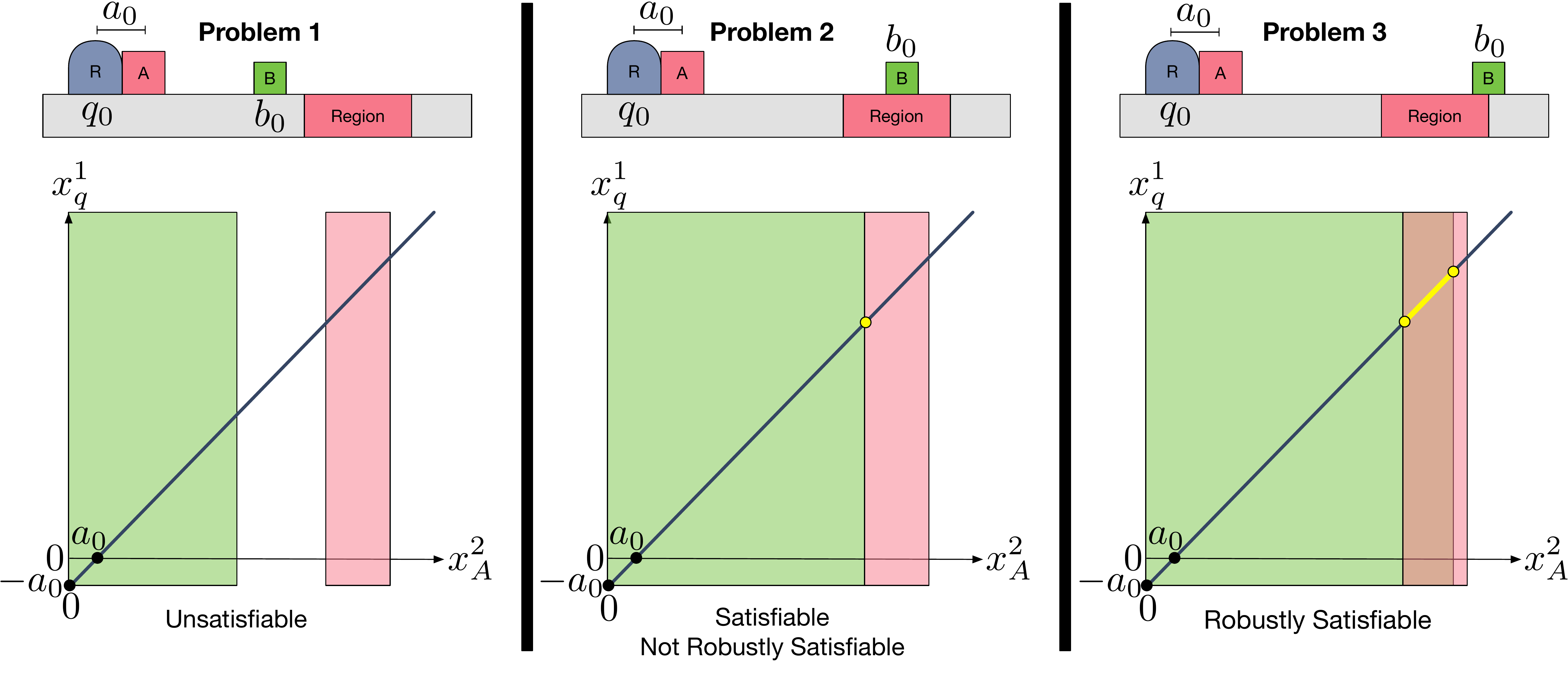}
\caption{From left to right: unsatisfiable, satisfiable (but not robustly satisfiable), and robustly satisfiable pick-and-place problem for plan skeleton $\vec{a} = (\id{MoveH}^A, \id{Place}^A)$. The top row displays each 1-dimensional problem varying $b_0$. The bottom row visualizes the plan parameter-space for free parameters $x_A^2, x_q^1$. Yellow regions of plan parameter-space correspond to pairs of $x_A^2, x_q^1$ satisfying the plan skeleton.} \label{fig:pp_satisfiable}
\end{figure*}

To better visualize a pick-and-place sample space, we investigate a 1-dimensional example where ${\cal Q}, {\cal S}_A, {\cal S}_B \subset \mathbb{R}$.
% \note{$\id{Stable}_o$ is not dimensionality reducing then}
Figure~\ref{fig:pp_satisfiable} fixes the plan skeleton $\vec{a} = (\id{MoveH}^A, \id{Place}^A)$ and investigates robustness properties of three problems varying the initial pose $b_0$ of block $B$.
The robot starts off holding block $A$ with grasp $a_0$, so in the initial state, $x_h^0 = A$ and $x_A^0 = a_0$.
The robot may only grasp block $A$ when $A$ touches its left or right side. 
Therefore, the set of grasps ${\cal G}_A = \{-a_0, a_0\}$ is finite.
Additionally, the kinematic constraint on $(x_o, x_o', x_q)$ results in a plane within $\mathbb{R}^3$.
\begin{equation*}
\id{Kin}_o = \langle (x_o', x_o, x_q), \{q + g = p \mid (g, p, q) \in \mathbb{R}^3\} \rangle
\end{equation*}
The goal constraint ${\cal C}_* = \{\id{Region}_A\}$ requires that $A$ be placed within a goal region.
The plan skeleton has the following free parameters: $x_q^1$ is the final robot configuration, $u_t^1$ is the $\id{MoveH}^A$ trajectory, and $x_A^2$ is the final pose of block $A$. 
Once again, $u_t^1$ can be uniquely determined given $x_q^1$.
The choices of these parameters must satisfy the following constraints:
\begin{align*}
\{\id{Motion}(q_0, u_t^1, x_q^1), \id{CFreeH}(u_t^1, a_0), \id{CFreeH}(u_t^1, a_0, b_0), \\
\id{Grasp}(a_0), \id{Stable}(x_A^2), \id{Kin}(a_0, x_A^2, x_q^1), \id{Region}(x_A^2)\}
\end{align*}
Varying $b_0$ only affects the $\id{CFreeH}(u_t^1, a_0, b_0)$ constraint.
The sample space $S$ is the 1-dimensional manifold given by $\id{Kin}(a_0, x_A^2, x_q^1)$.

Each plot visualizes values of $(x_A^2, x_q^1)$ that satisfy the $\id{Kin}_A$ (blue line), $\id{Region}_A$ (red rectangle) and $\id{CFreeH}_{A,B}$ (green rectangle) constraints.
For this example, we use $x_q^1$ as a surrogate for $u_t^1$ with respect to $\id{CFreeH}(u_t^1, a_0, b_0)$.
The yellow region indicates values of $(x_A^2, x_q^1)$ that satisfy the all the constraints and therefore result in a plan.
Problem 1 is unsatisfiable because the constraints have no intersection.
Problem 2 has only a single plan and therefore has zero measure with respect to $S$.
Because of this, Problem 2 satisfiable but not robustly satisfiable.
Problem 3 has a 1-dimensional interval of plans on $S$, so it is robustly satisfiable.
However, this set of solutions has zero measure with respect to $\mathbb{R}^2$.
Without the identification of the sample space $S$, this problem would not be deemed robustly satisfiable.
% \note{I could treat the environment as a variable with some sort of hairy initial configuration/pose}

\section{Conditional Samplers} \label{sec:samplers}
% Introduce conditional samplers before the math? The math will be used to motivate why they work?

Now that we have identified sample spaces that arise from dimensionality-reducing constraints, we can design samplers to draw values from these spaces. 
Traditional samplers either deterministically or nondeterministically draw a sequence of values $s = (v^1_p, v^2_p, ...)$ from the domain ${\cal Z}_p$ of a single parameter $p$.
In order to solve problems involving dimensionality-reducing constraints using sampling, we must extend this paradigm in two ways.
First, we must intentionally design samplers that draw values of several variables involved in one or more dimensionality-reducing constraints.
Second, we need to construct samplers conditioned on particular values of other variables in order to sample values at the intersection of several dimensionality-reducing constraints.
Thus, our conditional treatment of samplers will closely mirror the treatment of constraints.
%As suggested through our construction of the sample space using conditional constraints, we will sample this space using conditional samplers.

%\begin{defn}
%A {\em conditional sampler} $\psi$ is a function from a set of input values $\bar{z}_I$ for input parameters $I$ to a sampler. 
%%We generate output values from the sampler $\smplr{}$(\id{inps}) using \proc{sample}($\smplr{}$(\id{inps})).
%The sampler generates a sequence of output values $\psi(\bar{z}_I)$ for output parameters $O$ using \proc{sample}($\psi(\bar{z}_I)$).
%\end{defn}

%\begin{defn}
%A {\em sampler} $s = \langle s, O, {\cal C} \rangle$ is a sequence of output values $s = (\bar{z}_O^1, \bar{z}_O^2, ...)$ for parameters ${\cal O}$ that satisfy a set of constraints ${\cal C}$.
% Variables $O$
%\end{defn}

\note{Decide should I use $v$ or not?}

\begin{defn}
A {\em conditional sampler} $\psi = \langle I, O, {\cal C}, f \rangle$ is given by a function $f(\bar{v}_I) = (\bar{v}_O^1, \bar{v}_O^2, ...)$ from input values $\bar{v}_I$ for parameter indices $I$ to a sequence of output values $\bar{v}_O$ for parameter indices $O$. %The {\em domain} of $f$ satisfies a set of constraints ${\cal D}$ on $I$. 
The {\em graph} of $f$ satisfies a set of constraints ${\cal C}$ on $I \cup O$.
\end{defn}
% I think I want to fill in the output values of these?

We will call any $\psi$ with no inputs $I = ()$ an {\em unconditional sampler}.
%The domain implicitly contains input parameter constraints $\id{Var}_p$ for $p \in I$. 
The graph implicitly contains output variable domain constraints $\id{Var}_p = \langle (z_p), {\cal Z}_p \rangle$ for $p \in O$.
The function $f$ may produce sequences that are enumerably infinite or finite.
Let $\proc{next}(f(\bar{v}_I)) = \bar{v}_O$ produce the next output values in the sequence.
The function $f$ may be implemented to nondeterministically produce a sequence using random sampling.
%To start, we will just consider a single outcome for the sequence $f(\bar{v}_I)$.
%Later, we will revisit the probabilistic nature of a sampler by making statements about its distribution of possible outcomes.
% Except of a set of outcomes with zero probability
% Probability that you miss an open set is zero
% https://en.wikipedia.org/wiki/Almost_surely
% https://en.wikipedia.org/wiki/With_high_probability
%Conditional samplers can be annotated with any constraints their input and corresponding output values are guaranteed to satisfy. 
%Additionally, they can be given input domain constraints representing necessary conditions on the inputs for the sampler to be defined. These annotations can prove algorithmically useful as they inform planners about the function of the blackbox samplers.
%\note{I should make a new way of expressing constraints that puts emphasis on the effects things}
It is helpful algorithmically to reason with conditional samplers for particular input values.
\begin{defn}
A conditional sampler {\em instance} $s = \psi(\bar{v}_I)$ is a conditional sampler paired with input values $\bar{v}_I$. % that satisfy the domain of $\psi$.
\end{defn}

% I definitely don't want to assume anything about the structure of the constraint. I just want to assert what it does
% Assume one output for simplicity to start?
%Could make the equivalence that $\phi$ is $\psi$?
% Leslie - Is there anything to force the sampler to somehow "cover" the output space?

% Should I allow use of conditional samplers not useful for a plan skeleton or even conditional samplers not attached to a constraint?
% Rejection sampling approach to sampling from the joint constraint using a surrogate network
% Sampling distribution conditioned on the evidence of the other constraints not directly sampled
% We can always condition on all the previous filled in values (as well as the skeleton)
%If a problem is robustly feasible then there exists a space in which the set of solutions has nonzero measure. 
% known space
%Naturally, an approach for solving these problems is then to sample that space. 
%Recall that these spaces are the result of composing preimage projections, which define a manifold given values of the projection parameters. 
%To sample from preimage projections, we specify {\em conditional samplers} $\smplr{}$, functions from projection parameters to a sampler instance on the corresponding manifold. 
% Could put in $Z_P \to Z_O \times Z_O \times ...$
%Intuitively, conditional samplers are functions which produce samplers specific to the input function values.

We typically design conditional samplers to draw values from the conditional constraint manifolds present within a domain.
%For example, consider a  for grasp $g$, configuration $q$, and pose $p$. 
For example, consider the conditional sampler $\psi_\id{IK}^o$ for the kinematic constraint $\id{Kin}_o$. 
\begin{align*}
%\psi_\id{IK}^o =& \langle (o, o'), \{\id{Grasp}(o), \id{Stable}(o')\}, (q), \\
\psi_\id{IK}^o =& \langle (x_o, x_o'), (x_q), \{\id{Kin}_o\}, \proc{inverse-kin}\}\rangle% Prune CFree grasps?
\end{align*}
%A conditional sampler for this projection, denoted ${Kin}(g, p) \to q$, would 
$\psi_\id{IK}^o$ has input parameters $I = (x_o, x_o')$ and output parameters $O = (x_q)$. 
Its function $f = \proc{inverse-kin}$ performs inverse kinematics, producing configurations $q$ that have end-effector transform $pg^{-1}$ for world pose $p$ and grasp pose $g$.
% $\proj{Kin}(g, p) \to q$
%For a 6 degree-of-freedom manipulator in SE(3), this would sample from a 0-dimensional manifold where there are at most 16 possible solutions.
For a 7 degree-of-freedom manipulator in SE(3), this would sample from a 1-dimensional manifold. % where there are at most 16 possible solutions.
\note{Describe domain conditions?}
\note{Talk about conditional samplers in practice?}

% I don't think any of my examples actually choose two values for things. It would be more general to sample from the marginal likely than to have a sampler for joint outputs
%Additionally, to directly sample the constraint instead of only the constraint manifold, the sampler must prune output values that exceed joint limits or result in collisions with fixed obstacles.
% Leslie - Isn't this from a different constraint (which we might, generally, handle via rejection, because it isn't dimensionality reducing?)

Like conditional constraints, conditional samplers can be composed in a {\em sampler sequence} $\vec{\psi} = (\psi_1, ..., \psi_k)$ to produce a vector of values for several parameters jointly.
A well-formed sampler sequence for a set of parameters $\Theta$ satisfies $\Theta =  \bigcup_{j = 1}^{n} O_j$ as well the conditions from theorem~\ref{thm:intersection}.
Each parameter must be an output of exactly one conditional sampler and later conditional samplers must only depend on earlier samplers.
%For sampler sequence $(s_1, ..., s_k)$ of unconditional samplers, the countable set of values produced is simply $s_1 \times ... \times s_k$.
The set of values generated by the sampler sequence $\vec{\psi}$ is given by
\begin{equation*}
F(\vec{\psi}) = \{\vec{v} \mid \vec{v}_{O_1} \in f_1(), \vec{v}_{O_2} \in f_2(\vec{v}_{I_2}), ..., \vec{v}_{O_k} \in f_k(\vec{v}_{I_k})\}.
%\bigcup_{v_{O_1} \in f_1()} \bigcup_{v_{O_2} \in f_1(v_{O_1})}
\end{equation*}

% sample strategy
% sampling process constructed by chaining several conditional samplers $\phi_1, ..., \phi_n$.
We are interested in identifying combinations of conditional samplers that will provably produce a solution for robustly feasible problems.
%In particular, we are interested in properties of the samplers that are {\em almost surely} true, {\it i.e.}, properties that hold with probability one.
Similar to ${\cal M}_{\cal C}$, for a set of constraints ${\cal C}$ and set of conditional samplers $\Psi$, let $\Psi_{\cal C}$ be the set of conditional samplers appropriate for each constraint.

\begin{defn} \label{defn:sufficient}
A set of conditional samplers $\Psi$ is {\em sufficient} for a robustly satisfiable plan skeleton $\vec{a}$ with respect to conditional constraint manifolds ${\cal M}$ if there exists a sampler sequence $\vec{\psi} \subseteq \Psi_{{\cal C}_{\vec{a}}}$ such that 
%almost surely $F(\vec{\psi}) \cap {\cal C}_{\vec{a}} \neq \emptyset$.
$F(\vec{\psi}) \cap {\cal C}_{\vec{a}} \neq \emptyset$ with probability one.
%that with probability one samples a parameter assignment satisfying ${\cal C}_{\vec{a}}$ within a finite number of calls to $\proc{sample}$.
%$\Psi$ is {\em sufficient} for a domain ${\cal D}$ with respect to ${\cal M}$ if for all robustly feasible ${\cal P} \in {\cal D}$, there exists a robustly satisfiable plan skeleton $\vec{a}$ for which $\Psi$ is sufficient.
% This makes no conditions on how these act
% This doesn't allow the strange circular samplers though
\end{defn}
% Need to make sure you reuse old values in the sampler sequence
\note{Might need repeated samplers. Should I fix the output of each sampler (even when used twice). Probably. I think just think of random samplers using the principle of deferred decisions}

%This definition can be extended to a domain. 

\begin{defn} \label{defn:sufficient-domain}
%A set of conditional samplers 
$\Psi$ is {\em sufficient} for a domain $\langle {\cal D}, {\cal M} \rangle$ if for all robustly feasible ${\cal P} \in {\cal D}$, there exists a robustly satisfiable plan skeleton $\vec{a}$ for which $\Psi$ is sufficient.
%satisfying definition~\ref{defn:sufficient}.
\end{defn}
% Could also do dense or covers. Sufficient is really a very weak property

In practice, the set of conditional samplers $\Psi({\cal S})$ is often a function of the common transition system ${\cal S}$ for a domain $\langle {\cal D}, {\cal M} \rangle$, generating a different set of conditional samplers $\Psi$ per domain. 
Alternatively, the transition system ${\cal S}$ can be thought of as a meta-parameter for each $\psi \in \Psi$.
For example, a robot configuration conditional sampler can be automatically constructed from the number of joints and joint limits within a transition system ${\cal S}$.
Similarly, a stable object placement conditional sampler can be parameterized by the stable surfaces defining the \id{Stable} constraint for a particular ${\cal S}$.
% Conditional sampler could also just take in the ${\cal S}$ as a meta-parameter

%%%%%%%%%%%%%%%

%\begin{thm}
%If each conditional sampler $\psi \in \vec{\psi}$ is dense for its corresponding conditional constraint, the sampler sequence $\psi \in \vec{\psi}$ is dense in the intersection of these constraints.
%%is dense in the sample space.
%%A sequence of samples is {\em dense} for their corresponding sets of oriented manifold.
%\begin{proof}
%% https://math.stackexchange.com/questions/623314/cartesian-product-and-closure
%% https://math.stackexchange.com/questions/302213/question-about-closure-of-the-product-of-two-sets
%% Closure of cartesian product is cartesian product of their closures
%%Use the definition of the manifold for the sample space to show this.
%% I could even just trace the samples to determine that it does work. Was my concern that about the continuity of the space? 
%% https://math.stackexchange.com/questions/313006/how-to-show-that-fs-subset-y-is-dense-when-f-is-continuous-and-surjecti
%\end{proof}
%\end{thm}

%Conditional samplers can directly sample conditional constraints by performing rejection sampling on the conditional constraint manifold. % containing each conditional constraint.
\note{Maybe this is how I should formalize a domain. Certain constraints are fixed. Other constraints are not. A algorithm is probabilistically complete if for ANY possible value of the constraints that satisfy the robustness condition. It will succeed.}
A conditional sampler must generally produce values covering its constraint manifold to guarantee completeness across a domain.
This ensures that the sampler can produce values within every neighborhood on the constraint manifold.
This property is advantageous because it is robust to other adversarial, worst-case constraints that only admit solutions for a neighborhood of values on the constraint manifold that the sampler is unable to reach.
%If there exists a solution, the sampler will cover it
In motion planning, a traditional sampler $s = (v^i)_{i=1}^\infty$, ...  is {\em dense} with respect a topological space $Z$ if the topological closure of its output sequence is $Z$~\citep{Lavalle06}. 
The {\em topological closure} of $s$ is the union of $s$ and its {\em limit points}, points $z \in Z$ for which every neighborhood of $z$ contains a point in $s$.
We extend this idea to conditional samplers for conditional constraint manifolds.
%\note{Densely sample coordinate space and take cartesian products}

\begin{defn}
A conditional sampler $\psi = \langle I, O, {\cal C}, f \rangle$ is {\em dense} with respect to a conditional constraint manifold $\langle I, O, M \rangle$ if $\forall \bar{v}_I \in \proj_I(M)$, $f(\bar{v}_I)$ is dense in $\proj_O(\proj_I^{-1}(\bar{v}_I))$ 
with probability one.
%with high probability.
% Note, that we could sample from a higher dimensional space in which case the chance of producing any correct samples is zero
\end{defn}
% We defined with respect to the constraints themselves rather than the manifolds
% The manifolds don't really exist so we can't sample from them
% The dense conditions only makes sense when talking about sampling for constraints. However, we could have other strategies that don't do this at all

\note{The finite number of manifolds thing is to just assert that we can make dense samplers period}
\note{This is where I might want to talk about inverses and composing functions}
\note{The constructive generation is different than charts in terms of its domain}
%$$\{\bar{z} \mid \bar{z}_1 \in \psi^1(), \bar{z}_2 \in \psi^2(\bar{z}_1), ...,  \psi^n(\bar{z}_{1, ..., n-1})\}$$
\note{Closure operator}
\note{I could try to compose and do the intersection?}
\note{Define a conditional sampler $\psi$ to be {\em probabilistically dense} for $\langle I, O, M\rangle$ if it is dense with probability one.}

This definition encompasses a large family of deterministic and nondeterministic sampling strategies including sampling $\proj_O(\proj_I^{-1}(\bar{v}_I))$ uniformly at random and independently.
The following theorem indicates that dense conditional samplers for the appropriate conditional constraints will result in a sufficient collection of samplers.
Thus, a set of dense conditional samplers for individual conditional constraint manifolds can be leveraged to be sufficient for any robustly satisfiable plan skeleton.

\begin{thm} \label{thm:sufficient}
%A set of conditional samplers $\Psi$ is sufficient for a robustly satisfiable plan skeleton $\vec{a}$ if for each conditional constraint manifold $\langle I, O, M \rangle \in {\cal M}$, there exists a conditional sampler $\psi \in \Psi$ that is dense for $\langle I, O, M \rangle$.
A set of conditional samplers $\Psi$ is sufficient for a domain $\langle {\cal D}, {\cal M} \rangle$ if for each conditional constraint manifold $\langle I, O, M \rangle \in {\cal M}$, there exists a conditional sampler $\psi \in \Psi$ that is dense for $\langle I, O, M \rangle$.
\begin{proof}

\note{Do I even have to invert the charts?}
Consider any robustly feasible ${\cal P} \in {\cal D}$ and by definition~\ref{defn:robustly-feasible} any robustly satisfiable $\vec{a}$ plan skeleton for ${\cal P}$.
By definition~\ref{defn:robustly-satisfiable}, there exists a sample space $S = \bigcap_{i=1}^n \widehat{M}_i$ formed from conditional constraint manifolds $\{\langle I_1, O_1, M_1 \rangle, ..., \langle I_n, O_n, M_n \rangle\} \subseteq {\cal C}_{\cal M}$ and a neighborhood of parameter values ${\cal N}(\vec{z}) \subseteq S$ satisfying ${\cal C}_{\vec{a}}$.
%centered around $\bar{z}$ 
%such that ${\cal N}(\vec{z}) \subseteq \bigcap_{C \in {\cal C}} \widehat{C}$
Consider each $i$ iteration in theorem~\ref{thm:intersection}. 
%The $S_i$ can be described as $S_{i-1}$ concatenated 
%by each projection preimage 
Let $\bar{y}_i \in \mathbb{R}^{d_i}$ be the coordinates introduced on the $i$th projection preimage where $d_i = \dim{M_i} - \dim{\bar{\cal Z}_{I_i}}$.
Consider the atlas for $S$ constructed by concatenating combinations of the charts for each projection preimage in the sequence.
There exists an open set in the coordinate space of $S$ centered around $\vec{z}$ that satisfies ${\cal C}_{\vec{a}}$.
Consider a subset of the coordinate space of ${\cal N}(\vec{z})$ given as $\bar{Y}_1 \times ... \times \bar{Y}_n \subset \mathbb{R}^{\dim{S}}$ where each $\bar{Y}_i \subset \mathbb{R}^{d_i}$ is an open set.
Any combination of values $(\bar{y}_1, ..., \bar{y}_n)$ where $\bar{y}_i \in \bar{Y}_i$ is contained within this set.
%, axis-aligned hypercube 
%There exists an open interval $Y_j$ for each coordinate $y_j$ such that any combination of coordinate values in each interval is contained within this hypercube.

Consider a procedure for sampling $(\bar{y}_1, ..., \bar{y}_n)$ that chooses values for $\bar{y}_i$ in a progression of $i$ iterations.
On iteration $i$, the value of $\bar{z}_{\Theta_{i-1}}$ is fixed from the choices of $\bar{y}_1, ..., \bar{y}_{i-1}$ on previous iterations.
Consider the submanifold defined by the projection preimage $\proj^{-1}_{I_i}(\bar{z}_{I_i})$ for the $i$th conditional constraint manifold $\langle I_i, O_i, M_i \rangle$. 
Recall that $I_i \subseteq \Theta_{i-1}$.
By assumption, there exists a conditional sampler $\psi \in \Psi$ that is dense for $\langle I, O, M \rangle$.
Thus, $\psi(\bar{z}_{I_i})$ densely samples $\proj^{-1}_{I_i}(\bar{z}_{I_i})$ producing output values $\bar{z}_{O_i}$.
Correspondingly, $\psi(\bar{z}_{I_i})$ densely samples the coordinate space of $\bar{y}_i$.
Upon producing $\bar{y}_i \in \bar{Y}_i$, the procedure moves to the next iteration.
Because $\bar{Y}_i$ is open within $\mathbb{R}^{d_i}$ and the sampling is dense, the sampler will produce a satisfying coordinate values $\bar{y}_i$ within a finite number of steps with probability one. 
After the $n$th iteration, $\bar{z}$ given by coordinates $(\bar{y}_1, ..., \bar{y}_n) \in \bar{Y}_1 \times ... \times \bar{Y}_n$ will satisfy constraints ${\cal C}_{\vec{a}}$.
\qed
\note{I should probably talk about the charts used}

%the conditional sampler $\psi_i$ for conditional constraint manifold $\langle I_n, O_n, M_n \rangle$ will densely sample the intervals corresponding to $i$th coordinates.
%Let $\bar{y}_I \in \mathbb{R}^d$ be the set of coordinates introduced by the $i$ projection preimage.
%When produces $\bar{y}_I \in (Y_{I_1} \times ... \times Y_{I_{|I|}})$, we proceed to the next iteration.
%Because $\psi_i$ is dense, with probability one, there is a $\bar{y}_I$. Once produced we consider the next iteration.
%\note{The sampler doesn't need to know about the phases}
% Proceed nondeterministically (as in Turing Machine?)
\end{proof}
\end{thm}

\note{Do I need to enforce that the samplers return to their values arbitrarily many times?}
\note{Exponential convergence? This would require talking about measure though.}

%It seems like the sample networks still work here. Can say that you need a strategy for sampling and testing. Want to ask the question if a solution exists, can this sample it?

% Do I want to say the samplers covers the domain or the problem?
% Do I just want to say that they have the capacity of generating a solution?
% Do I want to enforce they cover the full space?
% If they don't cover the space, surely one could construct constraints that capitalize on this
% The motion planning story requires covering the full space
% Is there a way to do this across the domain rather than on problem instances?
% Only want to require samplers that are useful
% What if a sampler avoids sampling all the space because it knows about constraints

% We actually do avoid sampling the manifolds when we sample IK solutions and discard those colliding with the environment. Should this be described as separate conditions for IK and static-collision free
% The manifold analysis is purely theoretical just to certify that problem is feasible for some samplers

\subsection{Motion Planning Samplers}

In order to apply theorem~\ref{thm:sufficient} to the sampling network (figure~\ref{fig:motion_dag}), we require conditional sampler for the $\id{Var}_q$ and \id{Motion} conditional constraint manifolds.
For $\id{Var}_q$, we provide an unconditional sampler $\psi_Q$ that is equivalent to a traditional configuration sampler in motion planning.
$$\psi_Q = \langle (), (x_q), \{\id{Var}_q\}, \proc{sample-conf} \rangle$$
Its function \proc{sample-conf} densely samples ${\cal Q}$ using traditional methods~\citep{Lavalle06}. % such as a uniformly at random sampler. %sampling ${\cal Q}$.
For example, one implementation of \proc{sample-conf} nondeterministically samples ${\cal Q}$ uniformly at random and independently.
% with probability density bounded below by zero.
%\proc{sample-conf} can be implemented nondeterministically by sampling ${\cal Q}$ uniformly at random or deterministically such as by using a low-discrepancy sequence~\citep{Lavalle06}.
%In practice,  \proc{sample-conf} can be implemented to avoid sampling configurations $q \notin Q_\id{free}$ using rejection sampling.
%This will not affect the sufficiency of samplers as $q \notin Q_\id{free}$ will always violate $\id{CFree}(t)$ for any trajectory where $t(0) = q$ or $t(1) = q$.
%However, this can result in improvements in sampling efficiency as fewer calls to \kw{next} are required to produce configurations that may satisfy plan skeleton constraints.
%Additionally, pruning the unsatisfiable configurations results in smaller overhead awe
For \id{Motion}, we provide a conditional sampler $\psi_T$ that simply computes the straight-line trajectory between two configurations.
%$$\psi_T = \langle (q, q'), \emptyset, (t), \{\id{Motion}(q, t, q')\}, \proc{straight-line} \rangle$$
%$$\psi_T = \langle (q, q'), \emptyset, (t), \{\id{Motion}(q, t, q'), \id{CFree}(t)\}, \proc{line} \rangle$$
$$\psi_T = \langle (x_q, x_q'), (u_t), \{\id{Motion}\}, \proc{straight-line} \rangle$$
$\proc{straight-line}$ generates a sequence with just a single value corresponding to the trajectory $t$ between $q, q'$.
% Bidirectional edges?
\note{Tomas - cyclic coordinates not unique}

%Because motion planning our motion planning transition system $\bar{x} = (x_q)$state and control , $\bar{u} = (u_t)$ 
Because our motion planning transition system exhibits little factoring, $\psi_Q$ samples entire states $(x_q) = \bar{x}$. 
Additionally, each trajectory $(u_t) = \bar{u}$ computed by $\psi_T$ corresponds to either a single transition when $u_t \in \id{CFree}$ or otherwise no transitions.
Thus, $\psi_Q$ and $\psi_T$ can be seen as directly sampling the entire state and control-spaces.

\note{No motivation to sample configurations not in collision with the network. Could add a collision constraint but it is subsumed}
\note{Only need to densely sample the set of solutions. In which case, don't even need dense. Can get away with just sampling a particular set of solutions that are a solution.}
\note{The reason we have the denseness is that we want to assume our samplers know nothing about the problem. Sampler that is invariant of the other constraints in the world}
\note{We are assuming that we have all the tests}

%Each parameter configuration along the plan skeleton is sampled directly from ${\cal Q} \to q$. 
%Each collision-free constraint is evaluated as a test ${CFree}(q, q')$. 
%The projection $\proj_{q, q'}{CFree}$ has the same dimensionality as ${\cal Q}^2$ (assuming $Q_{\it free}$ is not degenerate), so the composed constraint network exists. 
%Thus, only a single conditional sampler is needed to plan for a motion planning domain.
% What if we are in a problem where CFree is entirely lower-dimensional...
% This can be depicted graphically by converting the constraint network in a type of directed acyclic graph (DAG). The set of incoming edges from a constraint  
% And edges from parameters to constraints indicate that the constraint will be verified through a test.

%\subsection{Multi-modal motion planning}
%Multi-modal motion planning problems are admit a sampling network.
%\note{Finish this if we want to include it. Highlight why the non-factored form hurts sampling for some problems.}
%% Note the strategy is to show that many domains are chainable
%% Need to have conditions on the goal though
%
%\begin{figure}[ht]
%\centering
%\includegraphics[width=0.48\textwidth]{figures/mode_sample.pdf}
%\caption{Multi-modal motion planning sampling DAG.} \label{fig:simple_factor_graph}
%\end{figure}
%% Can always have a mode sampler that produces adjacent modes as well as goal modes

\subsection{Pick-and-Place Samplers}

%For \id{Kin}, a nonzero volume of poses and grasps admit an inverse kinematic solution.
%The same holds for \id{Motion} and \id{MotionH} but with respect to pairs of configurations and grasp transforms.
%Thus, the following set of conditional samplers is sufficient for this plan skeleton:
%The sampling network uses the following conditional samplers: 
%${Region \& Stable} \to p$, ${Grasp} \to p$, ${Kin}(g, p) \to q$, ${Motion}(q, q') \to \tau$, ${MotionH}(q, q', g) \to \tau$. 
%$\langle (), (p), \id{Region} \cap \id{Stable} \rangle$, $\langle (), (p), {Grasp} \rangle$, $\langle (g, p), (q), {Kin} \rangle$, $\langle (q, q'), (\tau), \id{Motion} \rangle$, $\langle (q, q', g), (\tau), \id{MotionH} \rangle$. 
%The projections for ${Kin}(g, p) \to q$, ${Motion}(q, q') \to \tau$, ${MotionH}(q, q', g) \to \tau$ have full dimensionality.

In addition to $\psi_\id{IK}^o, \psi_Q, \psi_T$, the pick-and-place sampling network (figure~\ref{fig:pp_dag}) requires the following unconditional samplers $\psi_G^o$ and $\psi_P^o$ for the \id{Grasp} and \id{Stable} constraints:
$$\psi_G^o = \langle \emptyset, (o), \{\id{Grasp}_o\}, \proc{sample-grasp}\rangle$$ % Prune CFree grasps?
$$\psi_P^o = \langle \emptyset, (o), \{\id{Stable}_o\}, \proc{sample-pose}\rangle.$$ % Prune CFree grasps?
%$$\psi_R^o = \langle (), \emptyset, (o), \{\id{Region}(o), \id{Stable}(o)\}, \proc{sample-region}\rangle$$ % Prune CFree grasps?
The implementation of \proc{sample-grasp} depends on the set of available grasps ${\cal G}_o$.
In our experiments, ${\cal G}_o$ is a finite set.
%If \id{Grasp} is a discrete set, \proc{sample-grasp} can simply enumerate the set.
%Otherwise, \proc{sample-grasp} must sample the free continuous parameters.
\proc{sample-pose} will typically sample $(x, y, \theta)$ 2-dimensional pose values in the coordinate frame of each surface and returns the poses in the world frame.
%Like $\psi_Q$, in practice, $\psi_P^o$ and $\psi_\id{IK}^o$ can be implemented to prune pose and configuration samples that are in collision with the fixed environment.

In contrast to the motion planning transition system, the pick-and-place transition system exhibits substantial factoring.
Factoring provides several benefits with respect to sampling. 
First, by exposing subsets of the states and controls, factoring enables the design of samplers that affect only several variables and constraints.
% Leslie - And, again, the same conditional sampler can be re-used to make many instances.
Otherwise, one would require specifying samplers that produce full states $\bar{x}$ as is typically done in multi-modal motion planning~\citep{HauserLatombe,HauserIJRR11,vega2016asymptotically}.
%This is particularly cumbersome in domains with many variables that do not always affect each other.
%This also limits the number of possible samplers one can design as only three parameters $\bar{x}, \bar{u}, \bar{x}'$ are exposed. 
%$$\langle (\bar{x}, \bar{x}'), \{\}, (\bar{u}), $$
%$$\langle (\bar{x}, \bar{u}), \{\}, (\bar{x}'), $$
%$$\langle (\bar{x}), \{\}, (\bar{x}'), $$
%$$\langle (\bar{u}), \{\}, (\bar{x}'), $$
% The mode people can get away with a little bit more because they have split mode states

\begin{figure*}[ht]
\centering
\includegraphics[width=0.99\textwidth]{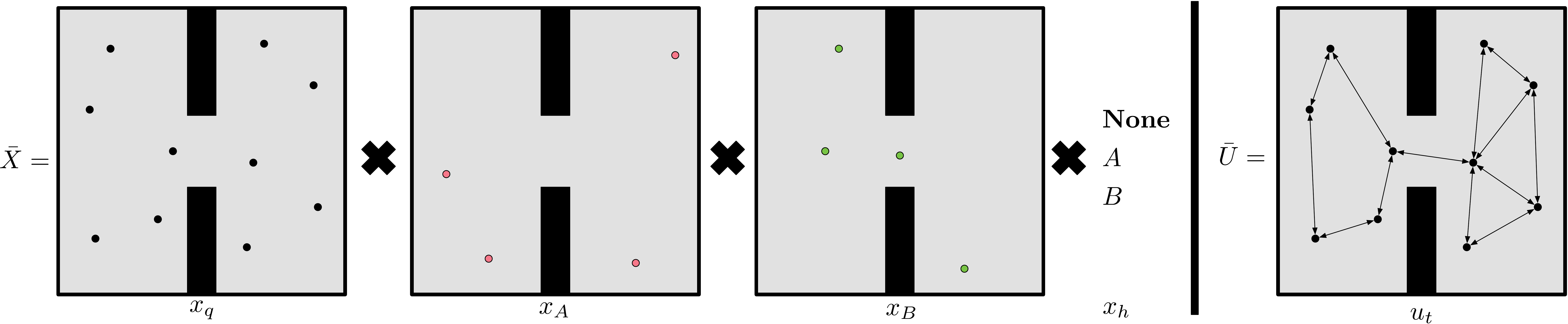}
\caption{A discretized state-space $\bar{X}$ and control-space $\bar{U}$ for a pick-and-place problem.} \label{fig:pp_samples}
\end{figure*}

Second, factoring can improve sample efficiency. 
A small set of samples for each state variable $X_i$ can lead to a large set of possible states $\bar{X} = X_1 \times ... \times X_n$ because of the decomposition of the state-space into domains of individual variables. 
Consider figure~\ref{fig:pp_samples} which displays 10 samples for $X_q$, 4 samples for $X_A$, and 4 samples for $X_B$. 
From only 16 total samples, we arrive at 160 possible arrangements of the robot and objects. 
While not all of these states are reachable, this large number of states presents more opportunities for solutions using these samples.
Factoring can improve sample efficiency with respect to control samples as well. 
Consider the roadmap of control trajectories in the right frame of figure~\ref{fig:pp_samples}.
The 26 trajectories (13 edges) result in possibly 208 $\id{Move}$ transitions due to $16$ possible arrangements of $A$ and $B$.
% 13*(4*4 + 4 + 4) = 312

\section{Algorithms}

\note{Tests here?}
\note{Eagerly apply tests like before}
\note{Focused algorithm can reason through things}

We have shown that, given a plan skeleton and sampling network, we can construct samplers that produce satisfying values. 
However, the input for a factored planning problem is just a transition system, initial set of states, and goal set of states. 
Thus, algorithms must search over plan skeletons and sampler sequences in order to produce solutions. 
%Many skeletons will not admit solutions due to 
%As in sampling-based motion planning, we are interested in identifying probabilistically complete algorithms. % that are efficient in practice.
%An algorithm is probabilistically complete for a domain, if for every robustly feasible problem within the domain, the algorithm will find a solution in finite time with high probability.
The concept of probabilistic completeness and the identification of probabilistically complete algorithms has been important for sampling-based motion planning.
We extend those ideas to sampling-based planning for factored transition systems.
\begin{defn}
An algorithm is {\em probabilistically complete} with respect to a domain $\langle {\cal D}, {\cal M} \rangle$ if for all robustly feasible problems ${\cal P} \in {\cal D}$, it will return a plan in finite time with probability one.
%it will almost surely return a plan in finite time.
\end{defn}

%$T$ is time
%$$\Pr[T \leq t] = \Pr[T = 1 \vee T = 2 \vee ...] = \sum_{i=1}^t\Pr[T = t]$$
% Expectation may not be finite if the $1 - \Pr[T \leq t]$ decays harmonically (not exponentially)
%Finite time is $\lim_{t \to \infty} \Pr[T \leq t]$
%$$\lim_{t \to \infty} $$
% Union of disjoint countable events equals the sum

%So far, we have focused on generating plan samples given a plan skeleton and sampling order for robustly feasible problems. However, 
%Thus, sampling-based factored planning can be viewed as a combined search over a plan skeleton and sampler order.
%Although this indicates that an algorithm can effectively sample a set of solutions for robustly feasible problems, the algorithm still must identify both the plan skeleton and the conditional sampler sequence. 
%Rather than design a special purpose algorithm for each domain, we present algorithms that take, as a hyper-parameter input, a set of conditional samplers for the domain. 
% This is like PRM and RRT, right?  Maybe say that?
%The algorithms are therefore {\em domain-independent} because the problem-specific knowledge is restricted to the constraints and samplers. 
%Rather than design a special purpose algorithm for each domain, 
We present algorithms that take as an input a set of conditional samplers $\Psi$ for the domain. % a hyper-parameter 
The algorithms are therefore {\em domain-independent} because the problem-specific knowledge is restricted to the constraints and samplers. 
%For these algorithms to work, the conditional samplers must satisfy the following properties within a domain.
We will show that these algorithms are probabilistically complete, given a set of sufficient conditional samplers $\Psi$ for conditional constraint manifolds ${\cal M}$.
%We will prove this by more generally showing that these algorithms will produce a solution if some sequence of sampling conditional samplers $\Psi$ produces the required samples.
Thus, the completeness of the resulting algorithm is entirely dependent on the conditionals samplers.
% appendix~\ref{appendix:dense}.
%%A naive algorithm for factored planning would perform a non-deterministic search over plan skeletons, sampler orders, and then samples. 
%A naive algorithm for factored planning iterates over and simultaneously samples for all plan skeletons. % and sampler orders.
%This can be seen as a search over plan structure that guides a search over continuous parameter values.
%However, this algorithm is clearly inefficient from both a combinatorial and sampling perspective. 
%It requires enumerating all plan skeletons up to the length of a solution plan skeleton.
%For typical applications of factored planning, the number of plans grows exponentially as, for instance, the number of objects in a manipulation domain increases. 
%%Possibly doubly exponentially...the length goes exponentially in the number of objects (kind of?) and num plans goes exponentially in length.
%% Something about sampling order
%Furthermore, it will sample each plan skeleton that, in a manipulation domain, will result in many expensive, unnecessary motion plan samples. 
%Finally, it will produce many redundant samples by keeping the samples for each plan skeleton independent from one another.
%% Once could also do replanning here (with all parameters) and attempt to fill things in
We give two algorithms, \proc{\incremental{}} and \proc{\focused{}}. 

\subsection{Incremental Algorithm}

\begin{figure}
\begin{footnotesize}
\begin{codebox}
\Procname{\proc{\incremental{}}(${\cal P}; \smplrSet$, \proc{discrete-search}):}
\li \elements{} = $\proc{initial-elements}({\cal P})$
\li \id{queue} = \proc{instantiate-samplers}(\elements{}; $\smplrSet$)
\li \While \kw{True}: \Do
\li $\langle \vec{a}, \vec{x}, \vec{u} \rangle$ = \proc{discrete-search}(${\cal P}$, \elements{})
\li \If $\vec{a} \neq \kw{None}$: \Then
\li \Return $\vec{u}$
\End
\li \id{processed}  = $\emptyset$
\li \proc{process-samplers}(\id{queue}, \id{processed}, \elements{};
\zi \;\;\;\;\;\;\;\;\;\;\;\;\;\;\;\;\;\;\;\;\;\;\;\;\;\;\;\;\;\;\;\;\proc{sample}, \kw{len}(\id{queue}))
\li \proc{push}(\id{queue}, \id{processed})
\End\End
\end{codebox}
\end{footnotesize}
\caption{The pseudocode for the \incremental{} algorithm.} \label{code:incremental}
\end{figure}

% Tomas - The creation of samplers from conditional samplers is crucial, so important to emphasize earlier.
The \incremental{} algorithm %performs a search over samples before plan structure.
alternates between generating samples and checking whether the current set of samples admits a solution.
It can be seen as a generalization of the PRM for motion planning and the iterative FFRob algorithm for task and motion planning~\citep{garrettIJRR2017} which both alternate between exhaustive sampling and search phases for their respective problem classes. 

\begin{figure}[h]
\begin{footnotesize}
%\begin{codebox}
%\Procname{\proc{elements}(${\cal C}, \bar{v}$):}
%%\li \Return $\{C(\bar{v}) \mid C \in {\cal C}, C \neq \id{Equal}\} \cup \{\id{Var}(v) \mid v \in \bar{v} \} $
%%\li \Return $\{C(\bar{v}) \mid C \in {\cal C}\} \cup \{\id{Var}(v) \mid v \in \bar{v} \}$ % Equal is fine
%\li \Return $\{C(\bar{v}) \mid C \in {\cal C}\}$ % This is also fine because it is used
%\end{codebox}
\begin{codebox}
\Procname{\proc{sample}($s$):}
\li $\psi(\bar{v}_I) = s; \langle I, O, {\cal C}, f \rangle = \psi$
\li $\bar{v}_O = \kw{next}(f(\bar{v}_I))$
\li \If $\bar{v}_O = \kw{None}$: \Then
\li \Return \{\}
\End
%\li \Return \proc{elements}(${\cal C}, \bar{v}_I + \bar{v}_O$)
\li \Return $\{C(\bar{v}_I + \bar{v}_O) \mid C \in {\cal C}\}$
\end{codebox}
\begin{codebox}
\Procname{\proc{instantiate-samplers}(\elements{}; $\Psi$):}
%\li \id{samples} = $\{v: [     ] \kw{for} \id{Var}_v e \in \Psi}$
\li \id{samples} = $\{v \mid C(\bar{v}) \in \elements{}, v \in \bar{v}\}$
\li \id{instances} = $\emptyset$
\li \For $\smplr = \langle I,  O, {\cal C}, f \rangle$ \kw{in} $\smplrSet$: \Do
\li \For $\bar{v}_I$ \kw{in} \kw{product}(\id{samples}, $|I|$): \Do
%\li \If \kw{all}($e$ \kw{in} \elements{} \For $e$ \kw{in} \proc{elements}(${\cal D}, \bar{v}_I$)): \Then
\li \id{instances} += $\{\smplr(\bar{v}_I)\}$
\End\End\End
\li \Return \id{instances}
\end{codebox}
\begin{codebox}
\Procname{\proc{process-samplers}(\id{queue}, \id{processed}, \elements{}; \proc{process}, $k$):}
\li \While (\kw{len}(\id{queue}) $\neq$ 0) \kw{and} (\kw{len}(\id{processed}) $< k$) : \Do
\li $s$ = \proc{pop}(\id{queue})
\li \elements{} += \proc{process}($s$)
\li \For $s'$ \kw{in} \proc{instantiate-samplers}(\elements{}; $\smplrSet$): \Do
\li \If $s'$ \kw{not} \kw{in} (\id{queue} + \id{processed}): \Then
\li \proc{push}(\id{queue}, $s'$)
\End\End
\li \id{processed} += $\{s\}$
\End
\end{codebox}
\end{footnotesize}
\caption{The pseudocode for the procedures shared by both the \incremental{} and \focused{} algorithms.} \label{code:shared}
\end{figure}

The pseudocode for the \incremental{} algorithm is displayed in figure~\ref{code:incremental}.
\proc{\incremental{}} updates a set \elements{} containing certified constraint elements $C(\bar{v}_P)$ where $C = \langle P, R \rangle$ and $\bar{v}_P \in R$.
 %tuples of values annotated with the constraint they satisfy, 
Intuitively, each constraint element $C(\bar{v}_P) \in \elements{}$ is a tuple of samples $\bar{v}_P$ that have been identified by \proc{\incremental{}} to satisfy constraint $C$.
As a result, \elements{} encodes the current discretization of the transition relation ${\cal T}$ corresponding to problem ${\cal P}$.
Namely, for a state and control triple $(\bar{x}, \bar{u}, \bar{x}') = \bar{v}$ and a clause ${\cal C}_a$, 
\begin{align*} %\label{eqn:implication}
\{C(\bar{v}_P) \mid C  &= \langle P, R \rangle \in {\cal C}_a \} \subseteq \elements{}  \nonumber \\
& \implies \bar{v} = (\bar{x}, \bar{u}, \bar{x}') \in {\cal T}.
\end{align*}
As \proc{\incremental{}} identifies new constraint elements and adds them to \id{elements}, it in turn identifies additional possible transitions.
%The addition constraint elements to \elements{} causes the set of identified transitions to increase.
\note{Rename discretized variables to use $\hat{X}$?}
\note{Could define a bunch of unreachable transitions from values we don't currently have}
%As a result, \elements{} identifies a finite set of transitions $(\bar{x}, \bar{u}, \bar{x'}) \in {\cal T}$ for problem ${\cal P}$.  

Additionally, \proc{\incremental{}} maintains \id{queue}, a first-in-first-out queue of sampler instances.
The procedure \proc{initial-elements} gives the set of elements resulting from initial samples present in problem ${\cal P}$.
%Recall that a constraint element is tuple of values annotated with the constraint they satisfy.
Define an {\em iteration} of \proc{\incremental{}} to be the set of commands in body of the \While loop.
On each iteration, \proc{\incremental{}} first calls \proc{discrete-search} to attempt to find a plan $\langle \vec{a}, \vec{x}, \vec{u} \rangle$ using \elements{}. 
%First, \proc{discrete-search} searches a discretized transition system for problem ${\cal P} = \langle {\cal C}_0, {\cal C}_*, \{{\cal C}_1, ..., {\cal C}_\alpha\} \rangle$, given \elements{}, a set of constraint elements on samples from each ${\cal X}_i$ and ${\cal U}_i$.
The procedure \proc{discrete-search} searches the current discretization of problem ${\cal P}$ given by \id{elements}.
We assume \proc{discrete-search} is any sound and complete discrete search algorithm such as a breadth-first search (BFS). 
%${\cal P} = \langle {\cal C}_0, {\cal C}_*, \{{\cal C}_1, ..., {\cal C}_\alpha\} \rangle$, using the samples from each ${\cal X}_p$ and ${\cal U}_p$ given by \elements{}.
%returns the sequence of clauses $\vec{a}$, states $\vec{x}$, and control inputs $\vec{u}$ that correspond to a plan.
We outline several possible implementations of \proc{discrete-search} in section~\ref{sec:search}.
%The simplest implementation explicitly constructs the discrete state-space and the discrete set of transitions.
%Then, it searches the resulting discrete graph using any sound and complete search algorithm such as breadth-first search (BFS).
%Advanced implementations instead use efficient artificial intelligence search algorithms that are able to exploit the factoring in the transition system through heuristics~\citep{helmert2006fast}.
%The procedure \proc{instances} produces the set of states and transitions formed from samples that also satisfy their respective conjunctive constraint clauses ${\cal C}$.
%, that operate more efficiently by considering the full pool of samples at once and using efficient search algorithms to identify solutions using only these sampled parameter values.s
If \proc{discrete-search} is successful, the sequence of control inputs $\vec{u}$ is returned. 
Otherwise, \proc{discrete-search} produces $\vec{a} = \kw{None}$.
In which case, \proc{\incremental{}} calls the \proc{process-samplers} subroutine to sample values from at most \kw{len}(\id{queue}) sampler instances using the function \proc{sample}. 
The procedure \proc{sample} in figure~\ref{code:shared} has a single sampler-instance argument $s$ and queries the \kw{next} set of output values $\bar{v}_O$ in the sequence $f(\bar{v}_I)$. 
%In the event that the sequence is enumerated, let $\proc{next}(f(\bar{v}_I)) = \kw{None}$.
If the sequence has not been enumerated, {\it i.e.} $\proc{next}(f(\bar{v}_I)) \neq \kw{None}$, it returns a set of constraint elements $C(\bar{v}_I + \bar{v}_O)$ that values $\bar{v}_I + \bar{v}_O$ satisfy together. % using the helper function \proc{elements}.
%The helper function \proc{elements} takes in a set of constraints ${\cal C}$ and a set of values $\bar{v}$ and returns the corresponding set of constraint elements, including elements for the implicit variable constraints $\id{Var}$.

The procedure \proc{process-samplers} iteratively instantiates and processes sampler instances $s$.
Its inputs are a \id{queue} of sampler instances, a set of already \id{processed} sampler instances, the set of constraint \elements{}, and two additional parameters that are used differently by \proc{\incremental{}} and \proc{\focused{}}: 
the procedure $\proc{process} \in \{\proc{sample}, \proc{sample-lazy}\}$ takes as input a sampler instance and returns a set of elements, and $k$ is the maximum number of sampler instances to process. 
On each iteration, \proc{process-samplers} pops a sampler instance $s$ off of \id{queue}, adds the result of \proc{process} to \elements{}, and adds $s$ to \id{processed}. 
Constraints specified using tests, such as collision constraints, are immediately evaluated upon receiving new values.
The procedure \proc{instantiate-samplers} produces the set of sampler instances of samplers $\Psi$ formed from constraint elements \elements{}. 
%A simple implementation of \proc{instantiate-samplers} first recovers the set of samples \id{samples} from \id{elements}.
For each sampler $\psi$, \proc{instantiate-samplers} creates a sampler instance $s = \psi(\bar{v}_I)$ for every combination of values $\bar{v}_I$ for $\psi$'s input parameters that satisfy $\psi$'s input variable domain constraints.
%A more efficient implementation of \proc{instantiate-samplers} can dynamically update \id{instances} as \id{elements} grows.
% and more efficiently perform \kw{product}.
New, unprocessed sampler instances $s'$ resulting from the produced \elements{} are added to \id{queue}. 
\proc{process-samplers} terminates after $k$ iterations or when \id{queue} is empty.
Afterwards, \proc{\incremental{}} adds the \id{processed} sampler instances back to \id{queue} to be used again on later iterations.
This ensures each sampler instance is revisited arbitrarily many times.

% Many of the problems are declared infeasible at the start
% One could dynamically update at the search tree. The incremental algorithm is only a linear factor worse than this at most (because it repeats the search each iteration)
% Mention relationship to brute force search in the space (Hauser).
%While not included in the pseudocode, \proc{\incremental{}} can identify infeasible problems by terminating if $\vec{a} = \kw{None}$ and $\kw{len}(\id{queue}) = 0$.

% STRIPS Planning in Infinite Domains
% https://arxiv.org/pdf/1701.00287.pdf

%Suppose that $\proc{sample}(\psi(\id{inps}))$ returns one value?

\note{How do I deal with constants in my analysis? I guess they could always be zero dimensional conditional manifolds?}

\begin{thm} \label{thm:eager}
\proc{\incremental{}} is probabilistically complete for a domain $\langle {\cal D}, {\cal M} \rangle$ given a sufficient set of conditional samplers for $\langle {\cal D}, {\cal M} \rangle$. 
\begin{proof}
Consider any robustly feasible problem ${\cal P} \in {\cal D}$. 
By definitions 6 and 7, there exists a sampler sequence $\vec{\psi} = (\psi_1, ..., \psi_k)$ that with probability one, a finite number of calls to \proc{sample} produces values that are parameters in ${\cal C}_{\vec{a}}$ for some robustly satisfiable plan skeleton $\vec{a}$. 
In its initialization, \proc{\incremental{}} adds all sampler instances $s$ available from the ${\cal P}$'s constants.
On each iteration, \proc{\incremental{}} performs $\proc{sample}(s)$ for each sampler instance $s$ in \id{queue} at the start of the iteration. % to generate a new value. 
There are a finite number of calls to \proc{sample} each iteration.
The resulting constraints elements $\proc{sample}(s)$ are added to \elements{} and all new sampler instances $s'$ are added to \id{queue} to be later sampled. 
The output values from each sampler instance will be later become input values for all other appropriate conditional samplers. 
This process will indirectly sample all appropriate sampler sequences including $\vec{\psi}$. 
%Starting with $\psi_1(())$ each sampler 
Moreover, because $s$ is re-added to \id{queue}, it will be revisited on each iteration.
Thus, $\proc{sample}(s)$ will be computed until a solution is found.
%By adding $\psi(\id{inps})$ to the back of the queue, the number of sampler and input pairs before it decreases in the queue before it is finite. 
Therefore, each sampler sequence will also be sampled not only once but arbitrarily many times. 
%\proc{\incremental{}} will produce satisfying values from $\vec{\psi}$ if they are available after a finite number of calls. 
\proc{\incremental{}} will produce satisfying values from $\vec{\psi}$ within a finite number of iterations. 
%And again, they are assumed to be available with probability one.

%We will assume \proc{discrete-search} is any sound and complete discrete search algorithm. % such a breadth-first search (BFS). 
Because \proc{discrete-search} is assumed to be sound and complete, \proc{discrete-search} will run in finite time and return a correct plan if one exists. 
On the first iteration in which a solution exists within \id{samples}, \proc{discrete-search} will produce a plan $\vec{a} \neq \kw{None}$.
And \proc{\incremental{}} will itself return the corresponding sequence of control inputs $\vec{u}$ as a solution.
\qed

%%%%%%%%%%

%As shown in theorem 1, for any robustly feasibly problem within the domain, with probability one, there exists a finite target sequence of sampler calls $\proc{sample}(\phi(z_P))$ which will produce the parameters for a solution. A required sample may be far down in the sequence of $\phi(z_P)$, in which case $\proc{sample}(\phi(z_P))$ will need to be called many times on the target sequence.
%\proc{\incremental{}} will iteratively try all compositions of conditional samplers. It constructs new sampler instances from conditional samplers called with all possible compatible existing samples. It maintains the set of all sampler instances in {\it queue} to ensure that each can be called arbitrarily many times. Thus, if outcomes of the conditional sampler sequences admit a solution, it eventually produce those samples. On the subsequent iteration, \proc{search} will find a solution assuming it is complete.
%%Thus, for robustly feasible problems and a sufficient set of conditional samplers, within a finite amount of time (with high probability) the composition of samplers will produce a sample in the nonzero solution measure set of solutions on the witness composed constraint manifold. 
\end{proof}
\end{thm}

\note{In practice can stop early}

Because \proc{\incremental{}} creates sampler instances exhaustively, it will produce many unnecessary samples. 
This results the combinatorial growth in the number of queued sampler instances as well as the size of the discretized state-space.
This motivates our second algorithm, which is able to guide the selection of samplers by integrating the search over structure and search over samples.
% Exponential growth in samples

\subsection{Focused Algorithm}

% Could describe as repeatedly calling the incremental algorithm actually

The \proc{\focused{}} algorithm %performs a combined search over plan structure and samples. 
uses {\em lazy samples} as placeholders for actual concrete sample values. 
Lazy samples are similar in spirit to symbolic references~\citep{Srivastava14}.
% which can be used in a plan and used to guide later sampling
%representing a possible output of each conditional sampler.
The lazy samples are optimistically assumed to satisfy constraints with concrete samples and other lazy samples via {\em lazy constraint elements}.
Lazy constraint elements produce an optimistic discretization of ${\cal T}$, characterizing transitions that may exist for some concrete binding of the lazy samples involved.
This allows \proc{discrete-search} to reason about plan skeletons with some free parameters. % without some concrete parameters
After finding a plan, \proc{\focused{}} calls samplers that can produce values for the lazy samples used.
As a result, \proc{\focused{}} is particularly efficient on easy problems in which a large set of samples satisfy each constraint.
This algorithm is related to a lazy PRM~\citep{bohlin2000path,dellin2016unifying}, which defers collision checks until a path is found. 
%lazy shortest path algorithm
However instead of just defering collision checks, \proc{\focused{}} defers generation of sample values until an optimistic plan is found. 
In a pick-and-place application, this means lazily sampling poses, inverse kinematic solutions, and trajectories. 
Because \proc{\focused{}} plans using both lazy samples and concrete samples, it is able to construct plans that respect constraints on the actual sample values. 
And by using lazy samples, it can indicate the need to produce new samples when the existing samples are insufficient. 

\note{Epoch instead of episode}
The pseudocode for the \focused{} algorithm is shown in figure~\ref{code:focused}.
The \focused{} algorithm uses the same subroutines in figure~\ref{code:shared} as the \incremental{} algorithm.
Once again, let an {\em iteration} of \proc{\focused{}} be the set of commands in body of the \While loop.
Define an {\em episode} of \proc{\focused{}} to be the set of iterations between the last \id{sampled} reset and the next \id{sampled} reset. 
Let the initialization of \id{sampled} in line 1 also be a reset.
On each iteration, the \proc{\focused{}} algorithm creates a new \id{queue} and calls \proc{process-samplers} to produce \mixed{}. It passes the procedure \proc{sample-lazy} rather than \proc{sample} in to \proc{process-samplers}. 
For each output $o$ of $\smplr$, \proc{sample-lazy} creates a unique lazy sample $l_o^\smplr$ for the combination of $\smplr$ and $o$. 
Then, for each lazy constraint element $e$ formed using $l_o^\smplr$, $s$ is recorded as the sampler instance that produces values satisfying the element using $e.\id{instance}$.
For a pose sampler instance $\psi_P^o()$, \proc{sample-lazy} creates a single lazy sample $\lzp{1}$ and returns a single lazy constraint element $\id{Stable}(\lzp{1})$:
\begin{equation*}
\proc{lazy-sample}(\psi_P()) = \{\id{Stable}(\lzp{1})\}.
\end{equation*}

\note{Just do the DAG version of this, so no holdout or nonunique samples?}

%The inputs \id{inp} may be lazy samples themselves. 
%In order to avoid producing an infinite number of lazy samples, $\id{out}$ becomes shared across sampler inputs after a fixed depth.
% Do versions with multiple levels

% Separate generic parameter per the output of each sampler
% In practice, I choose multiple ways of achieving things, but this time I will just use one this time
% I'm just going to ignore the whole sampler domain stuff for now...
% Well actually, there will be a separate sampler for each block (i.e. different variable) which is nice 
% Need back pointer to parameters...
\begin{figure}
\begin{footnotesize}
\begin{codebox}
\Procname{\proc{sample-lazy}($s$):}
%\li $\bar{z}_O = (\proc{LazySample}(s, o) \mid o \in O)$
%\li $\bar{z}_O = (\proc{LazySample}_\Psi(s, o) \mid o \in O)$
%\li $\bar{v}_O = (l_o^\psi \mid o \in O)$
%\li \For $l_o^\psi$ \kw{in} $\bar{v}_O$: \Do
%\li \If $l_o^\psi$.\id{instance} = \kw{None}: \Then
%\li $l_o^\psi.\id{instance} = s$
%\End\End
\li $\psi(\bar{v}_I) = s; \langle I, O, {\cal C}, f \rangle = \psi$
\li $\bar{l}^\psi_O = (l_o^\psi \mid o \in O)$
%\li $\id{lazy\_elements} = \proc{elements}({\cal C}, \bar{v}_I + \bar{l}^\psi_O$)
\li $\id{lazy\_elements} = \{C(\bar{v}_I + \bar{l}^\psi_O) \mid C \in {\cal C}\}$
\li \For $e \in \id{lazy\_elements}$: \Do
\li $e.\id{instance} = s$
\End
\li \Return $\id{lazy\_elements}$
%\li \Return $\proc{elements}({\cal C}, \bar{v}_I + \bar{v}_O$)
\end{codebox}
\begin{codebox}
\Procname{\proc{retrace-instances}(\id{target\_elements}, \id{elements}):}
\li \id{instances} = $\emptyset$
\li \For $e$ \kw{in} $(\id{target\_elements} \setminus \id{elements})$: \Do
\li $\psi(\bar{v}_I) = e.instance$
%\li \id{ancestors} = \proc{retrace-instances}($\proc{elements}(\psi.{\cal D}, \bar{v}_I)$)
%\li \id{ancestors} = \proc{retrace-instances}($\proc{elements}(\emptyset, \bar{v}_I)$)
%\li \id{ancestors} = \proc{retrace-instances}($\{\id{Var}(\bar{v}_i) \mid i \in I\}$)
\li \id{ancestors} = \proc{retrace-instances}($\{\id{Var}(\bar{v}) \mid \bar{v} \in \bar{v}_I\}$)
\li \If \id{ancestors} = $\emptyset$: \Do
\li \id{instances} += $\{\psi(\bar{v}_I)\}$
\End
\li \id{instances} += \id{ancestors}
\End
\li \Return \id{instances}
\End
\end{codebox}

\begin{codebox}
\Procname{\proc{\focused{}}(${\cal P}; \smplrSet$, \proc{discrete-search}):} 
\li \elements{} = $\proc{initial-elements}({\cal P})$
\li \id{new\_elements} = $\emptyset$; \id{sampled} = $\emptyset$
\li \While \kw{True}: \Do
\li \id{queue} = \proc{instantiate-samplers}(\elements{}; $\smplrSet$)
\li \mixed{} = \kw{copy}(\elements{})
\li \proc{process-samplers}(\id{queue}, \kw{copy}(\id{sampled}), \mixed{}; 
\zi \;\;\;\;\;\;\;\;\;\;\;\;\;\;\;\;\;\;\;\;\;\;\;\;\;\;\;\;\;\;\;\;\proc{sample-lazy}, $\infty$)
\li $\langle \vec{a}, \vec{x}, \vec{u} \rangle$ = \proc{discrete-search}(${\cal P}$, \mixed{})
\li \If $\vec{a} = \kw{None}$: \Then
\li \elements{} += \id{new\_elements}
\li \id{new\_elements} = $\emptyset$; \id{sampled} = $\emptyset$
\li \kw{continue}
\End
%\li \id{plan\_elements} = \proc{elements}(${\cal C}_{\vec{a}}, \vec{x}  + \vec{u})$
\li \id{plan\_elements} = $\{C(\vec{x}  + \vec{u}) \mid C \in {\cal C}_{\vec{a}}\}$
\li \If $\id{plan\_elements} \subseteq \elements{}$: \Then
\li \Return $\vec{u}$
\End
\li \For $s$ \kw{in} \proc{retrace-instances}(\id{plan\_elements}, \elements{}): \Then
%\li \If \id{inps} $\subseteq$ \elements{}: \Then
%\li \If \kw{all}($e$ \kw{in} \elements{} \For $e$ \kw{in} \proc{elements}(${\cal D}, \bar{z}_I$)): \Then
\li \id{new\_elements} += \proc{sample}($s$)
\li \id{sampled} += $\{s\}$
\End\End
\end{codebox}
\end{footnotesize}
\caption{The pseudocode for the \focused{} algorithm.} \label{code:focused}
\end{figure}
% I could make a bunch of lazy samples which specifically are in combination with particular objects. This would likely be pretty expensive though
% Will any ordering of \proc{sample-order} work? No, it may construct a space with bad measure
% Make these tied to abstract objects rather than tests because will evaluate all tests greedily

% Generator is more generic than sampler because it more clearly has state
\proc{discrete-search} performs its search using \mixed{}, a mixed set of \elements{} and \id{lazy\_elements}.
If \proc{discrete-search} returns a plan, \proc{\focused{}} first checks whether it does not require any \id{lazy\_elements}, in which case it returns the sequence of control inputs $\bar{u}$. 
Otherwise, it calls \proc{retrace-instances} to recursively extract the set of sampler instances used to produce the \id{lazy\_elements}.
\proc{retrace-instances} returns just the set of ancestor sampler instances that do not contain lazy samples in their inputs.
For each ancestor sampler instance $s$, \proc{\focused{}} samples new output values and adds any new constraint elements to \id{new\_elements}.
To ensure all relevant sampler instances are fairly sampled, each $s$ is then added to \id{sampled}. 
This prevents these sampler instances from constructing lazy samples within \proc{process-samplers} on subsequent iterations.
%Future invocations of \proc{discrete-search} will not be able to return plans dependent on these parameters. 
Additionally, elements are added to \id{new\_elements} before they are moved to \elements{} to limit the growth in sampler instances. 
When \proc{discrete-search} fails to find a plan, \id{new\_samples} are added to \elements{}, \id{sampled} is reset, and this process repeats on the next episode. 

%These restrictions are necessary for completeness because a sampler may need to be called arbitrarily many times and there are infinitely many samplers that could be formed from conditional samplers.

While not displayed in the pseudocode, the \focused{} algorithm has the capacity to identify infeasibility for some problems.
When \proc{discrete-search} fails to find a plan and \id{sampled} is empty, the problem is infeasible because the discretized problem with optimistic lazy samples is infeasible.
%In practice, we directly add \id{new\_elements} to \elements{} for a fixed number of iterations before temporarily placing them in \id{new\_samples}.
If no graph on conditional samplers $\Psi$ contains cycles, then a lazy sample can be created for each sampler instance rather than each sampler.
Then, \proc{retrace-instances} can sample values for lazy elements that depend on the values of other lazy elements. 
Finally, \id{new\_elements} can be safely added directly to \elements{}.
These modifications can speed up planning time by requiring fewer calls to \proc{solve-discrete}.
For satisficing planning, to bias \proc{discrete-search} to use few lazy samples, we add a non-negative cost to each transition instance corresponding to the number of lazy samples used and use a cost sensitive version of \proc{solve-discrete}.
In this context \proc{solve-discrete} can be thought of optimizing for a plan that requires the least amount of additional sampler effort.
Thus, plans without lazy samples have low cost while plans with many lazy samples have high cost.
% Discuss the MDP story

%\begin{lem} \label{lem:lazy}
%During each episode, the \proc{\focused{}} algorithm will \proc{sample} 
%\begin{proof}
%\end{proof}
%\end{lem}

\begin{thm}
\proc{\focused{}} is probabilistically complete for a domain $\langle {\cal D}, {\cal M} \rangle$ given a sufficient set of conditional samplers for $\langle {\cal D}, {\cal M} \rangle$. 
\begin{proof}
As in theorem~\ref{thm:eager}, consider any robustly feasible problem ${\cal P} \in {\cal D}$. 
By definitions~\ref{defn:sufficient} and~\ref{defn:sufficient-domain}, there exists a sampler sequence $\vec{\psi} = (\psi_1, ..., \psi_k)$ that with probability one, in a finite number of calls to \proc{sample} produces values that are parameters in ${\cal C}_{\vec{a}}$ for some robustly satisfiable plan skeleton $\vec{a}$. 
At the start of an episode, \elements{} implicitly represents a set of partially computed sampler sequences. 
We will show that between each episode, for each partially computed sampler sequence that corresponds to some plan skeleton, a next sampler in the sampler sequence will be called.
And both the new partial sampler sequence as well as the old one will be present within \elements{} during the next episode.
% This is because it is partially ordered

On each iteration, \proc{\focused{}} calls \proc{discrete-search} to find a plan that uses both real samples and lazy samples. 
%In the event that a plan is found using only \id{samples} (line 10), \proc{\focused{}} returns it as a solution. %Otherwise, 
\proc{\focused{}} calls \proc{sample} for each sampler instance $s$ corresponding to a lazy sample along the plan. 
Additionally, by adding $s$ to \id{sampled}, \proc{\focused{}} prevents the lazy samples resulting from $s$ from being used for any future iteration within this episode. 
This also prevents \proc{discrete-search} from returning the same plan for any future iteration in this episode. 
The set \elements{} is fixed for each episode because new samples are added to \id{new\_elements} rather than \elements{}. 
Thus, there are a finite number of plans possible within each episode. 
And the number of iterations within the episode is upper bounded by the initial number of plans. 
Each plan will either be returned on some iteration within the episode and then be blocked or it will be incidentally blocked when \proc{search} returns another plan. 
Either way, \proc{sample} will be called for a sampler instance $s$ on its remaining sampler sequence. 
When no plans remain, \proc{search} will fail to find a plan.
Then, \proc{\focused{}} resets, allowing each $s \in \id{sampled}$ to be used again, and the next episode begins.
% The set of plans stricly increases

Because each episode calls \proc{sample} for at least one $\psi$ along each possible partial sampler sequence, $\vec{\psi}$ will be fully sampled once after at most $k$ episodes. 
Moreover, each subsequent episode will sample $\vec{\psi}$ again as new partial sampler sequences are fully computed. 
Thus, $\vec{\psi}$ will be fully sampled arbitrarily many times.
%As previous indicated, each episode has a finite duration 
Consider the first episode in which a solution exists within \elements{}. 
\proc{discrete-search} is guaranteed to return a plan only using only \elements{} within this episode.
This will happen, at latest, when all plans using lazy elements are blocked by \id{sampled}.
Then, \proc{\focused{}} will itself return the corresponding sequence of control inputs as a solution. 
\qed

\end{proof}
\end{thm}

It is possible to merge the behaviors of the \incremental{} and \focused{} algorithms and toggle whether to eagerly or lazily \proc{sample} per conditional sampler.
This allows inexpensive conditional samplers to be immediately evaluated while deferring sampling of expensive conditional samplers.
This fusion leads to a variant of the \proc{\focused{}} algorithm where sampler instances switch from being lazily evaluated to eagerly evaluated when they are added to \id{sampled}.
In this case, \id{sampled} need not be reset upon \proc{discrete-search} failing to identify a plan.

\section{Discrete Search} \label{sec:search}

%As a hyper-parameter, \proc{discrete-search} requires a blackbox search procedure \proc{search} to find plans within the discretized problem. 
%Unlike many approaches for task and motion planning, the discrete search operates directly over raw samples such as configurations, poses, and trajectories. % rather than special purpose symbols.
% A unguided search in many high-dimensional state-spaces is prohibitively expensive.
% Brute force search is likely all that's available in low dimensional systems that aren't factorable, but for high-dimensional systems it is bad.

The procedure \proc{discrete-search} takes as input a factored transition problem ${\cal P}$ and a set of constraint elements \id{elements}.
%It returns a plan $\lange \vec{a}, \vec{x}, \vec{u} \rangle$ if it finds a plan
The set of constraint elements \id{elements} is used to derive \id{transitions}, a discretization of transition relation ${\cal T}$ for problem ${\cal P}$.
This is done by first extracting the discretized variable domain $Z_p$ for each parameter index $p$:
\begin{equation*}
%Z_p = \{v \mid \exists\; C(\bar{z}_P) \in \elements{}.\; p \in P, v = \bar{z}_p\}.
Z_p = \{\bar{v}_p \mid \exists\; C(\bar{v}_P) \in \elements{}.\; p \in P\}.
\end{equation*}
The discretized variable domains result a discretized state-space $\bar{X} = X_1 \times ... \times X_m$ and control-space $\bar{U} = U_1 \times ... \times U_n$.
% The discretized variable domain $Z_p$ comprising $\bar{X} = X_1 \times ... \times X_m, \bar{U} = U_1 \times ... \times U_n$ is also derived from \elements{}:
The discretized set of transitions is then
\begin{align*}
\id{transitions} &= \{(\bar{x}, \bar{u}, \bar{x}') \in \bar{X} \times \bar{Z} \times \bar{X} \mid \exists\; {\cal C}_a.\; \\
\forall\; C &= \langle P, R \rangle \in {\cal C}_a.\; C(\bar{z}_P) \in \elements{}\}.
\end{align*}

%As shown in equation~\label{eqn:implication}, 
% Explain hashing?
A straightforward implementation of \proc{discrete-search} is a breadth-first search (BFS) from $\bar{x}_0$ using \id{transitions} to define the set of directed edges $\{(\bar{x}, \bar{x}') \mid (\bar{x}, \bar{u}, \bar{x}') \in \id{transitions}\}$ defined on vertices $\bar{X}$. 
Note that the control samples $\bar{u}$ are used to identify transitions, but play no role in the BFS itself.
% the current discretization of the transition relation ${\cal T}$ derived from \elements{}:
% Transitions as edges
As an optimization, the outgoing edges from a state $\bar{x}$ can be dynamically computed by considering each clause ${\cal C}_a$, substituting the current values for $\bar{x}$, and identifying all combinations of $\bar{u}$ and $\bar{x}'$ resulting in $(\bar{x}, \bar{u}, \bar{x}') \in \id{transitions}$.
%\begin{equation} \label{eqn:elements}
%%\proc{elements}({\cal C}_a, \bar{x} + \bar{u} + \bar{x}') \subseteq \elements{}.
%%\{C(\bar{x} + \bar{u} + \bar{x}') \mid C \in {\cal C}_a\} \subseteq \elements{}.
%\{C(\bar{x} + \bar{u} + \bar{x}') \mid C \in {\cal C}_a\} \subseteq \elements{}.
%\end{equation} 
%The resulting state is $\bar{x}'$. 
%The control variables $\bar{u}$ are only used in the resulting plan. % to affirm the legality of a transition.
%Control variables are only used to determine the validity of an edge.
This can be further optimized by fixing the values of any $\bar{u}$, $\bar{x}'$ constrained by equality. 
%Thus, the number of outgoing edges is bounded by the number of combinations of sampled values for each free control or subsequent state variable.
While a BFS avoids explicitly constructing the full discretized state-space, it will still search the entire state-space reachable from $\bar{x}_0$ in fewer transitions than the length of the shortest plan. 
This can be prohibitively expensive for problems with significant factoring such as pick-and-place problems where many choices of objects to manipulate result in a large branching factor.

\subsection{Factored Planning}

% Artificial intelligence planning algorithms are 
The artificial intelligence community has developed many algorithms that are much more efficient than classical graph search algorithms for high-dimensional, factored problems. 
These algorithms exploit both the factored state representation and transitions with many equality constraints to guide search using domain-independent heuristics.
% sparsity of transition effects
Many heuristics are derived by solving an easier approximation of the original search problem.
This leads to both admissible~\citep{bonet2001planning} and empirically effective heuristics~\citep{HoffmannN01,helmert2006fast}.
These heuristics can frequently avoid exploring most of the discrete state-space and even efficiently identify many infeasible problem instances.

In our experiments, we use the efficient FastDownward planning toolkit~\citep{helmert2006fast} which contains implementations of many of these algorithms.
FastDownward, as well as many other planners, operate on states described as a finite set of discrete variables.
For example, the foundational STRIPS~\cite{Fikes71} planning formalism uses binary variables in the form of logical propositions.
We instead consider the Simplified Action Structures (SAS+)~\citep{backstrom1995complexity} planning formalism which allows variables with arbitrary finite domains.
This allows a factored transition system state $\bar{x}$ to also be a legal SAS+ state
%Because the current set of constraint elements \id{elements} involves a finite set of samples, 
where each state variable $x_p$ has a finite discretized domain $X_p$. 
%This simplifies the transformation from a factored transition problem ${\cal P}$ and set of constraint elements $\id{elements}$ to a SAS+
%Control variables $u_p$ also have finite discretized domains; however, they are solely used when grounding actions.
Discrete transitions in SAS+ are described using a precondition and effect action model.
% http://ai.cs.unibas.ch/_files/teaching/fs16/ai/slides/ai34.pdf
\begin{defn}
An {\em action} $\langle \id{pre}, \id{eff}\rangle$ is given by sets of constant equality conditions \id{pre} on $\bar{x}$ and \id{eff} on $\bar{x}'$. 
State variables $i$ omitted from \id{eff} are assumed to be constrained by pairwise equality constraints $x_i = x_i'$. 
\end{defn}
A single action will represent many different transitions if some state variables are not mentioned within $\id{pre}$.
Thus, action models can be advantageous because they compactly describe many transitions using a small set of actions. 
%This is particular evident when transitions 
%that are valid for any value of many variables (few variables mentioned within \id{pre}) and change only a small number of variables (few variables mentioned within \id{eff}). 

In order to use these algorithms, we automatically compile 
%discretized factored transition systems given by a a factored transition problem ${\cal P}$ and set of constraint elements $\id{elements}$ 
\id{transitions} into SAS+.
We could instead automatically compile to Planning Domain Definition Language (PDDL)~\citep{mcdermott1998pddl,edelkamp2004pddl2}, a standardized artificial intelligence planning language used in competitions. 
However, many PDDL planners first compile problem instances into a formalism similar to SAS+~\citep{helmert2006fast}, so we directly use to this representation.
%Our compilation uses derived predicates~\citep{edelkamp2004pddl2} to compactly factor transitions with many constraints.
%We compile the transition clauses into a set of STRIPS actions and compile each sampler and test into a stream. 

\subsection{Action Compilation}

A factored transition system with a discretized set of constraint elements can be compiled into SAS+ as follows.
First, the goal constraints ${\cal C}_*$ are converted into a `goal transition'
\begin{equation*}
{\cal C}_* \cup \{x'_\id{goal} = \kw{True}\} \cup \{x_1 = x_1', ..., x_m = x_m'\}
\end{equation*}
by adding a state variable $x_\id{goal}$ that is true when the goal constraints have been satisfied.
This allows the new goal ${\cal C}_*' = \{x'_\id{goal} = \kw{True}\}$ to be represented with a single equality constraint.
Because of this additional state variable, all existing transitions are augmented with an equality constraint $\{x_\id{goal} = x'_\id{goal}\}$.

%Then, a set of transitions is generated for each clause ${\cal C}_a$
% Lifted?
Each clause ${\cal C}_a$ is compiled into a set of actions by first identifying its set of possible parameters $P_a$, which is comprised of each $x_i, u_j$ present within $C \in {\cal C}_a$ as well as all of $\bar{x}'$.
The inclusion of the entirely of $\bar{x}'$ reflects that, after applying an action, each state variable may change.
Many clauses ${\cal C}_a$ contain constant and pairwise equality constraints that fully constrain some parameters.
Thus, the subset of free parameters $F_a \subseteq P_a$ is determined by defining a graph on parameters and samples where undirected edges are pairwise equality constraints.
Connected components in the graph are parameters and samples that are transitively constrained by equality.
For each connected component that does not contain sample, a single parameter $f \in P_a$ is selected to represent the component. 
%include a single variable for each connected component of variables transitively constrained by pairwise equality. 
%$F_a$ excludes variables transitively constrained by constant equality.
Finally, the clause is grounded by considering every binding of the free parameters $F_a$ satisfying ${\cal C}_a$.
This creates ground action $\langle \id{pre}, \id{eff}\rangle$ for each binding where 
$\id{pre}$ contains an equality constraint from each $x_i \in P_a$ to the bound value of its corresponding free parameter $f \in F_a$ and 
$\id{eff}$ contains an equality constraint from each $x_i'$ to the bound value of its corresponding free parameter $f \in F_a$ if $[x_i = x_i'] \notin {\cal C}_a$.
% I could also just directly take the Cartesian without worrying about equality and check that equality is satisfied
Consider the following actions generated for ${\cal C}_\id{Pick}^o$:
%again considering all combinations of free $\bar{x}, \bar{u}, \bar{x}'$ resulting in constraint elements included within \elements{}. Finally, each transition is converted into an action by dropping pairwise equality constraints.
%\begin{align*}
%{\cal C}_\id{Move}:& \big\{\langle \{x_q = q\}, \{x_q' = q'\} \rangle  \mid \exists q, t, q'.\;  \\
%&\{\id{Motion}(q, t, q'), \id{CFree}(t)\} \subseteq \elements{}\big\}
%\end{align*}
\begin{align*}
\big\{\langle &\id{pre}=\{x_o = p, x_h =  \kw{None}, x_q = q\}, \\
&\id{eff}=\{x_o = g, x_h = o\} \rangle \mid \exists g, p, q.\; \\
&\id{Kin}_o(g, p, q) \in \elements{}\big\}.
\end{align*}
${\cal C}_\id{Pick}^o$ and ${\cal C}_\id{Place}^o$ only have $3$ free parameters, so $F_\id{Pick}^o = \{p, g, q\}$.
This results in a compact descriptions of their transitions. Now consider the actions generated for ${\cal C}_{\id{Move}}$:
\begin{align*}
\big\{\langle &\id{pre}=\{x_q = q, x_h =  \kw{None}\} \cup \{x_o = p_o \mid o \in {\cal O}\}, \\
&\id{eff}=\{x_q' = q'\} \rangle \mid \exists q, t, q', p_1, .., p_{|{\cal O}|} .\; \\
&\{\id{Motion}(q, t, q'), \id{CFree}(t)\} \;\cup \\
&\{\id{CFree}_o(t, p_o) \mid o \in {\cal O}\} \subseteq \elements{}\big\}
\end{align*}
${\cal C}_{\id{Move}}$ and ${\cal C}_\id{MoveH}^o$ have $3 + |{\cal O}|$ free parameters because $F_\id{Move} = \{q, t, q', p_1, .., p_{|{\cal O}|}\}$. 
For non-unary discretization of each object variable, the number of transitions grows exponentially in $|{\cal O}|$.
Despite this, each $\id{eff}$ only involves one variable and there is only one control variable.
The rest of the state variables are solely used to determine action feasibility.
Additionally, each constraint has low arity: \id{Motion} involves 3 parameters and each $\id{CFree}_o$ constraint only involves 2 parameters.

\subsection{Axiom Compilation}

Low constraint arity allows us to further factor transitions by introducing {\em derived variables}~\citep{edelkamp2004pddl2}, variables evaluated from the core state variables $\bar{x}$ using rules known as {\em axioms}~\citep{helmert2006fast}. 
Axioms are known to be useful for compactly expressing planning problems~\citep{thiebaux2005defense}. 
Axioms have the same form $\langle \id{pre}, \id{eff}\rangle$ as actions.
However, they are automatically applied upon reaching a new state in contrast to actions, which are chosen by a planner.

For each non-equality constraint $C = \langle P, R \rangle \in {\cal C}_a$, we compute a parameterized boolean derived variable $d_C(\bar{z}_D)$.
The parameterization $D = P \setminus \bar{x}$ includes parameters for the control $\bar{u}$ and subsequent state $\bar{x}'$ but excludes the current state variables $\bar{x}$. 
Note that $\bar{u}$ and $\bar{x}'$ are included within $D$ to ensure that same control and subsequent state values are considered in each derived variable precondition. % $d_C(\bar{z}_D) = \kw{True}$.
An axiom $\langle \id{pre}, \id{eff}\rangle$ is computed for each constraint element $C(\bar{v}) \in \elements{}$ where $\id{pre} = \{p : v_p \mid p \in (P \cap \bar{x})\}$ and $\id{eff} = \{d_C(\bar{v}_D): \kw{True}\}$.
This allows $d_C(\bar{z}_D) = \kw{True}$ to be substituted for $C$ within the preconditions of any action involving $C$. % in return for dropping  from ${\cal C}_a$.
%The axioms serve the purpose of quantifying out $\bar{x}$ from the action forms of ${\cal C}_a$, removing them as parameters.
Using this substitution, an action is performable if, for each constraint $C$, the values of state variables $P \cap \bar{x}$ complete a constraint element within \elements{}.
As a result, $\bar{x}$ can be removed from the possible parameters $P_a$ of ${\cal C}_a$.
% by individually testing if each constraint is satisfied.
Because each $P_a$ now involves fewer parameters, the set of resulting actions and axioms instances is generally much smaller than before.

The axioms computed for \id{Motion} are
\begin{align*}
\big\{\langle&\id{pre}=\{x_q = q\}, \\
&\id{eff}=\{\id{Motion}(\cdot, t, q') = \kw{True}\} \rangle \mid \exists q, t, q'. \\
&\id{Motion}(q, t, q')  \in \elements{}\big\}.
\end{align*}
The axioms computed for $\id{CFree}_o$ are %within ${\cal C}_\id{Move}$ are as follows
\begin{align*}
\big\{&\id{pre}=\langle \{x_o = p\}, \\
&\id{eff}=\{\id{CFree}_o(t, \cdot) = \kw{True}\} \rangle \mid \exists t, p. \\
&\id{CFree}_o(t, p)  \in \elements{}\big\}.
\end{align*}
And ${\cal C}_{\id{Move}}$ can be modified to be the following:
\begin{align*} 
\big\{\langle&\id{pre}=\{\id{Motion}(\cdot, t, q') = \kw{True}, x_h =  \kw{None}\} \\ 
&\;\;\;\;\;\;\;\;\;\;\;\;\cup \{\id{CFree}_o(t, \cdot): \kw{True} \mid o \in {\cal O}\}, \\
&\id{eff}=\{x_q' = q\} \rangle \mid \exists t, q'.\; \\
&\{\id{Var}(t), \id{Var}(q')\} \subseteq \elements{}\big\}.
\end{align*}
The resulting \id{Motion} and $\id{CFree}_o$ axioms as well as ${\cal C}_{\id{Move}}$ actions all have 3 or fewer parameters.
And the number of actions and axioms need to describe a pick-and-place transition is linear in $|{\cal O}|$ rather than exponential in $|{\cal O}|$. 
% Low airity
%$O(|{\cal O}|*2^{|{\cal O}|})$
%A finite set of samples ${\cal X}_i$ defines a finite state-space
%Do I want to more formally keep track of the samples? ${\cal X}$
% Empirical counts of the number of operators for each of these}

\section{Tabletop Manipulation}~\label{sec:manip}

\begin{figure*}[h]
\centering
\includegraphics[width=0.99\textwidth]{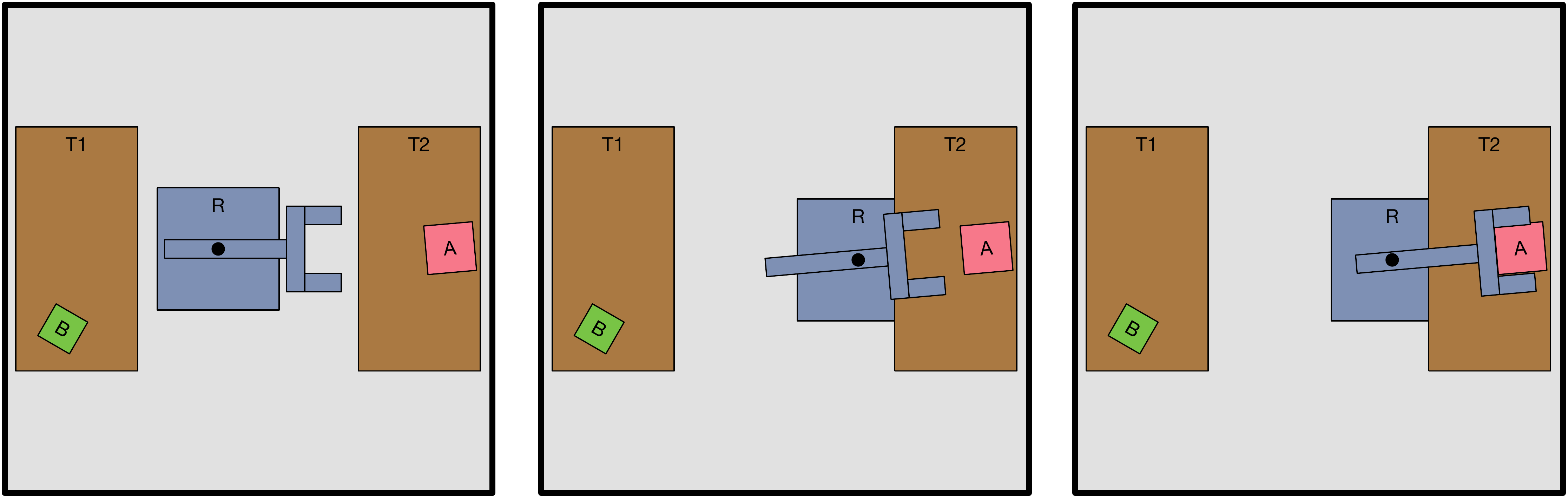}
\caption{Mobile manipulation example application.} \label{fig:rob_initial}
\end{figure*}

We seek to model tabletop manipulation problems involving a manipulator attached to a movable base as a factored transition system.
The previously presented pick-and-place factored transition system encompasses this application, and the specified samplers lead to probabilistically complete algorithms.
However, the previous formulation leads to poor performance in practice for high-dimensional robot configuration spaces as it attempts to construct control trajectories between all pairs of robot configurations.
%as many linear movements are required to cover the space. 
The resulting control-space is similar to a simplified Probabilistic Roadmap (sPRM)~\citep{Kavraki98probabilisticroadmaps}, which is known to be inefficient for high dimensional robot configuration spaces.
Instead, we model tabletop manipulation problems as transition systems in which multi-waypoint robot trajectories $u_m$ are control parameters.
This allows us to design samplers that call efficient motion planners to produce trajectories between pairs of configurations. 
%Additionally, we partition robot controls $\bar{u} = (u_t, u_m)$ into base trajectories $u_t$ and manipulator trajectories $u_m$. 
%\note{I should just think of $q$ as a base configuration}

\begin{figure}[h]
\centering
\includegraphics[width=0.33\textwidth]{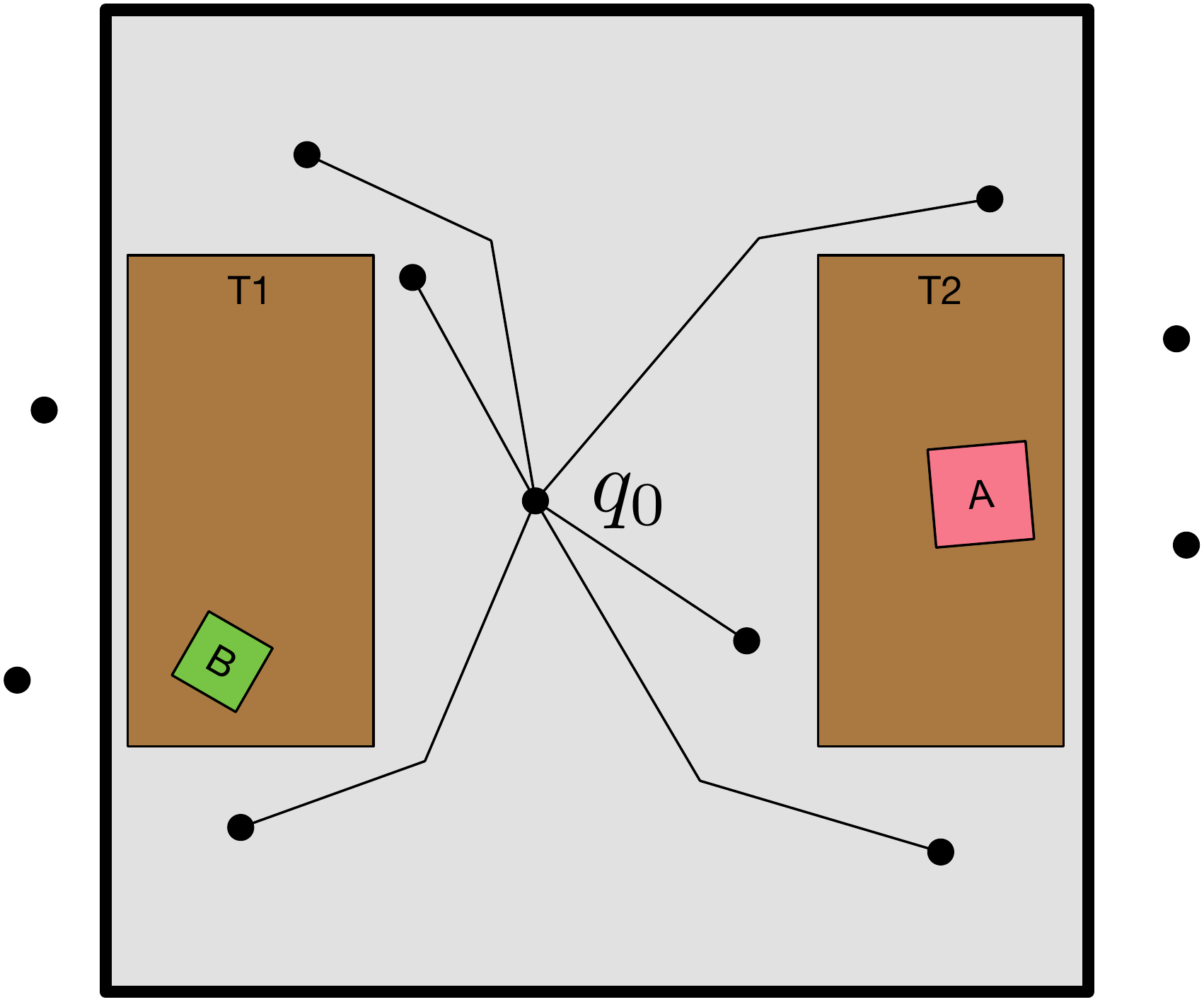}
\caption{Star roadmap comprised of trajectories. Base configurations outside the room are identified as unreachable.} \label{fig:star_roadmap}
\end{figure}

\note{I adopted a different convention for $\id{Manip}(x_o, x_o', x_q, u_m)$ than kin}
% Problems are easy, most constrained near goal 
% Star roadmap in manipulator space
% Carrying configuration?
% Combining limits the horizon
% Constraint space of actions for search
%Manipulator trajectories are responsible for the prehensile manipulation of moveable objects.
Rather than specify ${\cal C}_{\id{Pick}}^o$ and ${\cal C}_\id{Place}^o$ transitions as instantaneous contacts with each object,
we represent robot trajectories moving to, manipulating, and returning from an object as a single transition ${\cal C}_{\id{MPick}}^o$ or ${\cal C}_{\id{MPlace}}^o$.
%For tabletop manipulation, robot trajectories away from tables are safely executable for almost all possible poses of movable objects.
\begin{align*}
%{\cal C}_{\id{MPick}}^o =& \{\id{Stable}(x_o), \id{Grasp}(x_o'), \id{Manip}(x_o, x_o', x_q, u_m), \\
%& x_h = \kw{None}, x_h' = o,  \}\; \cup \\
%& \{x_{o'} = x'_{o'}, \id{CFree}_{o'}(u_m, x_{o'}) \mid o' \in {\cal O}, o \neq o'\}
{\cal C}_{\id{MPick}}^o =& \{\id{Stable}_o, \id{Grasp}_o', \id{Manip}, \\
& x_h = \kw{None}, x_h' = o,  \}\; \cup \\
& \{x_{o'} = x'_{o'}, \id{CFree}_{o'}) \mid o' \in {\cal O}, o \neq o'\}
\end{align*}
\begin{align*}
{\cal C}_\id{MPlace}^o =& \{\id{Grasp}_o, \id{Stable}_o', \id{Manip}', \\
& x_h = o, x_h' = \kw{None}\}\; \cup \\
& \{x_{o'} = x'_{o'}, \id{CFree}_{o'} \mid o' \in {\cal O}, o \neq o'\}
\end{align*}
% Could instead have a reachable condition on these and then plan a direct motion plan that avoid constraints
This behavior is enforced be a manipulation constraint $\id{Manip}$ on parameters $x_o, x_o', u_m$ representing poses $x_o, x_o'$ for object $o$ and trajectory $u_m$.
%ensures that the manipulation trajectories $u_m$ satisfy a motion that attaches the robot to object $o$ at pose $x_o$ with grasp $x_o'$.
Let $q_0$ be the initial robot configuration and $q_{\id{Kin}}$ be a kinematic solution for end-effector transform $x_o'x_o^{-1}$ grasping object $o$ with grasp $x_o^{-1}$ at placement $x_o'$.
The trajectory $u_m$ is the concatenation of a motion plan from $q_0 \to q_{\id{Kin}}$, a grasp plan, and a motion plan $q_{\id{Kin}} \to q_0$. 
Both motion plans are computed to be free of collisions with fixed obstacles. 
Additionally, one motion plan avoids collisions with $o$ placed at $x_o'$, and the other avoids collisions between $o$ held at grasp $x_o$ and fixed obstacles.
Because the robot always returns to $q_0$, the robot configuration need not be included as a state variable.

We structure the transition system this way based on the insight that the bulk of the robot's configuration space is unaffected by the placements of movable obstacles.
The moveable obstacles mostly only prevent the safe execution of manipulator trajectories above tabletops. 
Rather than directly plan paths between pairs of configurations manipulating objects, we instead plan paths to a home configuration chosen arbitrarily as the initial configuration $q_0$.
This approach guarantees the feasibility of the resulting plan while not inducing significant overhead.
Shorter, direct trajectories between pairs of base configurations can later be produced when post-processing a solution.
Figure~\ref{fig:star_roadmap} visualizes the set of $u_m$ trajectories as edges in a star roadmap~\citep{garrettIJRR2017} with $q_0$ as the root.

%\begin{align*}
%{\cal C}_\id{Pick}^o =& \{\id{Stable}_o (x_o), \id{Grasp}_o (x_o'), \id{Manip}_o(x_o, x_o', x_q, u_m), \\
%& x_q = x_q', x_h = \kw{None}, x_h' = o,  \}\; \cup \\
%& \{x_{o'} = x'_{o'}, \id{CFree}_{o'}(u_m, x_{o'}) \mid o' \in {\cal O}, o \neq o'\}
%\end{align*}
%\begin{align*}
%{\cal C}_\id{Place}^o =& \{\id{Grasp}_o (x_o), \id{Stable}_o (x_o'), \id{Manip}_o(x_o', x_o, x_q, u_m), \\
%& x_q = x_q', x_h = o, x_h' = \kw{None}\}\; \cup \\
%& \{x_{o'} = x'_{o'}, \id{CFree}_{o'}(u_m, x_{o'}) \mid o' \in {\cal O}, o \neq o'\}
%\end{align*}
%\begin{itemize}
%\item ${\cal C}_{ToHome} = \{{Motion}(q, \tau, q'), h = h'\} \cup \{p_{o'} = p'_{o'} \mid o' \in {\cal O}\}$ 
%\item ${\cal C}_{Pick}^o = \{\langle {Stable}_o (p_o), {Grasp}_o (p'_o), q = q', h = \kw{None}, h' = o, {Manip}_o (p_o', p_o, q, \mu) \} \cup \{p_{o'} = p'_{o'}, {CFree}_{o'}(\mu, p_{o'}) \mid o' \in {\cal O}, o \neq o'\}$
%\end{itemize}

\note{Placement instead of Stable or Pose}
We specify the following samplers to produce values satisfying the constraints in this transition system.
The grasp sampler $\psi_P^o$ and placement $\psi_P^o$ sampler are the same as before.
%with the exception that \proc{sample-pose} prunes poses that are in collision with fixed obstacles.
%\begin{equation*}
%\psi_G^o = \big\langle (), \emptyset, (x_o), \{\id{Grasp}(x_o)\}, \proc{sample-grasp}\big\rangle
%\end{equation*}
%\begin{equation*}
%\psi_P^o = \big\langle (), \emptyset, (x_o), \{\id{Stable}(x_o)\}, \proc{sample-pose}\big\rangle
%\end{equation*}
The manipulation sampler $\psi_M^o$ is similar to $\psi_{IK}^o$:

\begin{align*}
%\psi_M^o = \big\langle (x_o, x_o'), \{\id{Stable}(x_o), \id{Grasp}(x_o')\}, (x_q, u_m),& \\
\psi_M^o = \big\langle (x_o, x_o'), (u_m), \{\id{Manip}_o\}, \proc{sample-manip}\big\rangle&.
\end{align*}
The procedure \proc{sample-manip} samples a nearby base pose via inverse reachability.
From this base pose, it performs manipulator inverse kinematics to identify a grasping configuration to perform the pick or place. 
If \proc{sample-manip} fails to find a kinematic solution, it samples a new base pose.
Otherwise, it calls a sampling-based motion planner twice to find motion plans to this grasping configuration and back.
These motion plans are computed to not be in collision with fixed obstacles or $o$ both when it is on the table and when it is held.
Additionally, each call has a timeout meta parameter to ensure termination. 
The timeout for each sampler instance of $\psi_M^o$ is increased after each call allowing \proc{sample-manip} to have an unbounded amount of time collectively over all of its calls.

%Finally, the reachability sampler $\psi_R$ calls \proc{rrt-connect} in the base configuration space from $q_0 \to x_q$.
%\begin{equation*}
%\psi_R = \big\langle (x_q), \emptyset, (u_t), \{\id{Reachable}(u_t, x_q)\}, \proc{rrt-connect}\big\rangle
%\end{equation*}

%\begin{equation*}
%\psi_T = \big\langle (x_q), \emptyset, (u_t), \{\id{Reachable}(u_t, x_q), \proc{birrt} \} \big\rangle
%\end{equation*}
%\begin{equation*}
%\psi_{\id{grasp}_o} = \big\langle (), \emptyset, (x_o), \{\id{Grasp}(x_o), \proc{sample-grasp}\}\big\rangle
%\end{equation*}
%\begin{equation*}
%\psi_{\id{stable}_o} = \big\langle (), \emptyset, (x_o), \{\id{Stable}(x_o), \proc{sample-pose}\}\big\rangle
%\end{equation*}
%\begin{align*}
%\psi_{\id{manip}_o} =& \big\langle (x_o, x_o'), \{\id{Stable}(x_o), \id{Grasp}(x_o')\}, (x_q), \\
%&\{\id{Kin}(x_o, x_o', x_q), \proc{sample-ik}\}\big\rangle
%\end{align*}

\note{Correctness condition of planners using samplers?}

\section{Example Mobile Manipulation Problem}

To illustrate both the \incremental{} and \focused{} algorithms, we work through their steps on an example mobile manipulation problem.
Consider the problem in figure~\ref{fig:rob_initial} with two movable objects $A, B$ and two tables $T1, T2$.
States are $\bar{x} = (x_A, x_B, x_h)$ and controls are $\bar{u} = u_m$.
The initial state is $\bar{x}_0 = (a_0, b_0, \kw{None})$ and the goal constraints are ${\cal C}_* = \{\id{Region}_A\}$ indicating that object $A$ is placed on $T1$.
For simplicity, we assume that each moveable object has a single grasp $a_g$ or $b_g$.
The values $a_0, b_0, a_g, b_g \in \text{SE}(3)$ represent continuous transformations.
Similarly, manipulations $m$ are full body motion plans from $q_0$ to a grasping configuration and back.
%Additionally, new placements can only be generated on table $T2$. 
For the example, we assume that the conditional samplers never fail to produce an appropriate value.

\subsection{Incremental Algorithm}

Table~\ref{fig:inc_table} traces the sampler instances $S_i$ for which \proc{sample} is called paired with the resulting element for each iteration $i$ of the \incremental{} algorithm.
The set of \elements{} available on each iteration is the union of the previously sampled elements $S_j, j < i$. 
The \incremental{} algorithm fails to find a plan for 2 iterations and finally finds the following plan $\pi_3$ on the 3rd iteration.
\begin{align*}
\vec{a}_3 &= [{\cal C}_\id{MPick}^A, {\cal C}_\id{MPlace}^A] \\
\vec{x}_3 &= [(a_0, b_0, \kw{None}),  (a_g, b_0, A), (a_1, b_0, \kw{None})] \\
\vec{u}_3 &= [m_1, m_3]
\end{align*}

Notice that the number of sampler instances sampled per iteration grows quickly.
Additionally, samples are generated for both objects $A$ and $B$ despite the task only requiring manipulating $A$.

\note{Write the \id{Var} or tests?}

% Poses are even grasps are odd
% Starting the manipulations from 0
\renewcommand{\arraystretch}{1.25}
\begin{table*}
    \begin{footnotesize}
    \begin{tabular}{| l | l | l | l |}
    \hline
    %It. & \elements{} \\ \hline
   \rowcolor[gray]{.9}  $S_0$ & $\proc{initial-elements}({\cal P}): \{\id{Stable}(a_0), \id{Stable}(b_0))\}$ \\ \hline\hline

    $\vec{a}_1$ & \kw{None} \\ \hline
    \rowcolor[gray]{.9} $S_1$ & $\psi_G^a(): \id{Grasp}(a_g), \psi_G^b(): \id{Grasp}(b_g), \psi_P^A(): \id{Stable}(a_1), \psi_P^B(): \id{Stable}(b_1), \id{Region}(a_1)$  \\ \hline\hline
    
    $\vec{a}_2$ & \kw{None} \\ \hline
    \rowcolor[gray]{.9} $S_2$ & $\psi_M^A(a_0, a_g):  \id{Manip}(a_0, a_g, m_1), \psi_M^B(b_0, b_g): \id{Manip}(b_0, b_g, m_2), \psi_M^A(a_1, a_g): \id{Manip}(a_1, a_g, m_3),$ \\ 
    \rowcolor[gray]{.9} & $\psi_{M}^B(b_1, b_g): \id{Manip}(b_1, b_g, m_4), \psi_P^A(): \id{Stable}(a_2), \psi_P^B(): \id{Stable}(b_2), \id{Region}(a_2)$  \\
    \rowcolor[gray]{.9} & $\id{CFree}(m_1, b_0), \id{CFree}(m_1, b_1), \id{CFree}(m_1, b_2), \id{CFree}(m_2, a_0), \id{CFree}(m_2, a_1), \id{CFree}(m_2, a_2)$ \\ \hline\hline
    
    $\vec{a}_3$ & $\vec{a}_3 = [{\cal C}_\id{MPick}^A, {\cal C}_\id{MPlace}^A], \vec{x}_3 = [(a_0, b_0, \kw{None}), (a_g, b_0, A), (a_1, b_0, \kw{None})], \vec{u}_3=[m_1, m_2]$ \\ \hline
    \end{tabular}
    \end{footnotesize}
    \caption{Example walkthrough of the \incremental{} algorithm. Each $S_i$ displays the set of sampler instances for which \proc{sample} is called along with the new elements produced.} \label{fig:inc_table}
\end{table*}

\subsection{Focused Example}

Table~\ref{fig:foc_table} traces each iteration $i$ of the \incremental{} algorithm.
We will assume that \id{new\_elements} are directly added to \id{elements}.
% Can also fail after each iteration
As before, $S_i$ contains the sampler instances \proc{sample} paired with the resulting elements.
Elements certified by \id{CFree} and \id{Region} are individually added to $S_i$.
The set of \id{sampled} sampler instances are contained in $S_j, i < j$.
Let ${\cal E}_i$ be the set of elements using lazy samples generated by \proc{process-samplers} on each iteration.
The union of ${\cal E}_i$ and \elements{} is \mixed{}.
The plan returned on each iteration is denoted by $\pi_i = \langle \vec{a}_i, \vec{x}_i, \vec{u}_i \rangle$.
We denote the lazy sampler  for each sampler as follows:
\begin{equation*}
\proc{lazy-sample}(\psi_G^o()) = \{\id{Grasp}(\lzg{o})\}
\end{equation*}
\begin{equation*}
%\proc{lazy-sample}(\psi_P^o()) = \{\id{Pose}(\lzp{o}), \id{Stable}(\lzp{o})\}
\proc{lazy-sample}(\psi_P^o()) = \{\id{Stable}(\lzp{o})\}
\end{equation*}
\begin{equation*}
%\proc{lazy-sample}(\psi_M^o(x_o, x_o')) = \{\id{Manip}(x_o, x_o', \lzq{o}, \lzm{o})\}
\proc{lazy-sample}(\psi_M^o(x_o, x_o')) = \{\id{Manip}(x_o, x_o', \lzm{o})\}
\end{equation*}
%\begin{equation*}
%\proc{lazy-sample}(\psi_T(x_q)) = \{\id{Reachable}(\lzt, x_q)\}.
%\end{equation*}

\note{If the only structure is a DAG, then you can avoid worrying about shared lazy or even having to move them to a lazy pool}
\note{I could go back to using full samples}
\begin{table*}
    \begin{footnotesize}
    \begin{tabular}{| l | l | l | l |}
    \hline
   \rowcolor[gray]{.9} $S_0$ & $\proc{initial-elements}({\cal P}): \{\id{Stable}(a_0), \id{Stable}(b_0)\}$ \\ \hline\hline
    
    ${\cal E}_1$ & $\id{Grasp}(\lzg{A}), \id{Grasp}(\lzg{B}), \id{Stable}(\lzp{A}), \id{Region}(\lzp{A}), \id{Stable}(\lzp{B}), \id{Manip}(a_0, \lzg{A}, \lzm{A}), \id{Manip}(b_0, \lzg{B}, \lzm{B}),$   \\
    & $\id{Manip}(\lzp{A}, \lzg{A}, \lzm{A}), \id{Manip}(\lzp{B}, \lzg{B}, \lzm{B}), \id{CFree}(\lzm{A}, b_0), \id{CFree}(\lzm{A}, \lzp{B}), \id{CFree}(\lzm{B}, a_0), \id{CFree}(\lzm{B}, \lzp{A})$ \\ \hline
    
    %$\vec{a}_1$ & ${\cal C}_\id{Pick}^A, {\cal C}_\id{MPlace}^A$ \\ \hline
    %$\vec{x}_1$ & $(a_0, b_0, \kw{None}), (\lzg{A}, b_0, A), (\lzp{A}, b_0, \kw{None})$ \\ \hline
    %$\vec{u}_1$ & $\lzm{A}, \lzm{A}$ \\ \hline
    $\vec{\pi}_1$ & $\vec{a}_1 = [{\cal C}_\id{MPick}^A, {\cal C}_\id{MPlace}^A], \vec{x}_1=[(a_0, b_0, \kw{None}), (\lzg{A}, b_0, A), (\lzp{A}, b_0, \kw{None})], \vec{u}_1=[\lzm{A}, \lzm{A}]$ \\ \hline
   \rowcolor[gray]{.9}  $S_1$ & $\psi_G^A(): \id{Grasp}(a_g), \psi_P^A(): \id{Stable}(a_1), \id{Region}(a_1)$ \\ \hline\hline
    
    ${\cal E}_2$ & $\id{Grasp}(\lzg{B}), \id{Stable}(\lzp{B}), \id{Manip}(a_0, a_g, \lzm{A}), \id{Manip}(a_1, a_g, \lzm{A}), \id{CFree}(\lzm{A}, b_0), \id{CFree}(\lzm{A}, \lzp{B}), \id{CFree}(\lzm{B}, a_0),$  \\
    & $\id{CFree}(\lzm{B}, a_1), \id{Manip}(b_0, \lzg{B}, \lzm{B}), \id{Manip}(\lzp{B}, \lzg{B}, \lzm{B})$  \\ \hline
    $\vec{\pi}_2$ & $\vec{a}_2 = [{\cal C}_\id{MPick}^A, {\cal C}_\id{MPlace}^A], \vec{x}_2=[(a_0, b_0, \kw{None}), (a_g, b_0, A), (a_1, b_0, \kw{None})], \vec{u}_2=[\lzm{A}, \lzm{A}]$ \\ \hline
    %$\vec{a}_1$ & ${\cal C}_\id{MPick}^A, {\cal C}_\id{MPlace}^A$ \\ \hline
    %$\vec{x}_2$ & $(a_0, b_0, \kw{None}), (a_g, b_0, A), (a_1, b_0, \kw{None})$ \\ \hline
    %$\vec{u}_2$ & $\lzm{A}, \lzm{A}$ \\ \hline
    \rowcolor[gray]{.9} $S_2$ & $\psi_M^A(a_0, a_g):  \id{Manip}(a_0, a_g, m_1), \psi_M^A(a_1, a_g): \id{Manip}(a_1, a_g, m_2), \id{CFree}(m_1, p_0), \id{CFree}(m_2, p_0)$ \\ \hline\hline
   
    ${\cal E}_3$ & $\id{Grasp}(\lzg{B}), \id{Stable}(\lzp{B}), \id{CFree}(m_1, \lzp{B}), \id{CFree}(m_2, \lzp{B}), \id{Manip}(b_0, \lzg{B}, \lzm{B})$  \\
    & $\id{Manip}(\lzp{B}, \lzg{B}, \lzm{B}), \id{CFree}(\lzm{B}, a_0), \id{CFree}(\lzm{B}, a_1)$ \\ \hline
    $\vec{a}_3$ & $\vec{a}_3 = [{\cal C}_\id{MPick}^A, {\cal C}_\id{MPlace}^A], \vec{x}_3 = [(a_0, b_0, \kw{None}), (a_g, b_0, A), (a_1, b_0, \kw{None})], \vec{u}_3=[m_1, m_2]$ \\ \hline
    \end{tabular}
    \end{footnotesize}
    \caption{Example walkthrough of the \focused{} algorithm. Each ${\cal E}_i$ displays the set of lazy elements at the start of the iteration. Each $S_i$ displays the set of sampler instances for which \proc{sample} is called along with the new elements produced.} \label{fig:foc_table}
\end{table*}
On the first iteration, the sampler instances $\psi_G^a(), \psi_P^a()$ are sampled to produce values for $\lzg{A}, \lzp{A}$ respectively.
On the second iteration, $\psi_M^a(a_0, a_g)$ and $\psi_M^a(a_1, a_g)$ generate the manipulations required for the ${\cal C}_\id{MPick}^A$ and ${\cal C}_\id{MPlace}^A$ transitions given the new grasp $a_g$ and placement $a_1$. 
%On the third iteration, $\psi_T(q_1), \psi_T(q_2)$ sample motion plans to base configurations required for the manipulations. 
On the final iteration, a plan $\pi_3$ not requiring any lazy samples is generated, resulting in a solution.

The \focused{} algorithm samples fewer values than incremental by only sampling values determined to be useful for completing lazy samples and satisfying constraints along a plan.
More specifically, it avoids sampling values for object $B$ altogether, saving time by not computing expensive motion plans.
This behavior becomes even more prevalent in problems with many moveable objects, such as ones arising from human environments.

\note{Do I want to assume collision stuff is automatically produced?}
\note{Add constraints to the sequence?}
\note{Pose precondition on collision sampler}
\note{Make the collision stuff right?} 
\note{Ensure the surface stuff is correct}
\note{What if I just used computer output for these?}
\note{What about the initial values here? How do I tell that they are poses?}

\subsection{Additional Example Scenarios}

We sketch out several additional problems and outline how, in particular, the \focused{} algorithm will proceed in each of these scenarios. 

\subsubsection{Sampling Failure:}

We previously assumed that each sampler successfully generated output values satisfying its constraints.
In general, samplers may fail to do so because of timeouts or even because no sample exists.
For example, suppose $\psi_M^A(a_0, a_g)$ fails to produce a collision-free inverse kinematic solution, resulting in a failure.
After the failure, $\psi_M^A(a_0, a_g)$ will be added to \id{sampled}, preventing it from being sampled on the next iteration.
Without $\id{Manip}(a_0, a_g, m_1)$ or $\id{Manip}(a_0, a_g, \lzm{A})$, the \focused{} algorithm will fail to find a plan.
In that case, $\id{sampled}$ is reset allowing $\psi_M^A(a_0, a_g)$ to be sampled again on the next episode.
This cycle will automatically repeat with increased timeouts for as long as $\psi_M^A(a_0, a_g)$ fails to produce a manipulation, as picking object $A$ is 
required for any solution to this problem.

\note{Is the there exists a set of things with a set of successors that will admit a solution thing actually general?}

\subsubsection{Obstructions:}

\begin{figure}[h]
\centering
\includegraphics[width=0.3\textwidth]{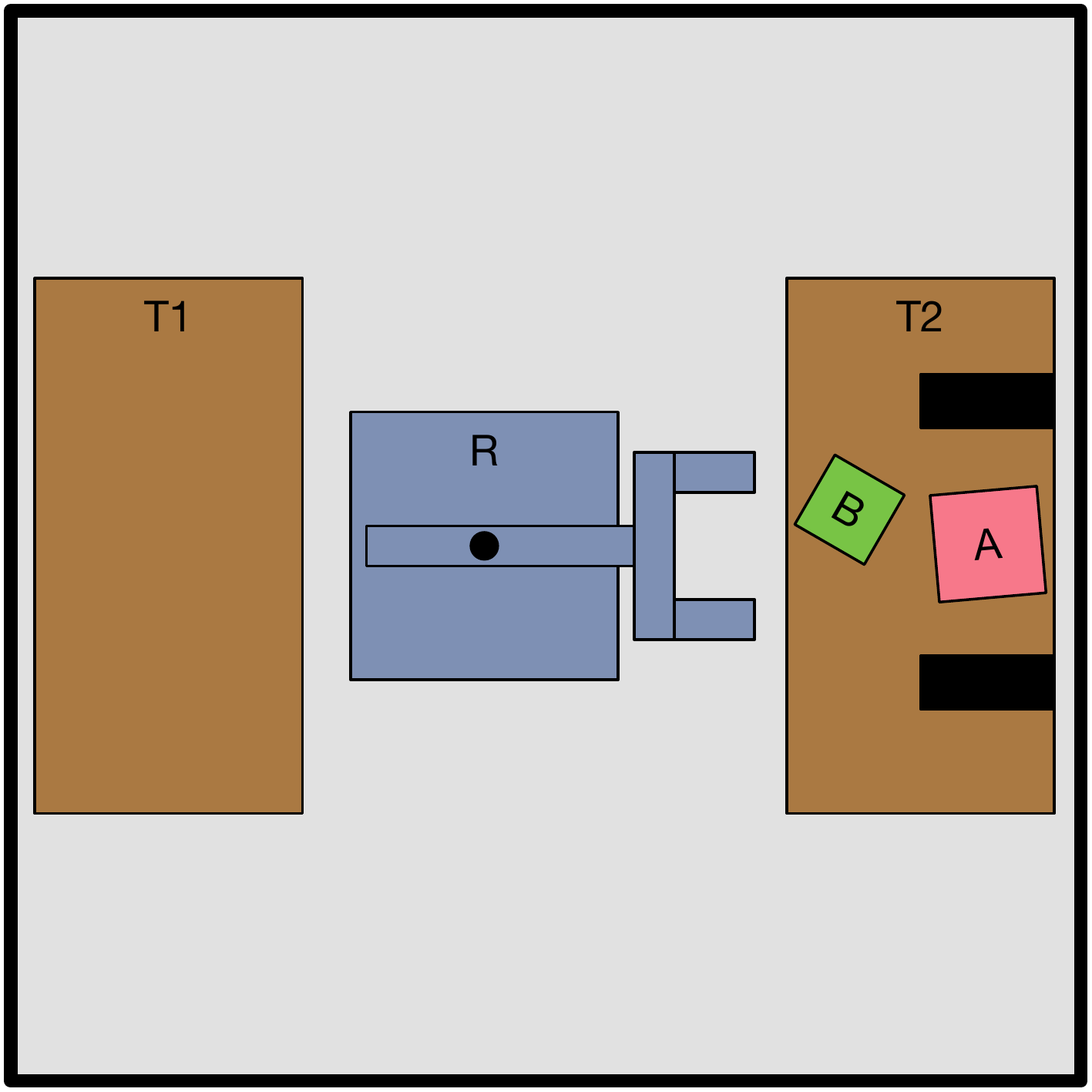}
\caption{Mobile manipulation problem where object $B$ is obstructing manipulations that pick object $A$.} \label{fig:rob_obstruction}
\end{figure}

Suppose that object $A$ is initially obstructed by object $B$ as in figure~\ref{fig:rob_obstruction}.
While $\psi_M^A(a_0, a_g)$ can produce a manipulation $m_1$, it cannot be performed because 
it violates the collision constraint $\id{CFree}(m_1, b_0)$. 
However, the lazy pose $\lzp{B}$ is optimistically assumed to not be in collision with $m_1$.
Thus, a valid plan involves first moving $B$ to $\lzp{B}$ before picking $A$. 
In the event where the sampled value for $\lzp{B}$ is still in collision with $m_1$,
an additional value can be generated on the next episode.
Lazy samples allow the \focused{} algorithm to reason about trajectories $u_m$ that do not yet correspond to a feasible transition because they violate one or more collision constraints.
By sampling concrete values for the lazy samples corresponding to these violated constraints, it can attempt to find transitions for which the control is feasible.

\subsubsection{Regrasp:}

Consider the regrasp experiment in figure~\ref{problem:regrasp} where the robot is unable to pick and place the goal object using the same grasp.
Because of this, it is forced to place the goal object at an intermediate location to change grasps.
The \focused{} algorithm will only create one lazy grasp sample $\lzg{A}$ for object $A$.
On the subsequent iteration, at least one of $\psi_M^A(a_0, a_g)$ and $\psi_M^A(a_*, a_g)$ will fail to sample a manipulation.
Both will be added to $\id{sampled}$ causing the \proc{discrete-search} to fail to find a plan on the next iteration.
After $\id{sampled}$ is reset, the \focused{} algorithm is able to use $\lzg{A}$ to produce the second grasp and arrive at a solution to the problem.

\section{Experiments}

We implemented both algorithms and tested them on a suite of tabletop manipulation problems.
All experiments used the same core factored transition system and same set of conditional samplers as those described in section~\ref{sec:manip}.
%The {\it Push}, {\it Wall}, and {\it Stacking} problems in section~\ref{sec:diverse} contains additional clauses and conditional samplers for pushing and stacking.
%The {\it Dinner} problem in section~\ref{sec:diverse} contains additional clauses for cleaning and cooking.
% No shared poses or grasps
%We give Python code for our experiment transition system in the extended version of this paper~\citep{TODO}.
%The conditional samplers for placement, grasps, inverse kinematics, and motion plans were implemented using OpenRAVE~\citep{openrave}. % with the rest of the implementation in Python.
We wrote our conditional samplers in Python, building on top of the OpenRAVE robotics development environment~\citep{openrave}.
We used the Open Dynamics Engine~\citep{smith2005open} for collision checking.

Each movable object was limited to four side-grasps except for in {\it Experiment 1} where each object has a single-top grasp.
Thus, the grasp conditional sampler $\psi_G$ simply enumerates this finite set.
The side-grasp restriction increases the difficulty of our benchmarks as it creates more opportunities for objects to obstruct each other.

Our placement conditional sampler $\psi_P$ randomly samples poses from a mixture distribution composed of a uniform distribution over stable placements and a uniform distribution over stable placements not in collision given the initial state. 
This strong bias towards initially collision-free placements accelerates the generation of unobstructed placements, particularly in problems where there are many movable objects.

Our manipulation conditional sampler $\psi_M$ samples base poses from a precomputed distribution of 2D base poses, each relative to a 2D object pose, that for some previous query admitted a kinematic solution.
This ``learned" base pose sampler has a greater likelihood of producing base poses that admit kinematic solutions than a sampler that generates base poses uniformly at random in a region near the desired end-effector pose.
% base poses relative to the object pose
% Could also uniformly sample
We use IKFast~\citep{diankov2010automated} for inverse kinematics.
Finally, we implemented $\psi_M$'s sampling-based motion planner using RRT-Connect (Bidirectional Rapidly-exploring Randomized Trees)~\citep{KuffnerLaValle}.
%We sampled placements uniformly-at-random 
%Our placement conditional sampler biases its sampling distribution to produce pose not in collision with the initial object poses.
%UberPool - driving to pick up multiple passengers with fuel and things 

We considered two FastDownward~\citep{helmert2006fast} configurations for the
\proc{\incremental{}} and \proc{\focused{}} algorithms:  {\em H} uses the
FastForward heuristic~\citep{HoffmannN01} in a lazy greedy search and {\em No-H} is a breadth-first search.
%Both configurations benefit from a compilation process that can quickly detect some infeasible problems using admissible heuristics.
%{\em Incremental-H} can be seen as comparable to FFRob by~\cite{garrettIJRR2017}.
%FastDownward automatically performs unreachability detection to prune some parts of the state-space. 
% FastDownward automatically does some pruning and infeasibility detection
%: {\em Incremental, No H}, {\em Incremental, H}, {\em Focused, No H}, {\em Focused, H}.

All trials were run on 2.8 GHz Intel Core i7 processor with a 120 second time limit.
Our Python implementation of the \proc{\incremental{}} and \proc{\focused{}} algorithms can be found here:
\url{https://github.com/caelan/factored-transition-systems}.
We also have developed a similar suite of algorithms for an extension of the PDDL~\citep{mcdermott1998pddl} called STRIPStream~\citep{garrett2017strips} available at
\url{https://github.com/caelan/stripstream}.
%\url{https://github.com/caelan/ss}.
%\url{https://github.com/caelan/pddlstream}.
%\proc{STRIPS}tream, like STRIPS~\citep{Fikes71}, operates on a propositional representation of the world and is more expressive than factored transition systems.
Videos of the experiments are available at \url{https://youtu.be/xJ3OeMAYmgc}.
%\url{https://www.youtube.com/playlist?list=PLNpZKR7uv5ARyjs2IVg9NTa5zcyjPyR-2}.
Base trajectories are post-processed by computing a direct trajectory between pairs of base configurations.

\subsection{Scaling Experiments}

\begin{figure}[h]
\centering
\includegraphics[width=0.45\textwidth]{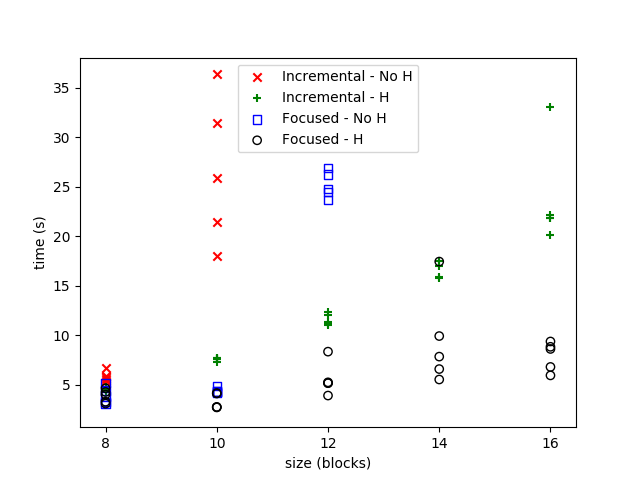}
\caption{{\it Experiment 1}: total runtime of the algorithms over 5 trials per problem size.} \label{fig:scatter1}
\end{figure}

\begin{figure}[h]
\centering
\includegraphics[width=0.45\textwidth]{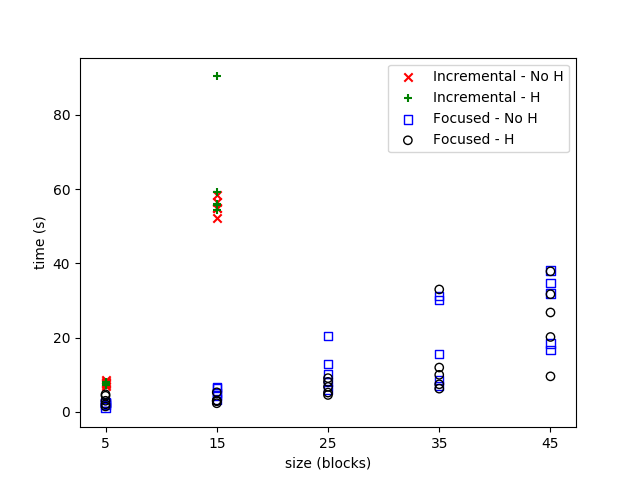}
\caption{{\it Experiment 2}: total runtime of the algorithms over 5 trials per problem size.} \label{fig:scatter2}
\end{figure}

\begin{figure}[h]
\centering
\includegraphics[width=0.45\textwidth]{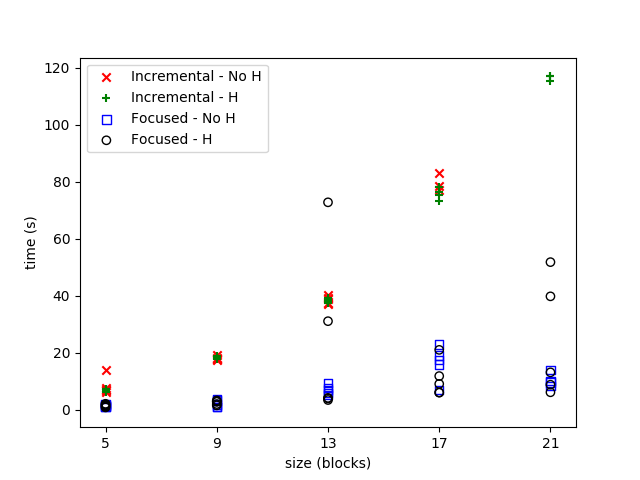}
\caption{{\it Experiment 3}: total runtime of the algorithms over 5 trials per problem size.} \label{fig:scatter3}
\end{figure}

We performed three scaling experiments on pick-and-place problems.
All experiments considered five problem sizes, varying the number of objects.
We performed five trials using randomly (with the exception of {\it Experiment 2}) generated problem instances for each problem size. % and algorithm configuration.
Each scatter plot in figures~\ref{fig:scatter1},~\ref{fig:scatter2}, and~\ref{fig:scatter3} display the total runtime of each configuration per trial. 
Timeouts are indicated by the omission of a trial.

{\it Experiment 1} in figure~\ref{fig:exp3} is the ``grid@tabletop'' benchmark~\citep{krontirisRSS2015} where each object has a specified goal pose.
%{\it Experiment 1} in figure~\ref{fig:exp1} is the ``grid@tabletop'' benchmark~\citep{krontirisRSS2015,krontiris2016icra,garrettIJRR2017} where each object has a specified goal pose.
The initial placements are randomly generated.
The table size scales with the number of objects.
%Each object has a single top-grasp.
As shown in figure~\ref{fig:scatter1}, {\em Focused-H} solved all problem instances and {\em Incremental-H} solved all but one (size=14) indicating that use of a heuristic is necessary for problems with long-horizons.

\begin{figure}[h]
\centering
\includegraphics[width=0.49\textwidth]{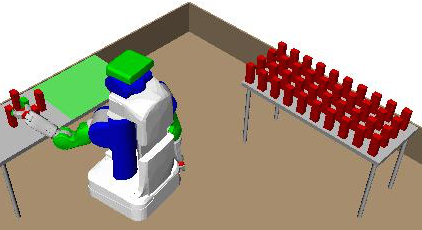}
\caption{{\it Experiment 2}: the robot must place the green object in the green region.} \label{fig:exp2}
\end{figure}

{\it Experiment 2} in figure~\ref{fig:exp2} has the goal that a single green object be placed in the green region.
The green object is obstructed by four red objects. 
The number of distracting red objects on the right table is varied between 0 and 40.
%Each object has four side-grasps. 
This experiment reflects many real-world environments where the state-space is enormous but many objects do not substantially affect a task.
As can be seen in figure~\ref{fig:scatter2}, both {\em Focused-No-H} and {\em Focused-H} solved all problem instances showing that the \proc{\focused{}} algorithm is able to avoid producing samples for objects until they are relevant to the task. 
% Dantam

%\begin{figure}[ht]
%\centering
%\raisebox{.2\height}{\includegraphics[width=0.24\textwidth]{figures/distract_small}}
%\includegraphics[width=0.24\textwidth]{figures/distract_plot.png}
%\caption{Experiment 2 - large statef-space.} \label{fig:exp2}
%\end{figure}

% I seem to be using not the vector config but the approach config here
{\it Experiment 3} in figure~\ref{fig:exp3} has the goal that a single blue object be moved to a different table.
The blue object starts at the center of the visible table, and the red objects are randomly placed on the table.
The table size scales with the number of objects.
%Each object has four side-grasps. 
As shown in figure~\ref{fig:scatter3}, {\em Focused-H} solved all instances and {\em Focused-No-H} solved all but one (size=21).
% Srivastava, Dantam

%\begin{figure}[ht]
%\centering
%\raisebox{.1\height}{\includegraphics[width=0.24\textwidth]{figures/several_small}}
%\includegraphics[width=0.24\textwidth]{figures/several_plot.png}
%\caption{Experiment 3 - several obstructing objects.} \label{fig:exp3}
%\end{figure}

\subsection{Diverse Experiments} \label{sec:diverse}

We experimented on several additional problems to show that the factored transition system framework and algorithms can be successfully applied to problems involving pushing, stacking, and discrete state variables.
We also experimented on two tricky pick-and-place problems that require regrasping and require violating several goal constraints along a plan to achieve the goal.
%We used the same factored transition system and conditional samplers for all problems.
%In each problem, each object has four side-grasps.
We conducted 40 trials per problem and algorithm, and each trial once again had a 120 second time limit.
The success percentage of each algorithm (\%) and mean runtime in seconds for successful trials are displayed in table~\ref{table:results}.
We also show the reported statistics for the best configuration of the HBF~\cite{GarrettIROS15} and FFRob~\cite{garrettIJRR2017} algorithms when applicable.

\subsubsection{Descriptions}
\begin{figure}[h]
\centering
     \includegraphics[width=0.26\textwidth]{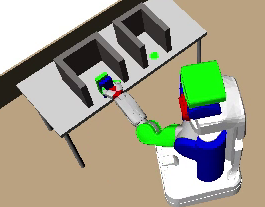}
     \includegraphics[width=0.22\textwidth]{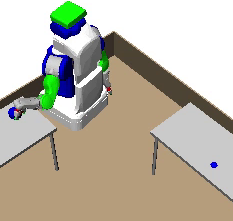}
\caption{{\it Regrasp} (left): a forced regrasp problem. {\it Push} (right): a problem requiring pushing, picking, and placing.} \label{problem:regrasp}
\end{figure}

The regrasp problem ({\it Regrasp}) in figure~\ref{problem:regrasp} is Problem 3 of~\cite{GarrettIROS15}.
The goal constraints are that the green object be at the green pose and the blue object remain at its current pose.
%We also performed five trials on the non-monotonic, regrasp problem in figure~\ref{fig:exp1}.
%Because the blue object must be moved out of the way, this problem is nonmonotonic.
The robot must place the green object at an intermediate pose to obtain a new grasp in order to insert it in the thin, right cupboard.
%All configurations solved all trials in less than 5 seconds. 
This indicates that the algorithms can solve pick-and-place problems where even non-collision constraints affect the plan skeleton of solutions. 

The first pushing problem ({\it Push}) in figure~\ref{problem:regrasp} has the goal constraint that the short blue cylinder on the left table be placed at the blue point on the right table. 
Because the blue cylinder is short and wide, the robot is unable to grasp it except by side grasps at the edges of each table.
Thus, the robot must first push the cylinder to the edge of the left table, pick the cylinder, place the cylinder on the edge of the right table, and push the cylinder to the goal point.
This problem introduces an additional transition relating to trajectories corresponding movements between two poses.
Additionally, it requires a conditional sampler to generate push Cartesian trajectories and motion plans per pairs of poses on the same table.

%\begin{figure}[h]
%     \centering
%     \includegraphics[width=0.44\textwidth]{new_figures/push_small}
%     \caption{The second state and last state on a plan for problem 1-2.}
%     \label{problem:push}
%\end{figure}

\begin{figure}[h]
     \centering
     \includegraphics[width=0.44\textwidth]{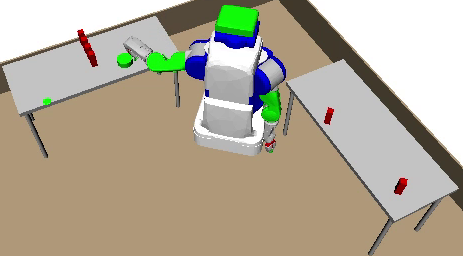}
     \caption{{\it Wall}: a pushing problem involving a wall of blocks.} \label{problem:push_wall}
\end{figure}

The second pushing problem ({\it Wall}) in figure~\ref{problem:push_wall} is Problem 2 of~\cite{GarrettIROS15} where the goal constraint is that the short green cylinder be placed at the green point.
A wall of moveable objects initially blocks the robot from pushing the the green cylinder to the goal point.
However, if several of these blocks are moved, the robot can execute a sequence of pushes to push the green cylinder directly to its goal.

\begin{figure}[h]
	\centering
	\includegraphics[width=0.44\textwidth]{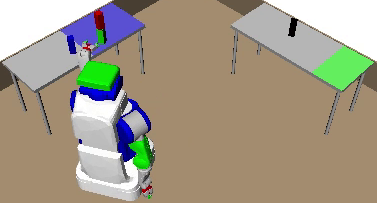}
	\caption{{\it Stacking}: a problem requiring unstacking and stacking.} \label{problem:stacking}
\end{figure}

The stacking problem ({\it Stacking}) in figure~\ref{problem:stacking} is Problem 4 of~\cite{GarrettIROS15}. 
The goal constraints are that the blue block be contained within the blue region, the green block be contained within the green region, and the black block be on top of the blue block.
The robot must unstack the red block to safely move the green block.
This problem requires a modification of the transition system to account for stability constraints.
Additionally, it requires a conditional sampler that produces poses of the black block on the blue block given poses of the blue block. 

\begin{figure}
     \centering
     \includegraphics[width=0.44\textwidth]{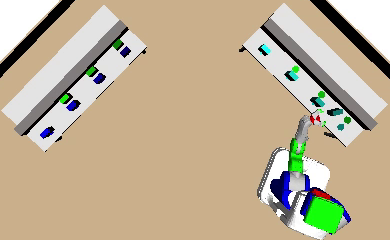}
     \caption{{\it Nonmon.}: a nonmonotonic pick-and-place problem.} \label{problem:ffrob}
\end{figure}

The pick-and-place problem ({\em Nonmon.}) in figure~\ref{problem:ffrob} is Problem 3-2 of~\cite{garrettIJRR2017}.
The goal constraints are that the green blocks be moved from their initial pose on the left table to their corresponding pose on the right table.
Additionally, there are goal constraints that each blue and cyan block end at its initial pose.
This is a highly nonmonotonic problem as solutions require violating several goal constraints satisfied by the initial state.

\begin{figure}
     \centering
     \includegraphics[width=0.44\textwidth]{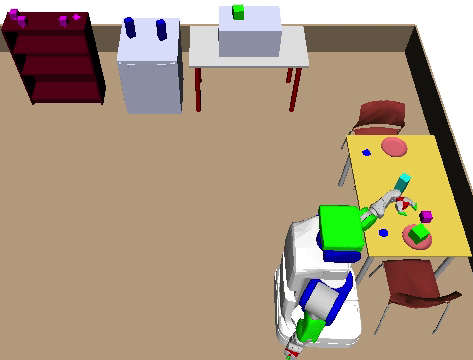}
     \caption{{\em Dinner}: a task and motion planning problem. } \label{problem:dinner}
\end{figure}

The task and motion planning problem ({\it Dinner}) in figure~\ref{problem:dinner} is Problem 5 of~\cite{garrettIJRR2017}.
The state contains an additional discrete state variable for each block indicating whether it is {\em dirty}, {\em clean}, or {\em cooked}.
The transition relation contains additional clauses to clean a dirty block when it is placed on the dishwasher and to cook a clean block when it is placed on the microwave.
The goal constraints are that the green blocks (``cabbage'') be cooked and placed on the plates, the blue blocks (``cups'') be cleaned and placed at the blue points, the cyan block (an unnecessary ``cup'') be cleaned, and the pink blocks (``turnips'') remain placed on the shelf.
To reach the cabbage, the robot must first move several turnips and the later replace them to keep the kitchen tidy.

\subsubsection{Results}
Each algorithm performed comparably on the first four problems ({\it Regrasp}, {\it Push}, {\it Wall}, {\it Stacking}).
When compared to HBF~\citep{GarrettIROS15}, {\em Focused-H} has a slightly lower average runtime for {\it Regrasp} and {\it Stacking} and about the same average runtime for  {\it Wall}. 
However, the average runtime for all algorithms on these problems is less than 15 seconds.
Only the heuristically informed algorithms where able to consistently solve the larger last two problems  ({\it Nonmon.}, {\it Dinner}).
For problem {\it Nonmon.}, {\em Incremental-H} slightly outperformed {\em Focused-H} because this problem requires manipulating each object. 
Thus, {\em Incremental-H} and {\em Focused-H} produce comparable sets of samples, but {\em Incremental-H} has less overhead.
Both algorithms performed significantly better than best algorithm of~\cite{garrettIJRR2017}. % which had a 72\% success rate and took on average 135 seconds.
For problem {\it Dinner}, the heuristic guided {\em Incremental-H} and {\em Focused-H} planners were able to quickly solve the problem reinforcing the point that search guidance is necessary for problems over long horizons.
These algorithms compare favorably to the best algorithm of~\cite{garrettIJRR2017}. % which had a 76\% success rate and took on average 44 seconds.

%\begin{table}
%\centering
%%\includegraphics[width=0.31\textwidth]{images/separate28.png}
%\begin{footnotesize}
%\begin{tabular}{||c||c|g||c|g||c|g||c|g||}
%\hline
%& \multicolumn{2}{c||}{\em Incr.}&\multicolumn{2}{c||}{\em Incr. - H}&\multicolumn{2}{c||}{\em Focus}&\multicolumn{2}{c||}{\em Focus - H}\\
%\hline
%{\bf Problem} & {\bf \%} & {\bf t} & {\bf \%} & {\bf t} & {\bf \%} & {\bf t} & {\bf \%} & {\bf t}
%\\ \hline
%{\it Regrasp} & 98 & 1 & 100 & 2 & 98 & 1 & 95 & 1
%\\ \hline
%{\it Push} & 100 & 11 & 100 & 13 & 100 & 13 & 100 & 9
%\\ \hline
%{\it Wall} & 95 & 10 & 98 & 13 & 100 & 6 & 100 & 8
%\\ \hline
%{\it Stacking} & 100 & 9 & 100 & 9 & 100 & 2 & 100 & 3
%\\ \hline
%{\it Nonmon.} & 25 & 21 & 98 & 15 & 0 & - & 88 & 43
%\\ \hline
%{\it Dinner} & 0 & - & 100 & 27 & 0 & - & 98 & 22
%\\ \hline
%\end{tabular}
%\end{footnotesize}
%\caption{The success percentage (\%) and mean runtime (t) for the diverse experiments over 40 trials.}
%\label{table:results}
%\end{table}

\begin{table*}
\centering
\begin{footnotesize}
\begin{tabular}{||c||c|g||c|g||c|g||c|g||c|g||c|g||}
\hline
&\multicolumn{2}{c||}{HBF (2015)}&\multicolumn{2}{c||}{FFRob (2017)}&\multicolumn{2}{c||}{\em Incr.}&\multicolumn{2}{c||}{\em Incr. - H}&\multicolumn{2}{c||}{\em Focus}&\multicolumn{2}{c||}{\em Focus - H}\\
\hline
{\bf Problem} & {\bf \%} & {\bf t} & {\bf \%} & {\bf t} & {\bf \%} & {\bf t} & {\bf \%} & {\bf t} & {\bf \%} & {\bf t} & {\bf \%} & {\bf t}
\\ \hline
{\it Regrasp} & 100 & 6 & - & - & 98 & 1 & 100 & 2 & 98 & 1 & 95 & 1
\\ \hline 
{\it Push} & - & - & - & - & 100 & 11 & 100 & 13 & 100 & 13 & 100 & 9
\\ \hline
{\it Wall} & 100 & 7 & - & - & 95 & 10 & 98 & 13 & 100 & 6 & 100 & 8
\\ \hline
{\it Stacking} & 97 & 12 & - & - & 100 & 9 & 100 & 9 & 100 & 2 & 100 & 3
\\ \hline
{\it Nonmon.} & - & - & 72 & 135 & 25 & 21 & 98 & 15 & 0 & - & 88 & 43
\\ \hline
{\it Dinner} & - & - & 74 & 44 & 0 & - & 100 & 27 & 0 & - & 98 & 22
\\ \hline
\end{tabular}
\end{footnotesize}
\caption{The success percentage (\%) and mean runtime in seconds (t) for the diverse experiments over 40 trials compared to the reported results for the HBF~\citep{GarrettIROS15} and FFRob~\citep{garrettIJRR2017} algorithms. A dash (-) indicates that no data was available.}
\label{table:results}
\end{table*}

\note{Motion, multi-goal motion, 2D pick-and-place}
\note{Set a random seed per sampler and compare plan to the best possible one}

\section{Conclusion}

%The experiments demonstrate that these domain-independent algorithms scale well on challenging problems.

%We introduced factored transition systems for modeling discrete-time planning problems for hybrid systems.
We introduced factored transition systems for specifying planning problems in discrete-time hybrid systems.
Factored transition systems can model motion planning, pick-and-place planning, and task and motion planning applications.
The transition dynamics for multi-object manipulation are significantly factorable.
Legal transitions can be expressed as the conjunction of constraints each involving only several state or control variables.

Conditional constraint manifolds enabled us to give a general definition of robust feasibility for factored transition systems. 
Under certain conditions, they allow us to describe a submanifold of plan parameter-space resulting from the intersection of dimensionality-reducing constraints.
Thus, robustness properties can be examined relative to this submanifold rather than the for plan parameter-space.
We introduced the idea of conditional samplers: samplers that given input values, produce a sequence of output values satisfying a constraint with the input values.
When appropriate conditional samplers are specified for each conditional constraint manifold, the resulting collection of samplers is sufficient for solving any robustly feasible problem.
Sampling benefits from factoring because a small collection of samples for each variable can correspond to a large number of combined states and transitions.

We gave two general-purpose algorithms that are probabilistically complete given sufficient samplers. 
The \incremental{} algorithm iteratively calls each conditional sampler and tests whether the set of samples is sufficient using a blackbox, discrete search.
The \focused{} algorithm first creates lazy samples representing hypothetical outputs of conditional samplers and then uses a blackbox, discrete search to identify which lazy samples could be useful.
We empirically demonstrated that these algorithms are effective at solving challenging pick-and-place, pushing, and task and motion planning problems.
The \focused{} algorithm is more effective than the \incremental{} algorithm in problems with many movable objects as it can selectively produce samples for only objects affecting the feasibility of a solution.
Additionally, both algorithms were more efficient when using a discrete search subroutine that exploited factoring in the search through domain-independent heuristics.

% Focused, FD solved all instances
% Eager, FD solved all but one rearrangement 14
% Focused, FD solves up to rearrangement 12
% Eager, Dijkstra solves up to rearrangement 10

%Experiments
%\begin{itemize}
%\item Factored representation vs non-factored representation
%\item No heuristic vs heuristic 
%\item Automatic axioms vs none
%\item Comparison to others
%\item Focused vs incremental - lazy planning
%\end{itemize}

\subsection{Future Work}

Future work involves developing additional algorithms for solving factored transition systems.
In particular, both the \incremental{} and \focused{} algorithms treat \proc{discrete-search} as a blackbox.
By directly integrating the search and sampling, an algorithm may be able to more directly target sampling based on the search state and possible transitions.
For example, Backward-Forward Search~\citep{GarrettIROS15} performs its search directly in the hybrid state-space instead of a discretized state-space. 
%By exploring samples useful for a relaxation of the problem, it is able to produce values for successor transitions.

Our formulation gives rise to several new opportunities for learning to improve sampling and search runtimes.
For instance, one could learn a policy to decide how frequently to sample each conditional sampler.
A high performing policy would balance the likelihood of a conditional sampler to produce useful samples, the overhead of computing samples, and the impact additional samples have on subsequent discrete searches.
Similarly, in the \focused{} algorithm, one could learn costs associated with using lazy samples reflective of the expected future planning time resulting from sampling from a particular conditional stream.
These costs could cause \proc{discrete-search} to produce plans that are likely realizable without too much overhead.

Finally, this work can be extended to optimal planning settings where there are nonnegative costs $c(\bar{u})$ on control inputs.
This will require adapting properties such as asymptotic optimality~\citep{karaman2011sampling} to the factored transition system setting and modifying the \incremental{} and \focused{} algorithms to achieve these properties.

\section{Acknowledgements}

We thank Beomjoon Kim, Ferran Alet, and Zi Wang for their feedback on this manuscript.
We gratefully acknowledge support from NSF grants 1420316, 1523767 and 1723381,  from AFOSR FA9550-17-1-0165, from ONR grant N00014-14-1-0486.
Caelan Garrett is supported by an NSF GRFP fellowship with primary award number 1122374.
Any opinions, findings, and conclusions or recommendations expressed in this material are those of the authors and do not necessarily reflect the views of our sponsors.


\begin{thebibliography}{69}
\providecommand{\natexlab}[1]{#1}
\providecommand{\url}[1]{\texttt{#1}}
\providecommand{\urlprefix}{URL }
\expandafter\ifx\csname urlstyle\endcsname\relax
  \providecommand{\doi}[1]{DOI:\discretionary{}{}{}#1}\else
  \providecommand{\doi}{DOI:\discretionary{}{}{}\begingroup
  \urlstyle{rm}\Url}\fi

\bibitem[{Alami et~al.(1994)Alami, Laumond and Sim\'eon}]{AlamiTwoProbs}
Alami R, Laumond JP and Sim\'eon T (1994) Two manipulation planning algorithms.
\newblock In: \emph{Workshop on Algorithmic Foundations of Robotics (WAFR)}.
\newblock \urlprefix\url{http://dl.acm.org/citation.cfm?id=215085}.

\bibitem[{Alami et~al.(1990)Alami, Sim{\'e}on and Laumond}]{Alami91}
Alami R, Sim{\'e}on T and Laumond JP (1990) A geometrical approach to planning
  manipulation tasks. the case of discrete placements and grasps.
\newblock In: \emph{International Symposium of Robotic Research (ISRR)}.
\newblock \urlprefix\url{http://dl.acm.org/citation.cfm?id=112736}.

\bibitem[{Alur et~al.(1995)Alur, Courcoubetis, Halbwachs, Henzinger, Ho,
  Nicollin, Olivero, Sifakis and Yovine}]{alur1995algorithmic}
Alur R, Courcoubetis C, Halbwachs N, Henzinger TA, Ho PH, Nicollin X, Olivero
  A, Sifakis J and Yovine S (1995) The algorithmic analysis of hybrid systems.
\newblock \emph{Theoretical computer science} 138(1): 3--34.

\bibitem[{Alur et~al.(2000)Alur, Henzinger, Lafferriere and
  Pappas}]{alur2000discrete}
Alur R, Henzinger TA, Lafferriere G and Pappas GJ (2000) Discrete abstractions
  of hybrid systems.
\newblock \emph{Proceedings of the IEEE} 88(7): 971--984.

\bibitem[{B{\"a}ckstr{\"o}m and Nebel(1995)}]{backstrom1995complexity}
B{\"a}ckstr{\"o}m C and Nebel B (1995) Complexity results for {SAS+} planning.
\newblock \emph{Computational Intelligence} 11(4): 625--655.

\bibitem[{Barraquand et~al.(1997)Barraquand, Kavraki, Latombe, Motwani, Li and
  Raghavan}]{Barraquand97}
Barraquand J, Kavraki L, Latombe JC, Motwani R, Li TY and Raghavan P (1997) A
  random sampling scheme for path planning.
\newblock \emph{International Journal of Robotics Research (IJRR)} 16(6):
  759--774.

\bibitem[{Barry et~al.(2013)Barry, Kaelbling and
  Lozano-P{\'e}rez}]{barry2013hierarchical}
Barry J, Kaelbling LP and Lozano-P{\'e}rez T (2013) A hierarchical approach to
  manipulation with diverse actions.
\newblock In: \emph{Robotics and Automation (ICRA), 2013 IEEE International
  Conference on}. IEEE, pp. 1799--1806.
\newblock
  \urlprefix\url{http://citeseerx.ist.psu.edu/viewdoc/summary?doi=10.1.1.365.1060}.

\bibitem[{Barry(2013)}]{barry2013manipulation}
Barry JL (2013) \emph{Manipulation with diverse actions}.
\newblock PhD Thesis, Massachusetts Institute of Technology.

\bibitem[{Berenson and Srinivasa(2010)}]{berenson2010probabilistically}
Berenson D and Srinivasa SS (2010) Probabilistically complete planning with
  end-effector pose constraints.
\newblock In: \emph{Robotics and Automation (ICRA), 2010 IEEE International
  Conference on}. IEEE, pp. 2724--2730.

\bibitem[{Bohlin and Kavraki(2000)}]{bohlin2000path}
Bohlin R and Kavraki LE (2000) Path planning using lazy {PRM}.
\newblock In: \emph{IEEE International Conference on Robotics and Automation
  (ICRA)}, volume~1. IEEE, pp. 521--528.
\newblock \urlprefix\url{http://ieeexplore.ieee.org/document/844107/}.

\bibitem[{Bonet and Geffner(2001)}]{bonet2001planning}
Bonet B and Geffner H (2001) Planning as heuristic search.
\newblock \emph{Artificial Intelligence} 129(1): 5--33.

\bibitem[{Cambon et~al.(2009)Cambon, Alami and Gravot}]{Cambon}
Cambon S, Alami R and Gravot F (2009) A hybrid approach to intricate motion,
  manipulation and task planning.
\newblock \emph{International Journal of Robotics Research (IJRR)} 28.
\newblock
  \urlprefix\url{http://journals.sagepub.com/doi/abs/10.1177/0278364908097884}.

\bibitem[{Dantam et~al.(2016)Dantam, Kingston, Chaudhuri and
  Kavraki}]{dantam2016tmp}
Dantam NT, Kingston Z, Chaudhuri S and Kavraki LE (2016) Incremental task and
  motion planning: A constraint-based approach.
\newblock In: \emph{Robotics: Science and Systems (RSS)}.
\newblock \urlprefix\url{http://www.roboticsproceedings.org/rss12/p02.pdf}.

\bibitem[{{de Silva} et~al.(2013){de Silva}, Pandey, Gharbi and
  Alami}]{deSilva}
{de Silva} L, Pandey AK, Gharbi M and Alami R (2013) Towards combining {HTN}
  planning and geometric task planning.
\newblock In: \emph{RSS Workshop on Combined Robot Motion Planning and AI
  Planning for Practical Applications}.
\newblock \urlprefix\url{https://arxiv.org/abs/1307.1482}.

\bibitem[{Dechter(1992)}]{dechter1992constraint}
Dechter R (1992) Constraint networks.
\newblock Technical report, Information and Computer Science, University of
  California, Irvine.
\newblock \urlprefix\url{http://www.ics.uci.edu/~csp/r17-survey.pdf}.

\bibitem[{Dellin and Srinivasa(2016)}]{dellin2016unifying}
Dellin CM and Srinivasa SS (2016) A unifying formalism for shortest path
  problems with expensive edge evaluations via lazy best-first search over
  paths with edge selectors.
\newblock \emph{International Conference on Automated Planning and Scheduling
  (ICAPS)} \urlprefix\url{https://arxiv.org/abs/1603.03490}.

\bibitem[{Deshpande et~al.(2016)Deshpande, Kaelbling and
  Lozano-P?erez}]{deshpande2016tamp}
Deshpande A, Kaelbling LP and Lozano-P?erez T (2016) Decidability of
  semi-holonomic prehensile task and motion planning.
\newblock \emph{Workshop on Algorithmic Foundations of Robotics (WAFR)}
  \urlprefix\url{http://lis.csail.mit.edu/pubs/deshpande-WAFR16.pdf}.

\bibitem[{Diankov(2010)}]{diankov2010automated}
Diankov R (2010) \emph{Automated construction of robotic manipulation
  programs}.
\newblock PhD Thesis, Robotics Institute, Carnegie Mellon University.

\bibitem[{Diankov and Kuffner(2008)}]{openrave}
Diankov R and Kuffner J (2008) Openrave: A planning architecture for autonomous
  robotics.
\newblock Technical Report CMU-RI-TR-08-34, Robotics Institute, Carnegie Mellon
  University.
\newblock
  \urlprefix\url{https://pdfs.semanticscholar.org/c28d/3dc33b629916a306cc58cbff05dcd632d42d.pdf}.

\bibitem[{Dornhege et~al.(2009)Dornhege, Eyerich, Keller, Tr{\"u}g, Brenner and
  Nebel}]{dornhege09icaps}
Dornhege C, Eyerich P, Keller T, Tr{\"u}g S, Brenner M and Nebel B (2009)
  Semantic attachments for domain-independent planning systems.
\newblock In: \emph{International Conference on Automated Planning and
  Scheduling (ICAPS)}. AAAI Press, pp. 114--121.
\newblock
  \urlprefix\url{https://www.aaai.org/ocs/index.php/ICAPS/ICAPS09/paper/viewPaper/754}.

\bibitem[{Dornhege et~al.(2013)Dornhege, Hertle and Nebel}]{dornhege13irosws}
Dornhege C, Hertle A and Nebel B (2013) Lazy evaluation and subsumption caching
  for search-based integrated task and motion planning.
\newblock In: \emph{IEEE/RSJ International Conference on Intelligent Robots and
  Systems (IROS) Workshop on AI-based robotics}.
\newblock
  \urlprefix\url{https://robohow.eu/_media/workshops/ai-based-robotics-iros-2013/paper08-final.pdf}.

\bibitem[{Edelkamp(2004)}]{edelkamp2004pddl2}
Edelkamp S (2004) Pddl2.2: The language for the classical part of the 4th
  international planning competition.
\newblock \emph{4th International Planning Competition (IPC'04), at ICAPS'04.}
  \urlprefix\url{https://pdfs.semanticscholar.org/4b3c/0706d2673d817cc7c33e580858e65b134ba2.pdf}.

\bibitem[{Erdem et~al.(2011)Erdem, Haspalamutgil, Palaz, Patoglu and
  Uras}]{Erdem}
Erdem E, Haspalamutgil K, Palaz C, Patoglu V and Uras T (2011) Combining
  high-level causal reasoning with low-level geometric reasoning and motion
  planning for robotic manipulation.
\newblock In: \emph{IEEE International Conference on Robotics and Automation
  (ICRA)}.

\bibitem[{Erol et~al.(1994)Erol, Hendler and Nau}]{erol1994htn}
Erol K, Hendler J and Nau DS (1994) Htn planning: Complexity and expressivity.
\newblock In: \emph{AAAI}, volume~94. pp. 1123--1128.

\bibitem[{Fikes and Nilsson(1971)}]{Fikes71}
Fikes RE and Nilsson NJ (1971) {STRIPS}: A new approach to the application of
  theorem proving to problem solving.
\newblock \emph{Artificial Intelligence} 2: 189--208.

\bibitem[{Garrett et~al.(2014)Garrett, Lozano-P\'{e}rez and
  Kaelbling}]{GarrettWAFR14}
Garrett CR, Lozano-P\'{e}rez T and Kaelbling LP (2014) {FFRob}: An efficient
  heuristic for task and motion planning.
\newblock In: \emph{Workshop on the Algorithmic Foundations of Robotics
  (WAFR)}.
\newblock
  \urlprefix\url{https://link.springer.com/chapter/10.1007%2F978-3-319-16595-0_11}.

\bibitem[{Garrett et~al.(2015)Garrett, Lozano-P{\'e}rez and
  Kaelbling}]{GarrettIROS15}
Garrett CR, Lozano-P{\'e}rez T and Kaelbling LP (2015) Backward-forward search
  for manipulation planning.
\newblock In: \emph{IEEE/RSJ International Conference on Intelligent Robots and
  Systems (IROS)}.
\newblock \urlprefix\url{http://lis.csail.mit.edu/pubs/garrett-iros15.pdf}.

\bibitem[{Garrett et~al.(2017{\natexlab{a}})Garrett, Lozano-Perez and
  Kaelbling}]{garrettIJRR2017}
Garrett CR, Lozano-Perez T and Kaelbling LP (2017{\natexlab{a}}) Ffrob:
  Leveraging symbolic planning for efficient task and motion planning.
\newblock \emph{The International Journal of Robotics Research}
  \doi{10.1177/0278364917739114}.
\newblock \urlprefix\url{https://arxiv.org/abs/1608.01335}.

\bibitem[{Garrett et~al.(2017{\natexlab{b}})Garrett, Lozano-P{\'e}rez and
  Kaelbling}]{garrett2017strips}
Garrett CR, Lozano-P{\'e}rez T and Kaelbling LP (2017{\natexlab{b}}) Strips
  planning in infinite domains.
\newblock \emph{arXiv preprint arXiv:1701.00287} .

\bibitem[{Hauser and Latombe(2010)}]{HauserLatombe}
Hauser K and Latombe JC (2010) Multi-modal motion planning in non-expansive
  spaces.
\newblock \emph{International Journal of Robotics Research (IJRR)} 29:
  897--915.

\bibitem[{Hauser and Ng-Thow-Hing(2011)}]{HauserIJRR11}
Hauser K and Ng-Thow-Hing V (2011) Randomized multi-modal motion planning for a
  humanoid robot manipulation task.
\newblock \emph{International Journal of Robotics Research (IJRR)} 30(6):
  676--698.
\newblock
  \urlprefix\url{http://journals.sagepub.com/doi/abs/10.1177/0278364910386985}.

\bibitem[{Helmert(2006)}]{helmert2006fast}
Helmert M (2006) The fast downward planning system.
\newblock \emph{Journal of Artificial Intelligence Research (JAIR)} 26:
  191--246.
\newblock \urlprefix\url{http://www.jair.org/papers/paper1705.html}.

\bibitem[{Henzinger(2000)}]{henzinger2000theory}
Henzinger TA (2000) The theory of hybrid automata.
\newblock In: \emph{Verification of Digital and Hybrid Systems}. Springer, pp.
  265--292.

\bibitem[{Hoffmann and Nebel(2001)}]{HoffmannN01}
Hoffmann J and Nebel B (2001) The {FF} planning system: Fast plan generation
  through heuristic search.
\newblock \emph{Journal Artificial Intelligence Research (JAIR)} 14: 253--302.
\newblock \urlprefix\url{http://dl.acm.org/citation.cfm?id=1622404}.

\bibitem[{Hsu et~al.(1997)Hsu, Latombe and Motwani}]{hsu1997path}
Hsu D, Latombe JC and Motwani R (1997) Path planning in expansive configuration
  spaces.
\newblock In: \emph{Robotics and Automation, 1997. Proceedings., 1997 IEEE
  International Conference on}, volume~3. IEEE, pp. 2719--2726.

\bibitem[{Jensen(1996)}]{jensen1996introduction}
Jensen FV (1996) \emph{An introduction to Bayesian networks}, volume 210.
\newblock UCL press London.

\bibitem[{Kaelbling and Lozano-P{\'e}rez(2011)}]{HPN}
Kaelbling LP and Lozano-P{\'e}rez T (2011) Hierarchical planning in the now.
\newblock In: \emph{IEEE International Conference on Robotics and Automation
  (ICRA)}.
\newblock \urlprefix\url{http://ieeexplore.ieee.org/document/5980391/}.

\bibitem[{Karaman and Frazzoli(2011)}]{karaman2011sampling}
Karaman S and Frazzoli E (2011) Sampling-based algorithms for optimal motion
  planning.
\newblock \emph{The International Journal of Robotics Research} 30(7):
  846--894.

\bibitem[{Kavraki et~al.(1998)Kavraki, Kolountzakis and
  Latombe}]{kavraki1998analysis}
Kavraki LE, Kolountzakis MN and Latombe JC (1998) Analysis of probabilistic
  roadmaps for path planning.
\newblock \emph{Robotics and Automation, IEEE Transactions on} 14(1): 166--171.

\bibitem[{Kavraki and Latombe(1998)}]{Kavraki98probabilisticroadmaps}
Kavraki LE and Latombe JC (1998) Probabilistic roadmaps for robot path
  planning.
\newblock \emph{Practical Motion Planning in Robotics: Current Approaches and
  Future Directions} .

\bibitem[{Kavraki et~al.(1995)Kavraki, Latombe, Motwani and
  Raghavan}]{kavraki1995randomized}
Kavraki LE, Latombe JC, Motwani R and Raghavan P (1995) Randomized query
  processing in robot path planning.
\newblock In: \emph{Proceedings of the twenty-seventh annual ACM symposium on
  Theory of computing}. ACM, pp. 353--362.

\bibitem[{Kavraki et~al.(1996)Kavraki, Svestka, Latombe and
  Overmars}]{Kavraki96}
Kavraki LE, Svestka P, Latombe JC and Overmars MH (1996) Probabilistic roadmaps
  for path planning in high-dimensional configuration spaces.
\newblock \emph{IEEE Transactions on Robotics and Automation} 12(4): 566--580.
\newblock \urlprefix\url{http://ieeexplore.ieee.org/document/508439/}.

\bibitem[{Krontiris and Bekris(2015)}]{krontirisRSS2015}
Krontiris A and Bekris KE (2015) Dealing with difficult instances of object
  rearrangement.
\newblock In: \emph{Robotics: Science and Systems (RSS)}. Rome, Italy.
\newblock
  \urlprefix\url{http://www.cs.rutgers.edu/~kb572/pubs/Krontiris_Bekris_rearrangement_RSS2015.pdf}.

\bibitem[{Krontiris and Bekris(2016)}]{krontiris2016icra}
Krontiris A and Bekris KE (2016) Efficiently solving general rearrangement
  tasks: A fast extension primitive for an incremental sampling-based planner.
\newblock In: \emph{International Conference on Robotics and Automation
  (ICRA)}. Stockholm, Sweden.
\newblock
  \urlprefix\url{http://www.cs.rutgers.edu/~kb572/pubs/fast_object_rearrangement.pdf}.

\bibitem[{Kuffner and {LaValle}(2000)}]{KuffnerLaValle}
Kuffner JJ Jr and {LaValle} SM (2000) {RRT-Connect}: An efficient approach to
  single-query path planning.
\newblock In: \emph{IEEE International Conference on Robotics and Automation
  (ICRA)}.

\bibitem[{Lagriffoul et~al.(2014)Lagriffoul, Dimitrov, Bidot, Saffiotti and
  Karlsson}]{lagriffoul2014efficiently}
Lagriffoul F, Dimitrov D, Bidot J, Saffiotti A and Karlsson L (2014)
  Efficiently combining task and motion planning using geometric constraints.
\newblock \emph{International Journal of Robotics Research (IJRR)} :
  0278364914545811\urlprefix\url{http://journals.sagepub.com/doi/abs/10.1177/0278364914545811?journalCode=ijra}.

\bibitem[{Lagriffoul et~al.(2012)Lagriffoul, Dimitrov, Saffiotti and
  Karlsson}]{LagriffoulDSK12}
Lagriffoul F, Dimitrov D, Saffiotti A and Karlsson L (2012) Constraint
  propagation on interval bounds for dealing with geometric backtracking.
\newblock In: \emph{IEEE/RSJ International Conference on Intelligent Robots and
  Systems (IROS)}.
\newblock \urlprefix\url{http://ieeexplore.ieee.org/document/6385972/}.

\bibitem[{Laumond(1998)}]{Laumond:1998:RMP:521883}
Laumond JPP (1998) \emph{Robot Motion Planning and Control}.
\newblock Secaucus, NJ, USA: Springer-Verlag New York, Inc.
\newblock ISBN 3540762191.

\bibitem[{{LaValle}(2006)}]{Lavalle06}
{LaValle} SM (2006) \emph{Planning Algorithms}.
\newblock Cambridge University Press.
\newblock \urlprefix\url{msl.cs.uiuc.edu/planning}.

\bibitem[{Lozano-P\'erez(1981)}]{LozanoPerez81}
Lozano-P\'erez T (1981) Automatic planning of manipulator transfer movements.
\newblock \emph{IEEE Transactions on Systems, Man, and Cybernetics} 11:
  681--698.
\newblock \urlprefix\url{http://ieeexplore.ieee.org/document/4308589/}.

\bibitem[{Lozano-P\'{e}rez et~al.(1987)Lozano-P\'{e}rez, Jones, Mazer,
  O'Donnell, Grimson, Tournassoud and Lanusse}]{handeyICRA87}
Lozano-P\'{e}rez T, Jones JL, Mazer E, O'Donnell PA, Grimson WEL, Tournassoud P
  and Lanusse A (1987) Handey: A robot system that recognizes, plans, and
  manipulates.
\newblock In: \emph{IEEE International Conference on Robotics and Automation
  (ICRA)}.
\newblock \urlprefix\url{http://ieeexplore.ieee.org/document/1087847/}.

\bibitem[{Lozano-P{\'e}rez and Kaelbling(2014)}]{lozano2014constraint}
Lozano-P{\'e}rez T and Kaelbling LP (2014) A constraint-based method for
  solving sequential manipulation planning problems.
\newblock In: \emph{IEEE/RSJ International Conference on Intelligent Robots and
  Systems (IROS)}. IEEE, pp. 3684--3691.
\newblock \urlprefix\url{http://lis.csail.mit.edu/pubs/tlpk-iros14.pdf}.

\bibitem[{McDermott et~al.(1998)McDermott, Ghallab, Howe, Knoblock, Ram,
  Veloso, Weld and Wilkins}]{mcdermott1998pddl}
McDermott D, Ghallab M, Howe A, Knoblock C, Ram A, Veloso M, Weld D and Wilkins
  D (1998) Pddl: The planning domain definition language.
\newblock Technical report, Yale Center for Computational Vision and Control.
\newblock
  \urlprefix\url{http://citeseerx.ist.psu.edu/viewdoc/summary?doi=10.1.1.51.9941}.

\bibitem[{Pandey et~al.(2012)Pandey, Saut, Sidobre and Alami}]{Pandey12}
Pandey AK, Saut JP, Sidobre D and Alami R (2012) Towards planning human-robot
  interactive manipulation tasks: Task dependent and human oriented autonomous
  selection of grasp and placement.
\newblock In: \emph{RAS/EMBS International Conference on Biomedical Robotics
  and Biomechatronics}.
\newblock
  \urlprefix\url{http://ieeexplore.ieee.org/abstract/document/6290776/}.

\bibitem[{Plaku and Hager(2010)}]{Plaku}
Plaku E and Hager G (2010) Sampling-based motion planning with symbolic,
  geometric, and differential constraints.
\newblock In: \emph{IEEE International Conference on Robotics and Automation
  (ICRA)}.
\newblock \urlprefix\url{http://ieeexplore.ieee.org/document/5509563/}.

\bibitem[{Sim{\'e}on et~al.(2004)Sim{\'e}on, Laumond, Cort{\'e}s and
  Sahbani}]{simeon2004manipulation}
Sim{\'e}on T, Laumond JP, Cort{\'e}s J and Sahbani A (2004) Manipulation
  planning with probabilistic roadmaps.
\newblock \emph{International Journal of Robotics Research (IJRR)} 23(7-8):
  729--746.
\newblock
  \urlprefix\url{http://journals.sagepub.com/doi/abs/10.1177/0278364904045471}.

\bibitem[{Smith(2005)}]{smith2005open}
Smith R (2005) Open dynamics engine.

\bibitem[{Srivastava et~al.(2014)Srivastava, Fang, Riano, Chitnis, Russell and
  Abbeel}]{Srivastava14}
Srivastava S, Fang E, Riano L, Chitnis R, Russell S and Abbeel P (2014)
  Combined task and motion planning through an extensible planner-independent
  interface layer.
\newblock In: \emph{IEEE International Conference on Robotics and Automation
  (ICRA)}.
\newblock \urlprefix\url{http://ieeexplore.ieee.org/document/6906922/}.

\bibitem[{Stilman and Kuffner(2006)}]{StilmanWAFR06}
Stilman M and Kuffner JJ (2006) Planning among movable obstacles with
  artificial constraints.
\newblock In: \emph{Workshop on Algorithmic Foundations of Robotics (WAFR)}.

\bibitem[{Stilman et~al.(2007)Stilman, Schamburek, Kuffner and
  Asfour}]{StilmanICRA07}
Stilman M, Schamburek JU, Kuffner JJ and Asfour T (2007) Manipulation planning
  among movable obstacles.
\newblock In: \emph{IEEE International Conference on Robotics and Automation
  (ICRA)}.
\newblock \urlprefix\url{http://ieeexplore.ieee.org/document/4209604/}.

\bibitem[{Thi{\'e}baux et~al.(2005)Thi{\'e}baux, Hoffmann and
  Nebel}]{thiebaux2005defense}
Thi{\'e}baux S, Hoffmann J and Nebel B (2005) In defense of pddl axioms.
\newblock \emph{Artificial Intelligence} 168(1-2): 38--69.

\bibitem[{Torrisi and Bemporad(2001)}]{torrisi2001discrete}
Torrisi FD and Bemporad A (2001) Discrete-time hybrid modeling and
  verification.
\newblock In: \emph{Proc. 40th IEEE Conf. on Decision and Control}, volume~3.
  IEEE, pp. 2899--2904.

\bibitem[{Toussaint(2015)}]{toussaint2015logic}
Toussaint M (2015) Logic-geometric programming: an optimization-based approach
  to combined task and motion planning.
\newblock In: \emph{AAAI Conference on Artificial Intelligence}. AAAI Press,
  pp. 1930--1936.
\newblock \urlprefix\url{https://www.ijcai.org/Proceedings/15/Papers/274.pdf}.

\bibitem[{Toussaint and Lopes(2017)}]{toussaint2017multi}
Toussaint M and Lopes M (2017) Multi-bound tree search for logic-geometric
  programming in cooperative manipulation domains.
\newblock In: \emph{Robotics and Automation (ICRA), 2017 IEEE International
  Conference on}. IEEE, pp. 4044--4051.

\bibitem[{Tu(2010)}]{tu2010manifolds}
Tu LW (2010) \emph{An Introduction to Manifolds}.
\newblock Springer.
\newblock \urlprefix\url{http://www.springer.com/gp/book/9781441973993}.

\bibitem[{Van Den~Berg et~al.(2009)Van Den~Berg, Stilman, Kuffner, Lin and
  Manocha}]{van2009path}
Van Den~Berg J, Stilman M, Kuffner J, Lin M and Manocha D (2009) Path planning
  among movable obstacles: a probabilistically complete approach.
\newblock In: \emph{Algorithmic Foundation of Robotics VIII}. Springer, pp.
  599--614.
\newblock
  \urlprefix\url{https://link.springer.com/chapter/10.1007%2F978-3-642-00312-7_37}.

\bibitem[{Vega-Brown and Roy(2016)}]{vega2016asymptotically}
Vega-Brown W and Roy N (2016) Asymptotically optimal planning under
  piecewise-analytic constraints.
\newblock In: \emph{Workshop on the Algorithmic Foundations of Robotics
  (WAFR)}.
\newblock \urlprefix\url{http://www.wafr.org/papers/WAFR_2016_paper_11.pdf}.

\bibitem[{Vendittelli et~al.(2015)Vendittelli, Laumond and
  Mishra}]{vendittelli2015decidability}
Vendittelli M, Laumond JP and Mishra B (2015) Decidability of robot
  manipulation planning: Three disks in the plane.
\newblock In: \emph{Algorithmic Foundations of Robotics XI}. Springer, pp.
  641--657.

\bibitem[{Wilfong(1988)}]{Wilfong89}
Wilfong GT (1988) Motion planning in the presence of movable obstacles.
\newblock In: \emph{Symposium on Computational Geometry}. pp. 279--288.
\newblock \urlprefix\url{https://link.springer.com/article/10.1007/BF01530890}.

\end{thebibliography}
\end{document}